\documentclass[a4paper, 10pt, conference]{ieeeconf}      

\IEEEoverridecommandlockouts                              

\usepackage{graphics} 
\usepackage{epsfig} 
\usepackage{amsmath} 
\usepackage{amssymb}  

\usepackage{booktabs}
\usepackage{nicefrac}       

\usepackage{graphicx}

\newcommand{\vecW}{\mathbf{W}}
\newcommand{\vecX}{\mathbf{X}}
\newcommand{\vecY}{\mathbf{Y}}

\newcommand{\R}{\mathbb{R}}

\newcommand{\yes}{\checkmark}
\newcommand{\no}{$\times$}
\usepackage[caption=false]{subfig}
\newcommand{\best}[1]{\textbf{#1}}

\usepackage{xcolor}   
\newcommand{\PAR}[1]{\noindent{\bf #1}}
\usepackage[pagebackref=true,breaklinks=true,letterpaper=true,colorlinks,bookmarks=false]{hyperref}
\usepackage[capitalize]{cleveref}
\crefname{section}{Sec.}{Secs.}
\Crefname{section}{Section}{Sections}
\Crefname{table}{Table}{Tables}
\crefname{table}{Tab.}{Tabs.}

\title{\LARGE \bf
HRFuser: A Multi-resolution Sensor Fusion Architecture\\for 2D Object Detection 
}

\author{Tim Br{\"o}dermann$^{1}$, Christos Sakaridis$^{1}$, Dengxin Dai$^{2}$ and Luc Van Gool$^{1,3}$
\thanks{$^{1}$Computer Vision Lab, ETH Zurich, 8092 Zurich, Switzerland {\tt\small \{timbr,csakarid,vangool\}@vision.ee.ethz.ch}}%
\thanks{$^{2}$VAS, MPI for Informatics, 66123 Saarbrücken, Germany {\tt\small ddai@mpi-inf.mpg.de}}%
\thanks{$^{2}$ESAT, KU Leuven, 3001 Leuven, Belgium}%
}

\begin{document}

\maketitle
\thispagestyle{empty}
\pagestyle{empty}

\begin{abstract}

    Besides standard cameras, autonomous vehicles typically include multiple additional sensors, such as lidars and radars, which help acquire richer information for perceiving the content of the driving scene. While several recent works focus on fusing certain pairs of sensors---such as camera with lidar or radar---by using architectural components specific to the examined setting, a generic and modular sensor fusion architecture is missing from the literature. In this work, we propose HRFuser, a modular architecture for multi-modal 2D object detection. It fuses multiple sensors in a multi-resolution fashion and scales to an arbitrary number of input modalities. The design of HRFuser is based on state-of-the-art high-resolution networks for image-only dense prediction and incorporates a novel multi-window cross-attention block as the means to perform fusion of multiple modalities at multiple resolutions. We demonstrate via extensive experiments on nuScenes and the adverse conditions DENSE datasets that our model effectively leverages complementary features from additional modalities, substantially improving upon camera-only performance and consistently outperforming state-of-the-art 3D and 2D fusion methods evaluated on 
    2D object detection metrics.
     The source code is publicly available at \url{https://github.com/timbroed/HRFuser}

\end{abstract}

\section{Introduction}
\label{sec:intro}

High-level visual perception is vital for the deployment of autonomous vehicles and robots. The primary sensors for such agents to perceive the surrounding scene are cameras, as they provide rich texture information at very high spatial resolution. This enables perception algorithms to achieve high accuracy in central tasks, such as object detection and semantic segmentation.

However, to attain full autonomy, systems require perception algorithms that perform robustly in all encountered conditions, but the quality of images degrades severely in adverse visual conditions, such as night-time, rainfall, snowfall, or fog.
Moreover, camera readings do not explicitly capture depth or other geometric attributes of the scene. 
Complementary characteristics to cameras are provided by other sensors: lidars and radars provide explicit range measurements, while radars and gated cameras feature robustness to adverse weather~\cite{seeing:through:fog}.
Thanks to developments in sensor technology,
these types of sensors are becoming cheaper and thus more commonly used in automated driving. Thus, exploiting \emph{all} measurements from the sensor suite of an autonomous system via sensor fusion is of utmost importance for accurate perception under all possible conditions.

\begin{figure}
    \centering
    \subfloat{\includegraphics[clip,width=0.33\linewidth,trim=30mm 20mm 150mm 60mm]{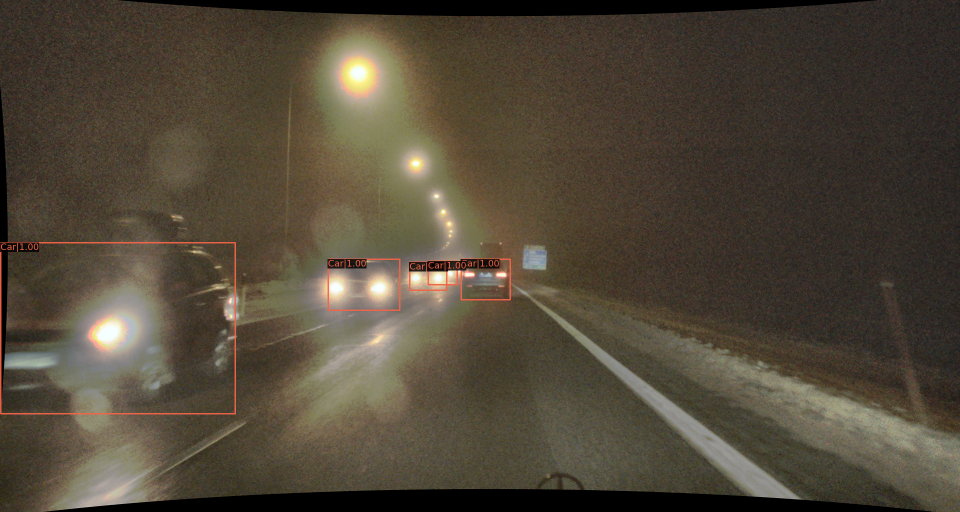}}
    \hfil
    \subfloat{\includegraphics[clip,width=0.33\linewidth,trim=60mm 80mm 177mm 70mm]{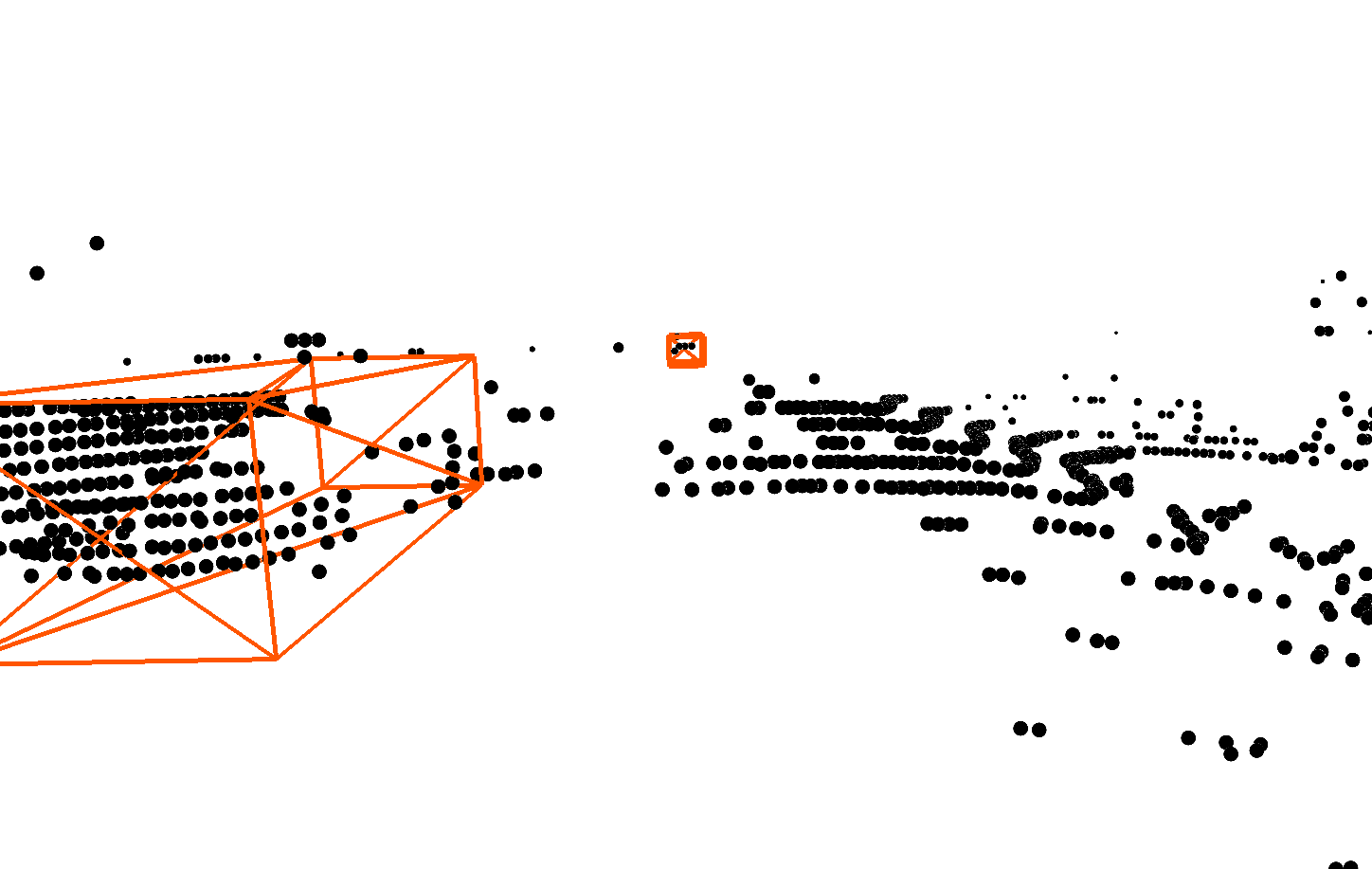}}
    \hfil
    \subfloat{\includegraphics[clip,width=0.33\linewidth,trim=30mm 20mm 150mm 60mm]{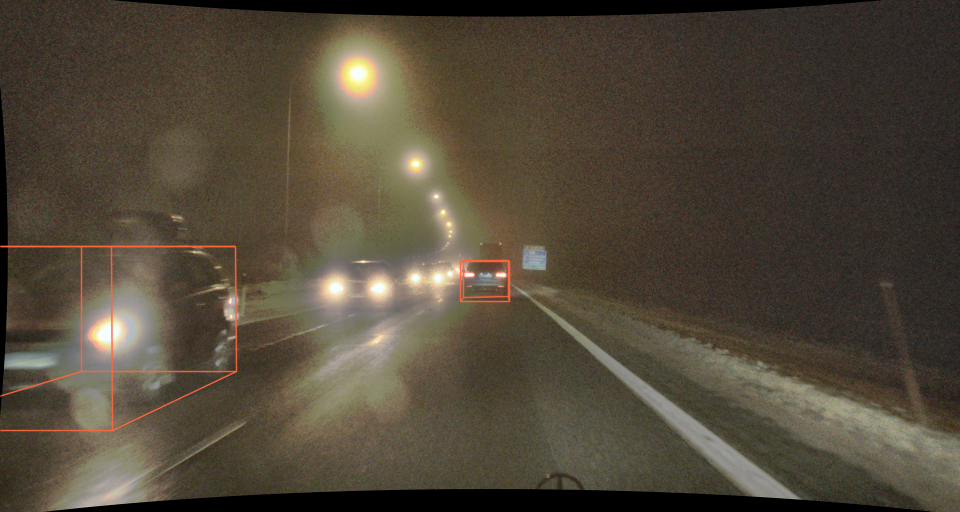}}
    \caption{ An example scene from DENSE~\cite{seeing:through:fog} with (left) 2D and (right) 3D object annotations. Multiple safety-relevant objects are missing in the (middle) point cloud due to weather deterioration and thus receive only 2D and not 3D annotations.
    }
    \label{fig:dense:2d:3d}
\end{figure}

Besides cameras, adverse weather conditions can also severely affect the measurements of lidars~\cite{influences:of:weather:on:laser:radar:systems,fog:simulation:lidar,snowfall:simulation:lidar}. 
This in turn results in lidar-based 3D object annotations being incomplete in such conditions. \cref{fig:dense:2d:3d} displays both 2D and 3D labels from the DENSE dataset~\cite{seeing:through:fog} and exemplifies why a significant amount of objects (41.91\% for the ``dense fog'' split of DENSE) can receive only a 2D annotation when correct lidar measurements are missing due to environmental factors such as fog or precipitation. As these measurements do not provide a complete and reliable signal for creating 3D annotations.
However, in difficult driving conditions, it is of utmost importance to detect \emph{all} safety-relevant objects even if their precise localization in 3D is not possible.

We thus focus on 2D object detection, which allows to train and evaluate detection models not only on standard data, such as nuScenes~\cite{nuScenes}, but also on extremely challenging data where 3D annotations are missing due to the factors mentioned above, such as DENSE~\cite{seeing:through:fog}.
We pursue this goal by building a modular architecture that treats the camera as the \emph{primary} modality and adaptively fuses features from an arbitrary number of additional, \emph{secondary} modalities in a modular and scalable manner.

Our network, named HRFuser, consists of a multi-resolution multi-sensor fusion architecture for 2D detection. The structure of HRFuser is based on the paradigm of preserving high-resolution representations throughout all layers of the backbone~\cite{hrnet,hrformer}. We extend this architectural paradigm to multiple modalities and propose an \emph{efficient} fusion design for our HRFuser, which scales well with the number of sensors. 
In particular, HRFuser includes parallel lightweight branches for each of the secondary input modalities.
Solely the primary camera branch constructs additional high-dimensional lower-resolution features.

We repeatable fuse the sensors at multiple levels and at all resolutions of the camera branch. 
To facilitate this, we propose a novel multi-window cross-attention (MWCA) block. This block efficiently performs an attention-based fusion of the camera with each additional sensor in parallel, reducing the quadratic complexity of attention via multiple non-overlapping spatial windows.
MWCA efficiently attends to the useful features of each sensor while ignoring noise, resulting in improved performance from \emph{all} added sensors, even from radar, which is highly noisy.

Our architecture is generic and modular, as it handles all additional sensors in the same way, except for basic pre-processing. This allows leveraging multiple sensors, such as lidar, radar, and gated cameras, without the need to create specialized architectural components dedicated to each individual sensor. Thus, HRFuser is directly applicable to an arbitrary number of sensors. Our novel MWCA fusion and our architecture design with light-weight branches for secondary modalities minimize the computational overhead to only +9.7\% flops and +1.9\% parameters for a single added modality, as detailed in \cref{subsubsec:exp:modablation}. HRFuser also inherits the benefits associated to processing camera features at multiple resolutions while preserving a high-resolution representation, allowing aggregation of global context without loss of fine spatial details.

We conduct a thorough experimental evaluation of our network for 2D detection on two major autonomous driving datasets, the adverse-condition-oriented DENSE~\cite{seeing:through:fog} and the large-scale nuScenes~\cite{nuScenes}. HRFuser substantially outperforms all state-of-the-art 2D sensor fusion and camera-only networks which are heavily engineered for dense prediction tasks. As well as state-of-the-art 3D object detection approaches evaluated in 2D. 
Detailed ablation studies evidence the benefit of our carefully designed network architecture and the novel MWCA fusion block compared to other fusion strategies.

\begin{figure*}
  \vspace{.5em}
    \centering
    \includegraphics[clip,width=\textwidth,trim=5mm 453mm 11mm 7mm]{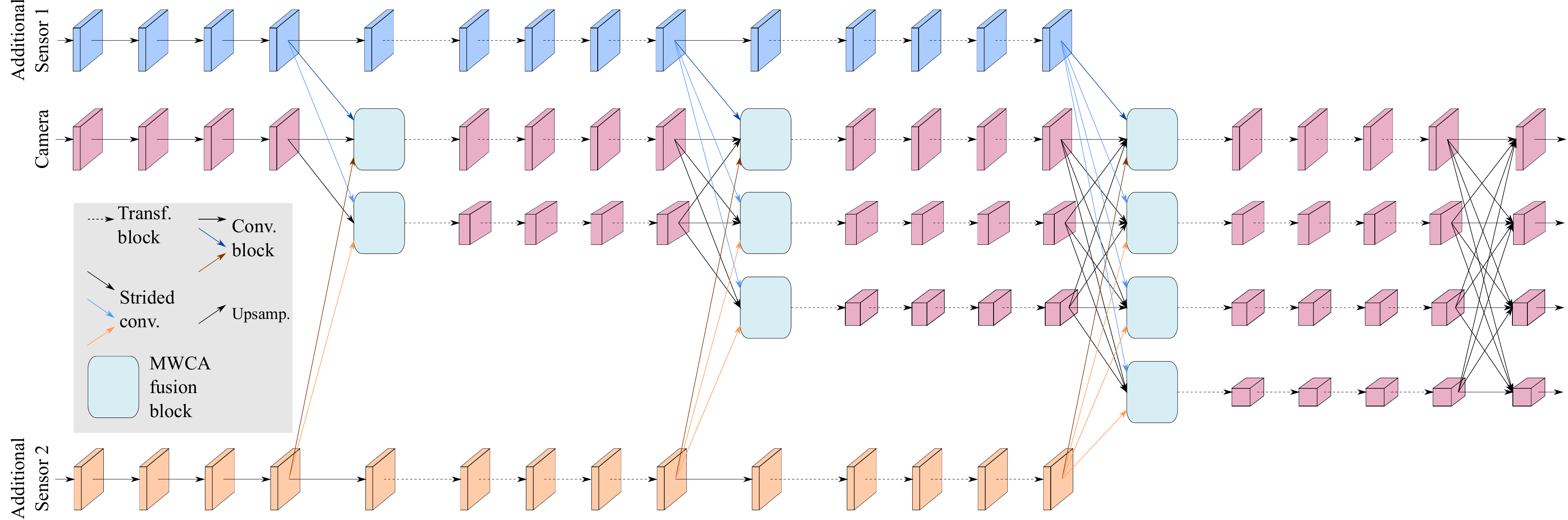}
    \caption{An instantiation of the overall architecture of our HRFuser backbone for the case where two additional sensors besides the camera are available. Feature maps are colored according to the sensor branch to which they belong. For brevity, we only show the backbone of the network and not the detection head. Transf.: transformer, Conv.: convolution, MWCA: multi-window cross-attention.}
    \label{fig:hrfuser}
  \vspace{-1em}
\end{figure*}

\section{Related Work}
\label{sec:related}

\PAR{Object detection} methods output bounding boxes for a given input scene. 
A popular line of work on 2D detection consists in the region-based CNN (R-CNN) framework~\cite{faster:rcnn,cascade:rcnn}, which employs a two-stage pipeline that first generates object proposals and then predicts the final boxes from the proposals. An alternative approach is single-stage detection~\cite{yolo}, which is typically faster but less accurate. Recent approaches 
improve the efficiency of the detectors~\cite{efficientdet:scalable:detection}, 
and adapt networks to adverse conditions such as fog~\cite{SFSU_synthetic}. HRNet~\cite{hrnet} constitutes a CNN backbone for detection that preserves a high resolution for intermediate representations, while aggregating global context via parallel lower-resolution branches. Recently, HRFormer~\cite{hrformer} has extended this idea by replacing most convolutional blocks of HRNet with transformer blocks, which facilitate context aggregation via attending to features from any location of the input.
HRFuser follows the architecture of HRNet and HRFormer with parallel streams of different resolutions, but it extends it to a multi-modal setting, by adding sensor-specific branches and a novel transformer-based fusion block, which allows us to simultaneously fuse information from multiple additional modalities in an adaptive manner.

\PAR{Sensor fusion for object detection} is the primary application of sensor fusion in visual perception, although other tasks~\cite{noise:aware:unsupervised:lidar:stereo:fusion,radar:ghost:target:detection:multimodal:transformers}
have also been studied. For a comprehensive overview of related work, we refer the reader to~\cite{multimodal:detection:segmentation:datasets:methods:challenges}.
The KITTI dataset~\cite{kitti} has catalyzed research in this area by providing recordings of driving scenes with multiple sensors, notably a lidar and a camera, along with object annotations.
Successors of KITTI include nuScenes~\cite{nuScenes} and Argoverse~\cite{argoverse:cvpr19}. Notably, nuScenes also includes radar readings, which are important in adverse-weather scenarios. Such scenarios are explicitly covered in~\cite{seeing:through:fog,CADC,ACDC}.
Based on these sets, several sensor fusion works have been presented that focus on improving lidar-based 3D detection by fusing information from the camera. This category of works ranges from early (low-level) fusion~\cite{pointpainting:3d:detection},
which directly combines the raw lidar data with raw image data or image features, and mid-level fusion~\cite{3d:cvf:3d:detection},
which combines lidar features with image-space features, to late fusion~\cite{deep:end:to:end:3d:person:detection}, which fuses the detection results from lidar and camera, asymmetric fusion~\cite{multiview:3d:detection:network}, which fuses the object-level representations from one modality with data-level or feature-level representations from the other, 
to bird’s-eye-view (BEV) based fusion~\cite{liu2022bevfusion}, which lifts mid-level camera features to a common BEV space. 
Other sensor fusion methods address multi-modal 2D detection; many focus on radar and camera sensors~\cite{deep:learning:based:radar:camera:fusion:architecture,radar:camera:fusion:joint:detection:distance:estimation}.
Methods that improve image-based 2D detection by fusing information only from lidar include~\cite{multimodal:cnn:pedestrian:classification}. Fewer previous works~\cite{seeing:through:fog,low:latency:low:level:fusion:automotive,chaturvedi:pay:Attention:Weather} 
fuse all three modalities, i.e.\ camera, lidar, and radar, for detection. We argue that using all three modalities is relevant, as they provide complementary characteristics which are essential for detection. While most recent works focus on fusing sensors with architectural components specific to individual sensors, we propose a modular fusion architecture for 2D object detection that easily scales to an arbitrary number of input modalities. 
Another feature of our network is the fusion at multiple levels and resolutions, which has also been applied in previous works~\cite{deep:continuous:fusion:3d:detection,seeing:through:fog}. Different from these methods, our approach keeps high-resolution representations for each modality \emph{throughout} the network in parallel with lower-resolution representations, which allows to better preserve details while also exploiting global context for classification.

\PAR{Transformers} \cite{attention:is:all:you:need} gained popularity in computer vision with the vision transformer~\cite{vit:transformers:for:image:recognition}. More recent methods use local windows~\cite{hrformer} and the Pyramid Vision Transformer (PVT)~\cite{pvt:pyramid:vision:transformer} introduces a spatial reduction attention  to reduce the memory footprint.
PVTv2~\cite{PVTv2:vision:transformer} improves upon PVT by
using a linear-complexity attention module.
To adaptively fuse two modalities with each other, cross-attention~\cite{MulT:multimodal:nlp} was introduced.
Different from these works, our transformer-based network handles several modalities instead of only two and fuses them at multiple resolutions, combining both global and local features of the input scene more effectively.
Moreover, we combine local-window attention with cross-attention, thereby reducing the memory footprint and enabling repeated attention-based fusion at high resolutions.

\section{HRFuser}
\label{sec:method}

With HRFuser, we extend the paradigm of preserving high-resolution representations throughout all layers of the backbone~\cite{hrnet,hrformer} to multiple modalities. To this end, we extend the HRFormer~\cite{hrformer} backbone with one additional high-resolution, but low-dimensional, branch for each added input modality besides the camera. These additional, or secondary, modalities are fused repeatedly at multiple resolutions into the branch of the primary modality.

\begin{figure*}
  \vspace{.5em}
    \centering  
    \includegraphics[clip,width=0.94\textwidth,trim=14mm 46mm 17mm 211mm]{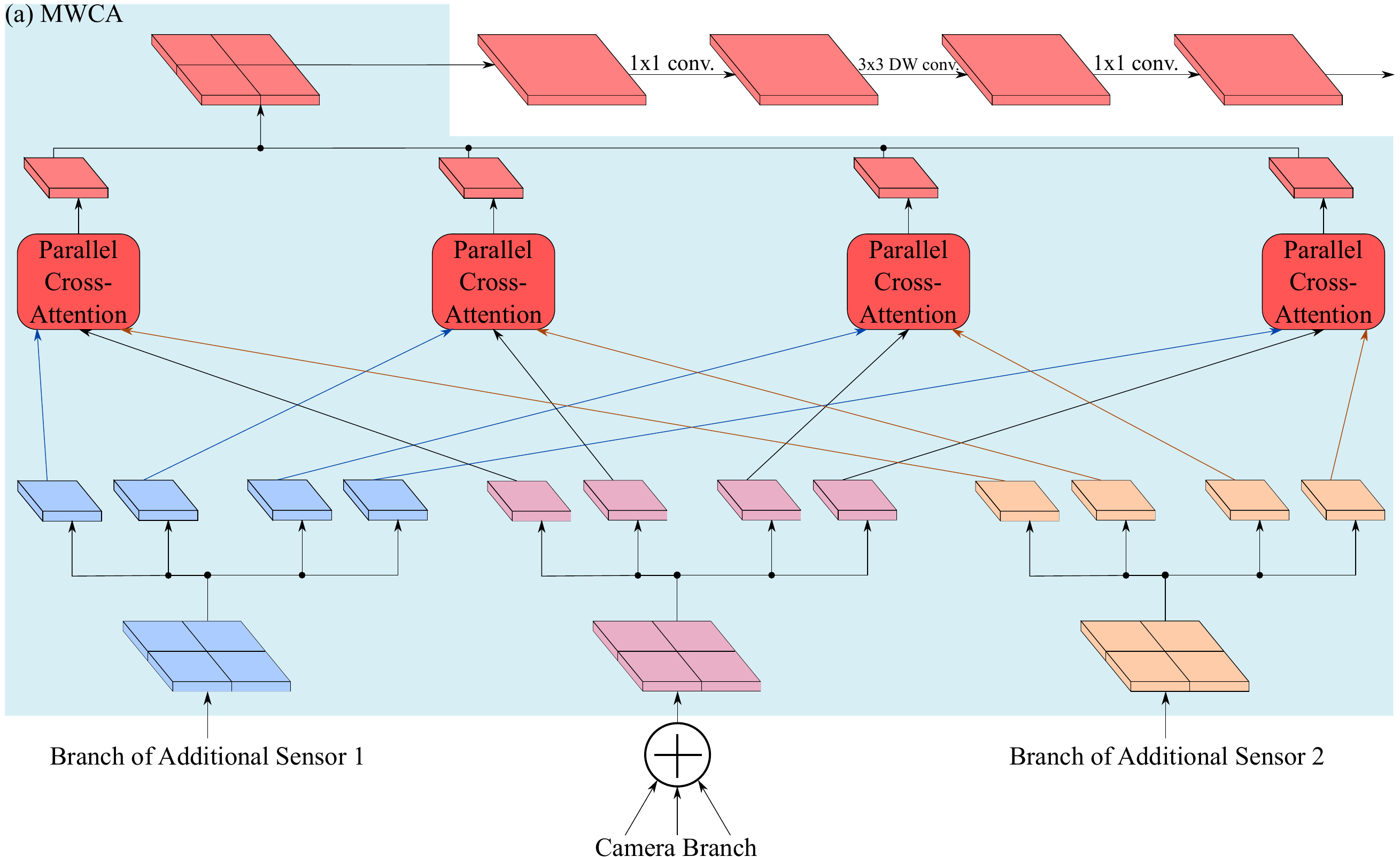}    
    \caption{Our multi-window cross-attention (MWCA) fusion block, consisting of (a) our MWCA and a subsequent feed-forward network. Inputs to the parallel cross-attention blocks are colored according to the sensor they come from. DW conv.: depth-wise convolution.}
    \label{fig:mwca}
  \vspace{-2.em}
\end{figure*}

\cref{fig:hrfuser} illustrates the general architecture of the multi-sensor fusion backbone of HRFuser. The design of the primary branch (camera) follows HRFormer, but is extended with a novel MWCA fusion block, which is further illustrated in \cref{fig:mwca}. The MWCA fusion block is inserted between the multi-resolution fusion module and the subsequent transformer block, allowing the features from the secondary modalities to be fused into the camera branch. All secondary branches continue for three stages and are fused with the primary branch at three levels and four different resolutions. They include feature maps at a single, high resolution, while in the camera branch we introduce lower resolutions as we proceed to later stages, progressively aggregating context. Before applying our MWCA fusion blocks, we add $3{\times}3$ strided convolutions to match the high-resolution secondary modalities to the lower-resolution streams of the primary modality. This down-sampling causes the same $7{\times}7$ local window to progressively cover a larger area of the secondary modalities feature map. Our design, therefore, allows to keep detail in all modalities with the high-resolution stream, while efficiently fusing via local windows and still taking local and more global relationships into account. 

Fusing multiple modalities in such an asymmetric way provides scalability to our method, 
as the complexity increases linearly with the number of added sensors.
We can include an arbitrary number of modalities by simply adding an extra secondary branch for each new modality and fusing it in parallel into the camera branch. We demonstrate this possibility in \cref{sec:experiments} by applying HRFuser to the DENSE dataset~\cite{seeing:through:fog} and utilizing a gated camera as the fourth sensor, besides the more common lidar and radar sensors.

The HRFuser backbone illustrated in \cref{fig:hrfuser} is followed by a neck which forms a feature pyramid by concatenating the upsampled outputs of all streams~\cite{hrnet}. This neck is in turn followed by a Cascade R-CNN head~\cite{cascade:rcnn}, following the widely used two-stage detector architecture. Cascade R-CNN introduces a sequence of detectors trained with increasing Intersection over Union (IoU) thresholds, setting a strong baseline for any given backbone.

\PAR{Multi-window cross-attention.} We propose a novel multi-window cross-attention (MWCA) block to fuse all modalities in parallel by applying multi-head cross-attention (CA) on multiple small non-overlapping local windows. In particular, MWCA limits the spatial extent of the cross-attention to small windows, addressing the quadratic complexity of attention and reducing the computational cost of each attention operation and allows to apply this operation to high-resolution feature maps. For each window, this results in $K^2$ tokens with dimensionality $D$, depending on the number of channels of the stream we fuse into. Compared to self-attention, CA fuses two modalities by applying attention with queries from the primary modality $\alpha$ and keys and values from the secondary modality $\beta$.

More formally, we partition the input feature map $\vecX$ of the primary modality $\alpha$ into a grid of $P$ non-overlapping spatial windows: $\vecX^\alpha \xrightarrow{\text{Split}} \{\vecX_1^\alpha,\,\vecX_2^\alpha,\,\dots,\,\vecX_P^\alpha\}$. Exactly the same partition is applied to the feature maps $\vecY^\beta$ of all secondary modalities $\beta \in \{1,\,\dots,\,M\}$: $\vecY^\beta \xrightarrow{\text{Split}} \{\vecY_1^\beta,\,\vecY_2^\beta,\,\dots,\,\vecY_P^\beta\}$. All input feature maps are vectorized across the spatial dimensions and have the same shape $\vecX,\vecY\in \R^{N{\times}D}$, where $N$ denotes the total number of spatial positions and $D$ denotes the number of channels, and each window is of size $K{\times}K$.

A local transformer applies parallel CA to each corresponding set of windows independently.
Parallel CA on the set of $p$-th windows is formulated as follows:

\setlength{\arraycolsep}{0.0em}
\begin{eqnarray}
    {{\rm{MultiHead}}(\vecX_p^\alpha,\vecY_p^\beta)} = {\rm{Concat}}[&{\rm{head}}(\vecX_p^\alpha,\vecY_p^\beta)_1, \nonumber \\
    &{\dots}, \nonumber \\
    &{\rm{head}}(\vecX_p^\alpha,\vecY_p^\beta)_H] \nonumber \\
    &{\in}\:\R^{K^2\times{}D},
    \label{test}
\end{eqnarray}
\begin{eqnarray}
    {\rm{head}}(\vecX_p^\alpha,\vecY_p^\beta)_{h} =\;&{\rm{Softmax}}\left[\frac{(\vecX_p^\alpha \vecW_q^{h,\beta}) (\vecY_p^\beta\vecW_k^{h,\beta})^T}
    {\sqrt{\nicefrac{D}{H}}} 
    \right] \nonumber \\
    &{\vecY_p^\beta}\,\vecW_v^{h,\beta} \in \R^{K^2\times{}\frac{D}{H}},
    \label{eq:parallel:ca:head}
\end{eqnarray}
\begin{eqnarray}
    \widehat{\vecX}_p =\;&{\vecX_p^\alpha} \nonumber {+}\:\sum_{\beta=1} ^{M}\left[\vecY_p^\beta + {\rm{MultiHead}}(\vecX_p^\alpha,\vecY_p^\beta) \vecW_o^\beta\right] \\
    &\in \R^{K^2\times{}D}
\end{eqnarray}
\setlength{\arraycolsep}{5pt}
where $\vecW_o^\beta  \in \R^{D \times D}$ and $\vecW_q^{h,\beta}$, $\vecW_k^{h,\beta}$, $\vecW_v^{h,\beta} \in \R^{D\times{}\frac{D}{H}}$ for $h\in\{1,\,\dots,\,H\}$ are weight matrices implemented by trainable linear projections. $H$ denotes the number of heads and $\widehat{\vecX}_p$ denotes the output of the parallel CA for the set of $p$-th windows.

\begin{figure}
  \vspace{1.em}
  \begin{center}
    \includegraphics[clip,width=0.6\columnwidth,trim=7mm 206mm 112mm 4mm]{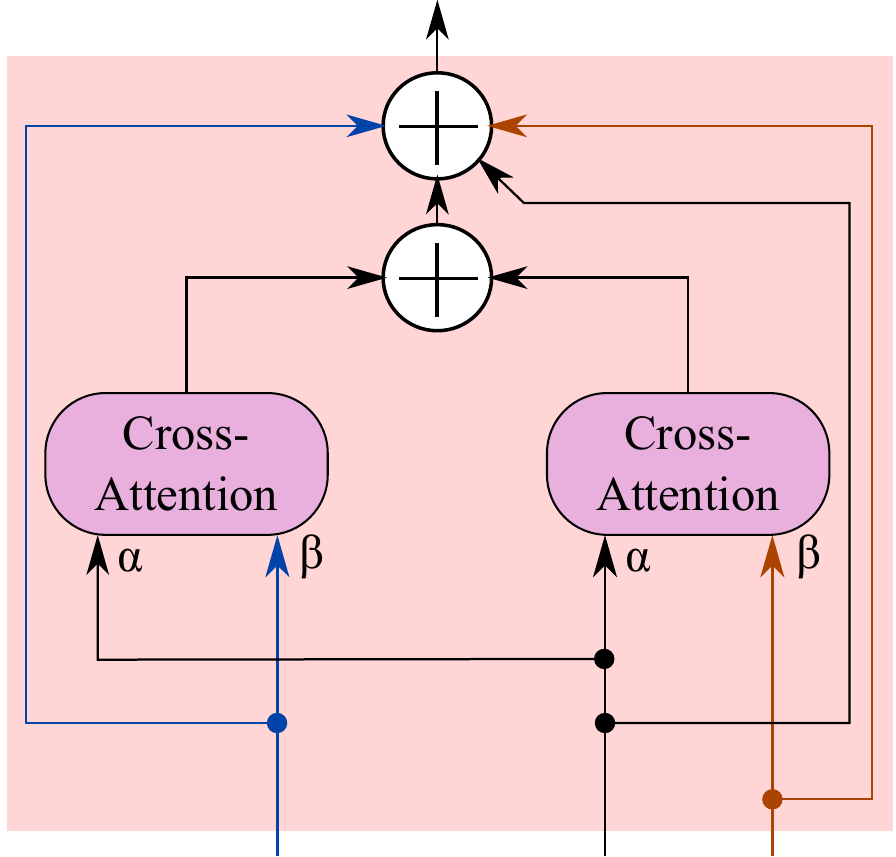}
  \end{center}
  \vspace{-1.em}
  \caption{Our parallel cross-attention block for the case where two additional sensors besides the camera are used. $\alpha$ denotes the primary modality (camera) and $\beta$ denotes the secondary modalities.}
  \label{fig:parallel:ca}
  \vspace{-1.5em}
\end{figure}

We arrange the outputs from all $P$ sets of windows back into a single feature map to get the final output of MWCA, $\vecX^{\rm{MWCA}}$:
\begin{equation}
    \left\{\widehat{\vecX}_1,\,\widehat{\vecX}_2,\,\dots,\,\widehat{\vecX}_P\right\} \xrightarrow{\text{Merge}} \vecX^{\rm{MWCA}}.  
    \label{eq:MWCA}
\end{equation}

\cref{fig:mwca} illustrates how we split up the input maps for each modality into non-overlapping windows and apply parallel CA across modalities within each window independently, before merging the resulting outputs back into a single feature map. \cref{fig:parallel:ca} illustrates parallel CA in more detail. To allow information exchange between the non-overlapping windows, we add a feed-forward network including $3{\times}3$ depth-wise convolution.

\PAR{Other architectural features.} Before feeding inputs to HRFuser, we project all secondary modalities onto the image plane of the camera, using perspective projection as proposed in~\cite{perception:aware:fusion:3D:semantic:segmentation}.
This yields an exact spatial correspondence between the input feature maps of different modalities, ensuring consistency among corresponding windows from different modalities in MWCA. All branches start with a CNN reducing the resolution by a factor of 4, followed by 4 stages consisting of multiple identical blocks. For all branches, we use basic bottleneck blocks to build the first stage~\cite{hrnet} and transformer blocks to build all subsequent stages and streams~\cite{hrformer}. We choose the parameters of each MWCA transformer ($H$, $D$) to be equal to the parameters of the subsequent transformer blocks of the respective stream. 
We include additional implementation details on different versions of HRFuser in Appendix~\ref{app:archit:details}.

\section{Experiments}
\label{sec:experiments}

\begin{table}
  \vspace{1.em}
  \caption{Comparison of 2D detection methods on nuScenes evaluated on 6 classes following~\cite{radar:camera:fusion:joint:detection:distance:estimation}. 
  The first group of rows uses the standard nuScenes set, while the second group uses the splits from~\cite{deep:learning:based:radar:camera:fusion:architecture}.
  C: camera, R: radar, L: lidar, (*): results taken directly from the respective paper.}
  \label{table:sota:nus}
  \smallskip
  \resizebox{\columnwidth}{!}{%
  \centering
  \setlength\tabcolsep{3pt}
  \small
  \begin{tabular}{lccccccc}
  \toprule
  Method & Modalities & AP & AP$_{0.5}$ & AP$_{0.75}$ & AP$_m$ & AP$_l$ & AR\\
  \midrule
  HRNetV2p-w18~\cite{hrnet} & C & 32.4 & 56.6 & 33.5  & 21.0 & 43.7 & 43.4\\ 
  HRFormer-T~\cite{hrformer} & C & 34.3 & 59.6 & 35.6  & 23.2 & 45.5 & 43.9\\ 
  HRFormer-B~\cite{hrformer} & C & 33.8 & 59.4 & 34.6 & 22.4 & 45.1 & 43.1 \\ 
  Radar-Camera Fusion\cite{radar:camera:fusion:joint:detection:distance:estimation}* & CR & 35.6 &  60.5 & 37.4 & - & - &42.1\\
  HRFuser-T & CRL & 38.3 & 65.3 & 40.1  & 26.8 & 49.9 & 48.3\\ 
  HRFuser-S & CRL & 38.5 & 65.6 & 40.2  & \best{27.2} & 49.9 & 48.1\\
  HRFuser-B  & CRL & \best{38.8} & \best{66.0} & \best{41.0}  & 26.9 & \best{50.7 }& \best{48.6}\\ 
  \midrule
  \midrule
  CRF-Net~\cite{deep:learning:based:radar:camera:fusion:architecture} & CR & 27.0 &  42.7 & 29.0 & 22.7 & 35.6 & 31.3\\ 
  HRFuser-T & CRL & \best{34.6} & \best{62.0} & \best{34.7} & \best{26.0} & \best{48.5} & \best{45.8}\\
  \bottomrule
  \end{tabular}}
  \vspace{-1.em}
\end{table}

We organize this section as follows. We first present our implementation details and experimental setup for multi-sensor 2D object detection on the two examined datasets, DENSE~\cite{seeing:through:fog} and nuScenes~\cite{nuScenes}. We then compare the 2D performance of our method to the state-of-the-art in 2D and 3D multi-sensor fusion and conduct detailed ablation studies on the utility of including additional sensors and the fusion mechanism.

\subsection{Experimental Setup}
\label{sec:exp:setup}

\begin{table*}[!tb]
  \vspace{1.em}
  \caption{Comparison of 2D detection methods on the DENSE test sets in AP. (*): results taken directly from the respective paper.}
  \label{table:sota:stf}
  \smallskip
  \centering
  \setlength\tabcolsep{4pt}
  \small
  \begin{tabular*}{\linewidth}{l @{\extracolsep{\fill}} cccccccccccc}
  \toprule
  Weather &  \multicolumn{3}{c}{\textbf{clear}} & \multicolumn{3}{c}{\textbf{light fog}} &  \multicolumn{3}{c}{\textbf{dense fog}} & \multicolumn{3}{c}{\textbf{snow/rain}}\\
  Difficulty & easy & mod. & hard & easy & mod. & hard & easy & mod. & hard & easy & mod. & hard \\
   \midrule
  Deep Entropy Fusion~\cite{seeing:through:fog}*~~~~  & 89.84 &	85.57 &	79.46 &	90.54 &	87.99 & 84.90 &	87.68 &	\best{81.49} &	76.69 &	88.99 &	83.71 &	77.85\\
  HRFuser-T & \best{90.15} &	\best{87.10} &	\best{79.48} &	\best{90.60} &	\best{89.34} &	\best{86.50} &	\best{87.93} &	80.27 &\best{	78.21} &	\best{90.05} &	\best{85.35} &	\best{78.09}\\
  \bottomrule
  \end{tabular*}
  \vspace{-1em}
\end{table*}

In all our experiments, we use a two-stage HRFuser network for 2D detection. The backbone of the network is structured as per \cref{sec:method} and its outputs are used to feed a Cascade R-CNN~\cite{cascade:rcnn} head which serves as the second stage of the network. 
We test a tiny (T), small (S) and base (B) version of HRFuser and implement them using the mmdetection framework~\cite{mmdetection}.

HRFuser is trained on DENSE for 60 epochs on batches of size 12 using AdamW with a base learning rate (LR) of 0.001. We apply a 500-step LR warm-up and reduce the LR by a factor of 10 at epochs 40 and 50. The training settings are the same for nuScenes, except that we use 12 epochs, a LR of 0.0001 and LR reductions at epochs 8 and 11. 
To accelerate learning of features 
from the less rich modalities such as radar, we randomly set inputs to zero during training~\cite{deep:learning:based:radar:camera:fusion:architecture,seeing:through:fog} with a chance of 50\% for DENSE and 20\% for nuScenes.

\PAR{DENSE}~\cite{seeing:through:fog} is a multi-modal driving dataset with 106k 2D and 68k 3D bounding boxes. The dataset provides camera images, lidar and radar points, and gated camera images, captured under a variety of normal and adverse weather conditions. The gated camera in DENSE captures images in the NIR band at 808nm with a time-synchronized flood-lit flash laser source.
Following the standard dataset splits in~\cite{seeing:through:fog}, we train only on clear-weather data and use adverse-condition data only for evaluation. We follow~\cite{seeing:through:fog} for basic sensor pre-processing, obtaining $1248{\times}360$ images with depth, intensity and height for lidar, and depth and velocity over ground for radar. Note that radar is missing the RCS channel, since this is not published with the rest of DENSE.
We train on the common KITTI classes car, pedestrian, and cyclist, and evaluate only on car using the KITTI evaluation framework~\cite{kitti}, similar to~\cite{seeing:through:fog}.

\PAR{NuScenes}~\cite{nuScenes} is a large-scale dataset (1.4M images) providing 3D data and annotations of a full autonomous vehicle sensor suite including 6 cameras, 1 lidar and 5 radars. We follow~\cite{deep:learning:based:radar:camera:fusion:architecture} for basic sensor pre-processing, creating radar images with range, radar cross-section (RCS) and velocity over ground, and lidar images with range, intensity and height. Compared to~\cite{deep:learning:based:radar:camera:fusion:architecture}, we do not accumulate radar data across time or filter them in any way. Unless otherwise stated, we use a subset of 10 nuScenes object classes following the mmdet3d~\cite{mmdet3d} framework: car, truck, trailer, bus, construction vehicle, bicycle, motorcycle, pedestrian, traffic cone, and barrier. 
To create 2D ground truth and to evaluate 3D approaches in 2D, we project the 3D bounding boxes onto each image plane by computing a rectangle convex hull of the projected corners, similar to~\cite{deep:learning:based:radar:camera:fusion:architecture,radar:camera:fusion:joint:detection:distance:estimation}. Whereby, we discard annotations that are labeled with the lowest visibility bin, thereby filtering out occluded boxes. 
We train on the official training set and evaluate on the validation set, due to the lack of a public benchmark for 2D detection. Evaluation uses the 2D COCO evaluation metrics~\cite{microsoft:coco}.

\subsection{Comparison to the State of the Art}
\label{subsec:exp:sota}

We compare multiple versions of HRFuser to state-of-the-art camera-only and multi-modal methods on nuScenes in \cref{table:sota:nus}. All versions of HRFuser outperform substantially all camera-only models. In particular, the fully-fledged HRFuser-B improves AP by 5.0\% compared to HRFormer-B and demonstrates analogous improvements in all other metrics. 
Moreover, all versions of HRFuser beat the radar-camera fusion method of~\cite{radar:camera:fusion:joint:detection:distance:estimation} by a large margin on the standard nuScenes split, showcasing the advantage of leveraging multiple complementary sensors---including lidar---with a single, modular architecture as ours over just using radar and camera. A substantial performance gain of +7.6 AP is also observed over CRF-Net~\cite{deep:learning:based:radar:camera:fusion:architecture} on the nuScenes split that is employed by~\cite{deep:learning:based:radar:camera:fusion:architecture} and using only the front camera for evaluation.

In \cref{table:sota:stf}, we compare our HRFuser-T to the fusion method of~\cite{seeing:through:fog} on DENSE. Our model clearly outperforms \cite{seeing:through:fog} across all weather conditions, showing in particular significant improvements in the cases of light fog and dense fog, in which it beats \cite{seeing:through:fog} by 1.6\% and 1.5\% on the ``hard'' setting, respectively. This finding showcases the ability of our model to generalize well to previously unseen, adverse conditions, which degrade the quality of the readings for some of the sensors, such as the camera and the lidar, by properly attending to the features from the sensors that are more robust to these conditions, such as the radar and the gated camera.

In~\cref{table:bev}, we compare our 2D HRFuser to 3D object detection approaches on nuScenes by projecting their 3D results to 2D, as indicated in~\cref{sec:exp:setup}.
HRFuser substantially outperforms the state-of-the-art 3D detection method BEVFusion~\cite{liu2022bevfusion} on all 2D metrics, demonstrating its effectiveness for 2D detection against 3D-based approaches. 
Note that this comparison is reasonably fair as 
correct 3D predictions will still be correct when evaluated against the projected 2D ground truth.

\begin{table}
  \caption
  {Comparison for 2D detection on nuScenes. ('): 3D$\to$2D projection is used to obtain 2D predictions.}
  \label{table:bev}
  \smallskip
  \centering
  \setlength\tabcolsep{4pt}
  \small
  \begin{tabular*}{\linewidth}{l @{\extracolsep{\fill}} cccccc}
  \toprule
  Method & AP & AP$_{0.5}$ & AP$_{0.75}$ & AP$_m$ & AP$_l$ & AR\\
   \midrule
  CenterPoint~\cite{yin2021center}' & 17.9 & 40.9 & 13.1 & 8.1 & 28.2 & 29.8 \\
   BEVFusion~\cite{liu2022bevfusion}'~~
  & 26.0 & 52.6 & 22.2 & 13.6 & 37.9 & 37.1\\
  HRFuser-B & \best{32.9} & \best{58.9} & \best{33.0}  & \best{23.8} & \best{44.1}& \best{43.5}\\
  \bottomrule
  \end{tabular*}
\end{table}


\subsection{Ablation Studies} 
\label{subsubsec:exp:modablation}

\begin{table}
  \caption
  {Ablations of input modalities on nuScenes. Results are in AP. EF: early fusion, C: camera, R: radar, L: lidar.}
  \label{table:nus:modality:ablations}
  \smallskip
  \centering
  \setlength\tabcolsep{4pt}
  \small
  \begin{tabular*}{\linewidth}{l @{\extracolsep{\fill}} cccc}
  \toprule
  Modalities & C & CR & CL & CRL\\
   \midrule
    HRFormer-T (EF) & 26.5 &25.7 & \best{28.2} & 27.7\\
  HRFuser-T (ours) & 26.5 & 27.9 & 31.2 & \best{31.5}\\
  \bottomrule
  \end{tabular*}
  \vspace{-1em}
\end{table}

\begin{table*}
  \vspace{1.em}
  \caption{Ablations of input modalities for HRFuser-T on the DENSE test sets in AP. C: RGB camera, R: radar, L: lidar, G: gated camera.}
  \label{table:stf:modality:ablation}
  \smallskip
  \resizebox{\textwidth}{!}{
  \centering
  \setlength\tabcolsep{3pt} 
  \footnotesize
  \begin{tabular}{ccccccc|ccc|ccc|ccc|ccc}
  \toprule 
   C & L & R & G  &  \multicolumn{3}{c}{\textbf{clear}} & \multicolumn{3}{c}{\textbf{light fog}} &  \multicolumn{3}{c}{\textbf{dense fog}} & \multicolumn{3}{c}{\textbf{snow/rain}}& Flops & Parameters & Inference \\
   &  &  &  & easy & mod. & hard & easy & mod. & hard & easy & mod. & hard & easy & mod. & hard & $[GFLOPs]$ &  $[M]$ & $[ms]$\\
   \midrule
  \yes & \no & \no & \no & 79.81 &	62.48 &	53.68 &	80.84 &	63.07 &	62.08 &	71.84 &	62.69 &	54.05 &	78.68 &	61.19 &	52.72 & 104.0 & 47.9 & 81.1\\
   \midrule
  \yes &  \yes & \no & \no & 89.91 &	85.16 &	78.68 &	90.47 &	88.44 &	80.55 &	87.39 &	78.32 &	71.13 &	89.21 &	79.88 &	76.19 & 114.1 (+9.7\%)& 48.8 (+1.9\%)& 103.3 (+27.4\%)\\
  \yes &  \no & \yes & \no & 88.48	&80.15	&76.25	&90.37	&86.40	&79.6	&88.51	&79.72	&71.87	&88.13	&78.85	&70.27 & 114.0 (+9.6\%) & 48.8 (+1.9\%) & 103.2 (+27.3\%)\\
  \yes &  \no & \no & \yes & 89.76	&85.37	&78.36	&90.56	&88.04	&80.47	&88.67	&80.64	&72.25	&89.62	&80.14	&76.58 & 114.0 (+9.6\%) & 48.8 (+1.9\%) & 103.0 (+27.0\%)   \\
  \yes &  \yes &\yes & \no & 89.88 &	85.17 &	78.64 &	90.46 &	87.87 &	80.51 &	88.10 &	80.11 &	72.01 &	89.40 &	80.02 &	76.11& 123.3 (+18.6\%)& 49.4 (+3.1\%)& 120.8 (+49.0\%)\\
  \yes &  \yes &\no & \yes & 90.14 & \best{87.18} & 79.44 & 90.62 & 89.17 & 80.95 & 88.56 & 80.33 & 72.21 & \best{90.09} & 85.32 & \best{78.09}& 123.2 (+18.5\%)& 49.4 (+3.1\%)& 121.0 (+49.2\%)\\
  \yes &  \no & \yes & \yes & 89.87 & 85.13 & 78.55 & \best{90.64} & 88.37 & 80.52 & \best{88.97} & \best{80.86} & \best{78.64} & 89.85 & 80.33 & 76.54& 123.2 (+18.5\%)& 49.4 (+3.1\%)& 121.3 (+49.6\%)\\
  \yes &  \yes & \yes & \yes & \best{90.15} &	87.10 &	\best{79.48} &	90.60 &	\best{89.34} &	\best{86.50} &	87.93 &	80.27 &	78.21 &	90.05 &	\best{85.35} &	\best{78.09}& 132.4 (+27.3\%)& 49.9 (+4.2\%)& 141.0 (+73.9\%)\\
  \bottomrule
  \end{tabular}
  }
  \vspace{-1.em}
\end{table*}

\PAR{Modalities.} \cref{table:nus:modality:ablations} investigates the contribution of each sensor on nuScenes, by
training HRFuser and a naive early fusion baseline with different subsets of input modalities.
HRFormer-T (Early Fusion)---which naively utilizes a concatenated input without any additional changes to HRFormer---performs 1.7\% better when adding lidar to the camera-only baseline. Note that the performance drops both times when we add the noisy radar to the input modalities.
In contrast, adding radar to HRFuser yields an improvement of 1.4\% over the camera-only baseline. 
The improvement is larger (4.7\%) when adding lidar, and is maximized (5.0\%) when combining all 3 sensors, showing the ability of our MWCA fusion to attend to the useful part of extra modalities---notably radar---while ignoring noisy content in them. 
HRFuser not only avoids a performance drop when adding radar but even gains additional performance. This result implies that the proposed method successfully pays attention to the relevant features.

We examine the effect of different modalities on DENSE in \cref{table:stf:modality:ablation}. A combination of all four modalities yields the overall best performance, except for the case of dense fog, where a combination of camera, radar and gated camera performs best. This is in line with the findings of~\cite{seeing:through:fog} and is due to the severe impact of fog on the lidar,
as the laser pulse has to travel to the object and back, which squares the attenuation due to the presence of fog. 
By contrast, radar and gated cameras are more robust to fog. Note that the used standard splits of DENSE investigate the generalization capabilities rather than the robustness of a model since training includes only clear-weather data. Thus, the effect of fog on lidar is unseen during training, and the network cannot learn how to deal with the introduced noise, as it does with the radar noise on nuScenes in the previous paragraph. Another finding is that adding the gated camera on top of lidar and radar provides a consistent improvement across conditions, evidencing the informativeness of the high-resolution features from this sensor, which is generally robust to adverse conditions.
Furthermore, we experimented with different sensors as primary modalities and found that choosing the information-dense RGB or gated cameras performed better than the sparse lidar and radar sensors. The higher spatial resolution may aid in guiding the fusion and attending to smaller details. For further details on the choice of primary modality, we refer the reader to Appendix~\ref{app:additional:ablations}.

\begin{table}
  \caption{Ablations of fusion strategies on nuScenes.
  }
  \label{table:architecture:ablations}
  \smallskip
  \resizebox{\columnwidth}{!}{%
  \centering
  \setlength\tabcolsep{3pt}
  \small
  \begin{tabular}{lcccccc}
  \toprule
  Method (Fusion Type) & AP & AP$_{0.5}$ & AP$_{0.75}$ & AP$_m$ & AP$_l$ & AR\\
   \midrule
  HRFormer-T & 26.5 & 49.9 & 25.3 & 18.2 & 37.0 & 26.8\\
  HRFormer-T (Early) & 27.7 & 51.6 & 26.5 & 18.4 & 38.8 & 38.9\\
  HRFuser-T (Addition) & 30.8 & 56.4 & 30.5  & 22.0 & 41.9 & 42.0\\
  HRFuser-T (MWCA$_{\text{onlyHighRes}}$) & 30.5 & 56.1 & 29.7 & 21.8 & 41.4 & 41.5 \\
  HRFuser-T (MWCA) & \best{31.5} & \best{57.4} & \best{31.1} & \best{22.7} & \best{42.5} &\best{42.3}\\
   \midrule
  HRFuser-T (PVTv2-CA~\cite{PVTv2:vision:transformer}) & 29.8 & 54.3 & 29.4 & 20.1 & 41.3 & 40.9\\
  HRFuser-T (PVTv2-Li-CA~\cite{PVTv2:vision:transformer}) & 29.5 & 54.2 & 28.6 & 19.9 & 41.0 & 40.6 \\
  \bottomrule
  \end{tabular}}
  \vspace{-1em}
\end{table}

\PAR{Fusion mechanism.} \cref{table:architecture:ablations} presents an ablation study on nuScenes regarding the fusion mechanism which is used in HRFuser, in order to verify the benefit of our MWCA fusion block. The reference is the camera-only HRFormer baseline. 
Early fusion achieves only a slight 1.2\% improvement in AP over the camera-only HRFormer. Using our proposed HRFuser with its multi-resolution fusion design, but with a simplified addition-based fusion block instead of MWCA, already yields a 
large 4.3\% improvement in AP over the camera-only baseline. Replacing addition with our proposed MWCA further improves performance consistently across all metrics, showcasing the utility of attention-based fusion for detection.
Limiting the fusion to only the high-resolution stream of the camera branch yields a 1.0\% reduction in AP, highlighting the importance of multi-resolution fusion.
We compare our MWCA to an alternative attention mechanism via the state-of-the-art transformer PVTv2~\cite{PVTv2:vision:transformer}, adapted for cross-attention (PVTv2-CA). 
For implementation details, we refer the reader to Appendix~\ref{app:additional:ablations}. 
Our MWCA fusion outperforms PVTv2-CA and the linear version PVTv2-Li-CA by 1.7\% and 2.0\% respectively, demonstrating the advantage MWCA.

\begin{figure*}
  \vspace{.5em}
\centering
\begin{tabular}{@{}c@{\hspace{0.05cm}}c@{\hspace{0.05cm}}c@{}}
\multicolumn{1}{c}{\footnotesize Ground Truth} &
\multicolumn{1}{c}{\footnotesize HRFormer~\cite{hrformer}} &
\multicolumn{1}{c}{\footnotesize HRFuser} \\
\vspace{-0.07cm}
\includegraphics[width=0.328\textwidth]{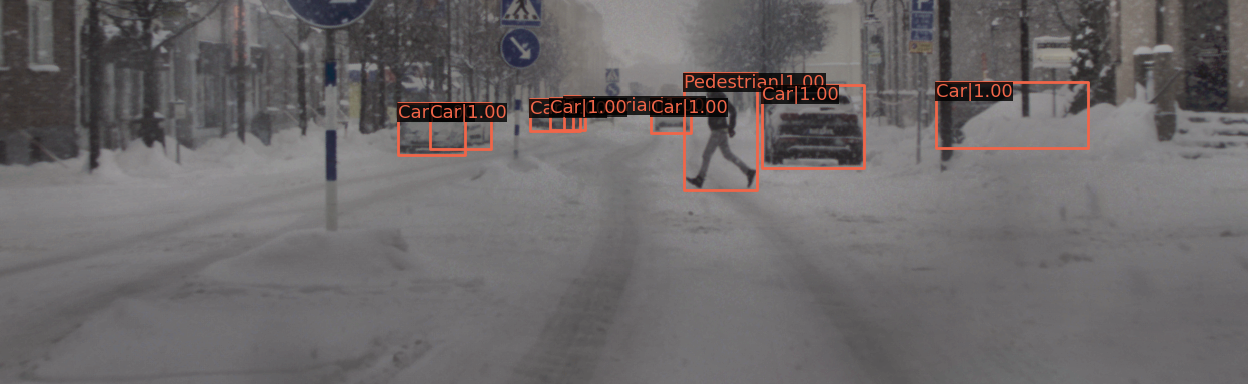} &
\includegraphics[width=0.328\textwidth]{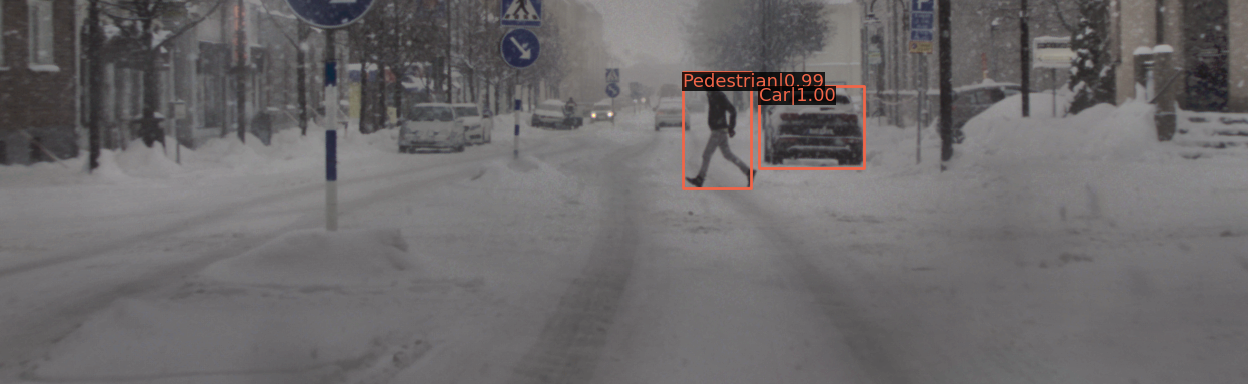} &
\includegraphics[width=0.328\textwidth]{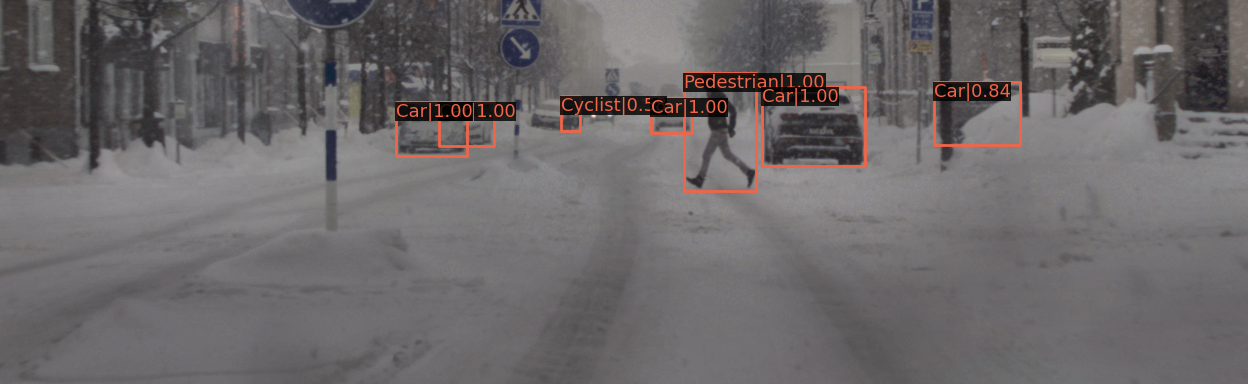} \\
\vspace{-0.07cm}
\includegraphics[width=0.328\textwidth]{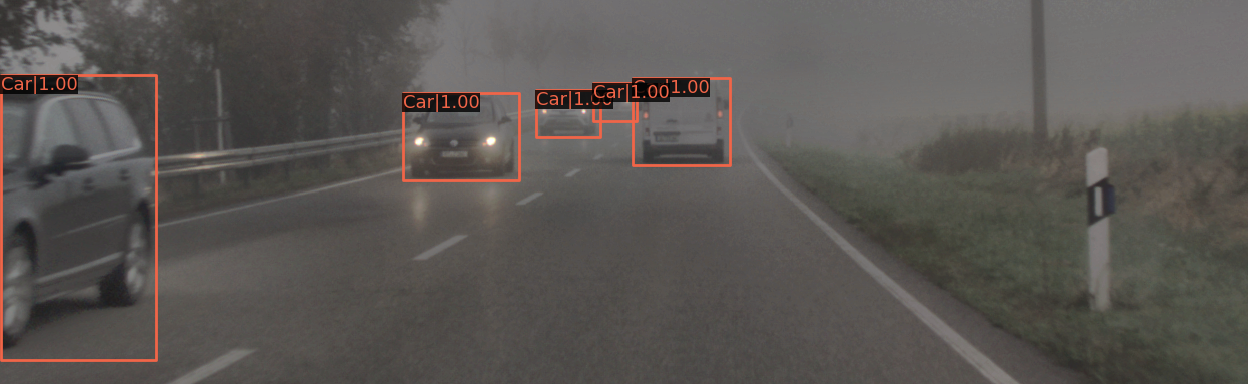} &
\includegraphics[width=0.328\textwidth]{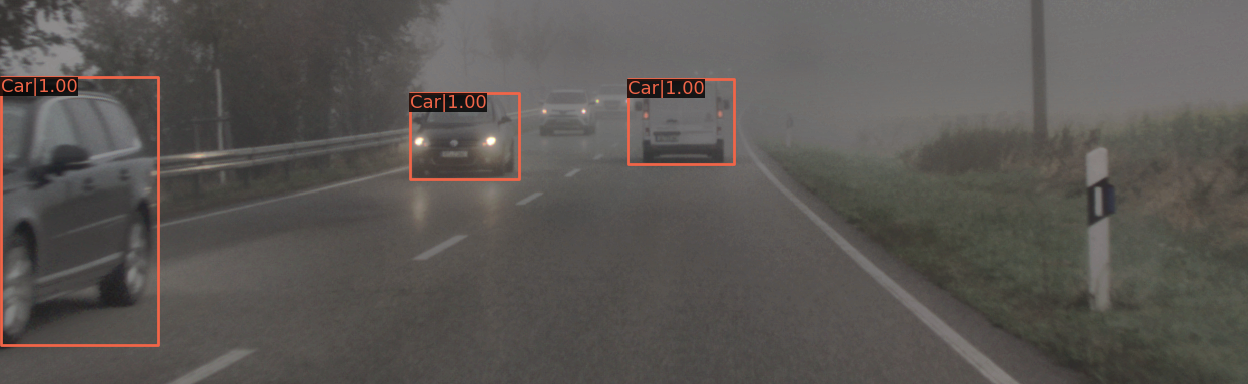} &
\includegraphics[width=0.328\textwidth]{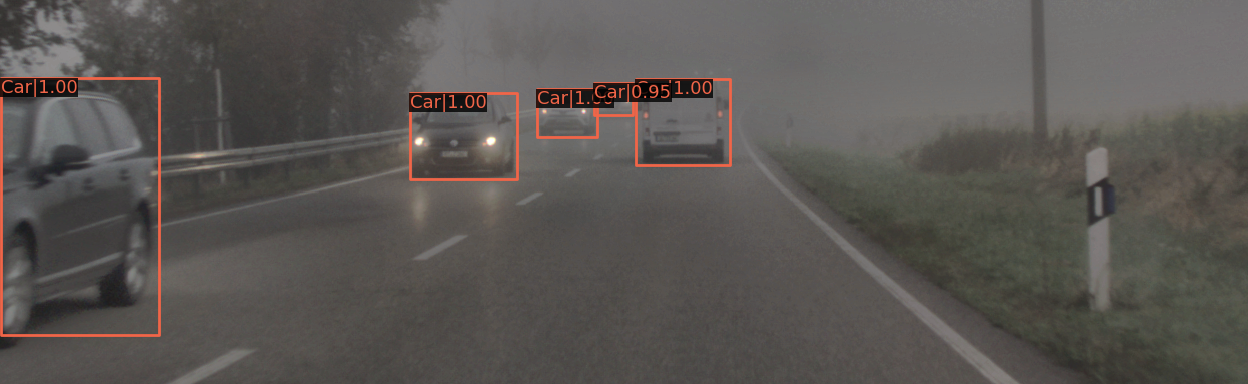} \\
\end{tabular}
\caption{Qualitative detection results on DENSE. Best viewed on a screen at full zoom.}
\label{fig:results:stf}
\vspace{-1.3em}
\end{figure*}

\begin{figure*}
\centering
\begin{tabular}{@{}c@{\hspace{0.05cm}}c@{\hspace{0.05cm}}c@{\hspace{0.05cm}}c@{}}
\subfloat{\footnotesize Ground Truth} &
\subfloat{\footnotesize HRFormer~\cite{hrformer}} &
\subfloat{\footnotesize 3D$\to$2D BEVFusion~\cite{liu2022bevfusion}}&
\subfloat{\footnotesize HRFuser} \\
\vspace{-0.07cm}
\includegraphics[width=0.245\textwidth]{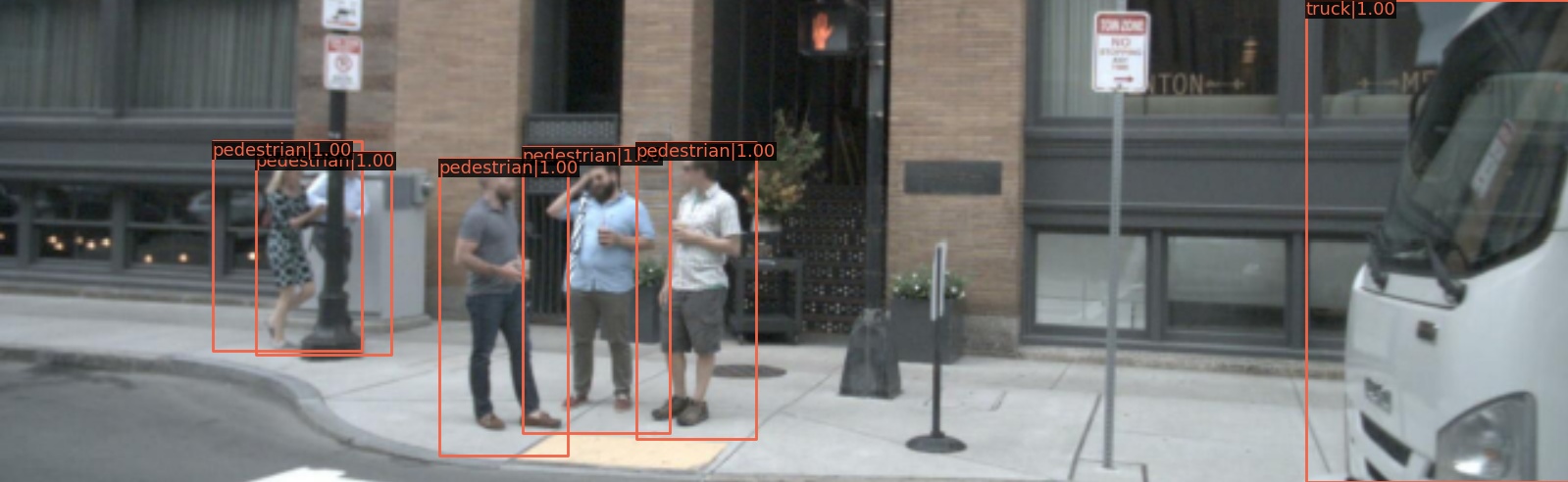} &
\includegraphics[width=0.245\textwidth]{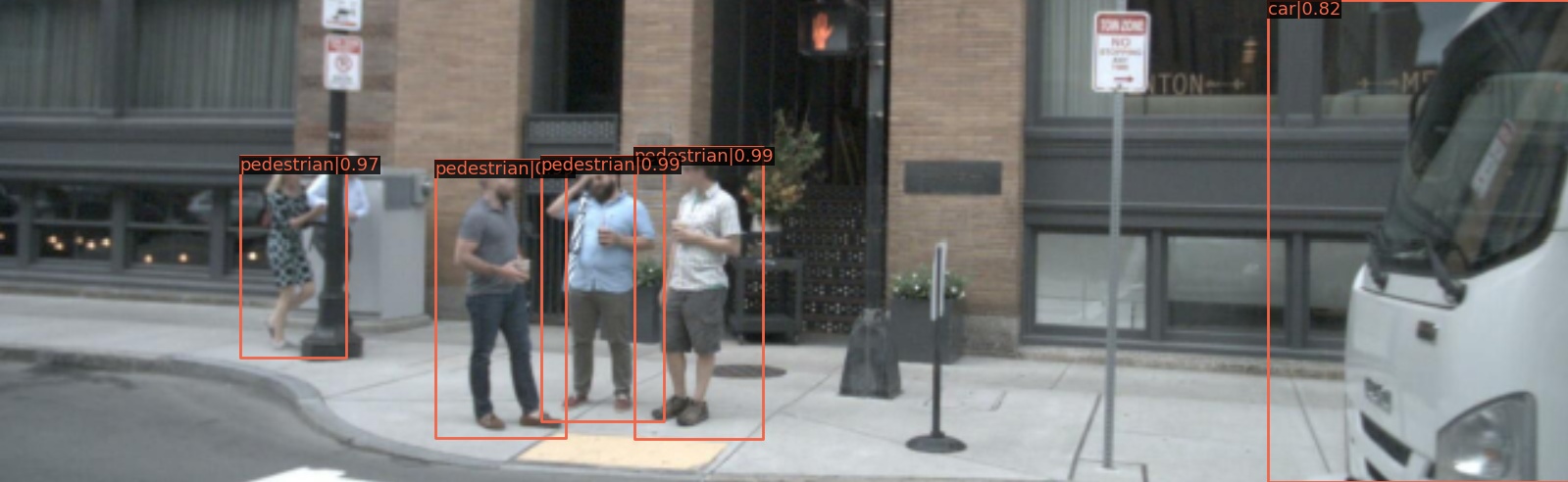} &
\includegraphics[clip,width=0.245\textwidth,trim=1 108 0 300]{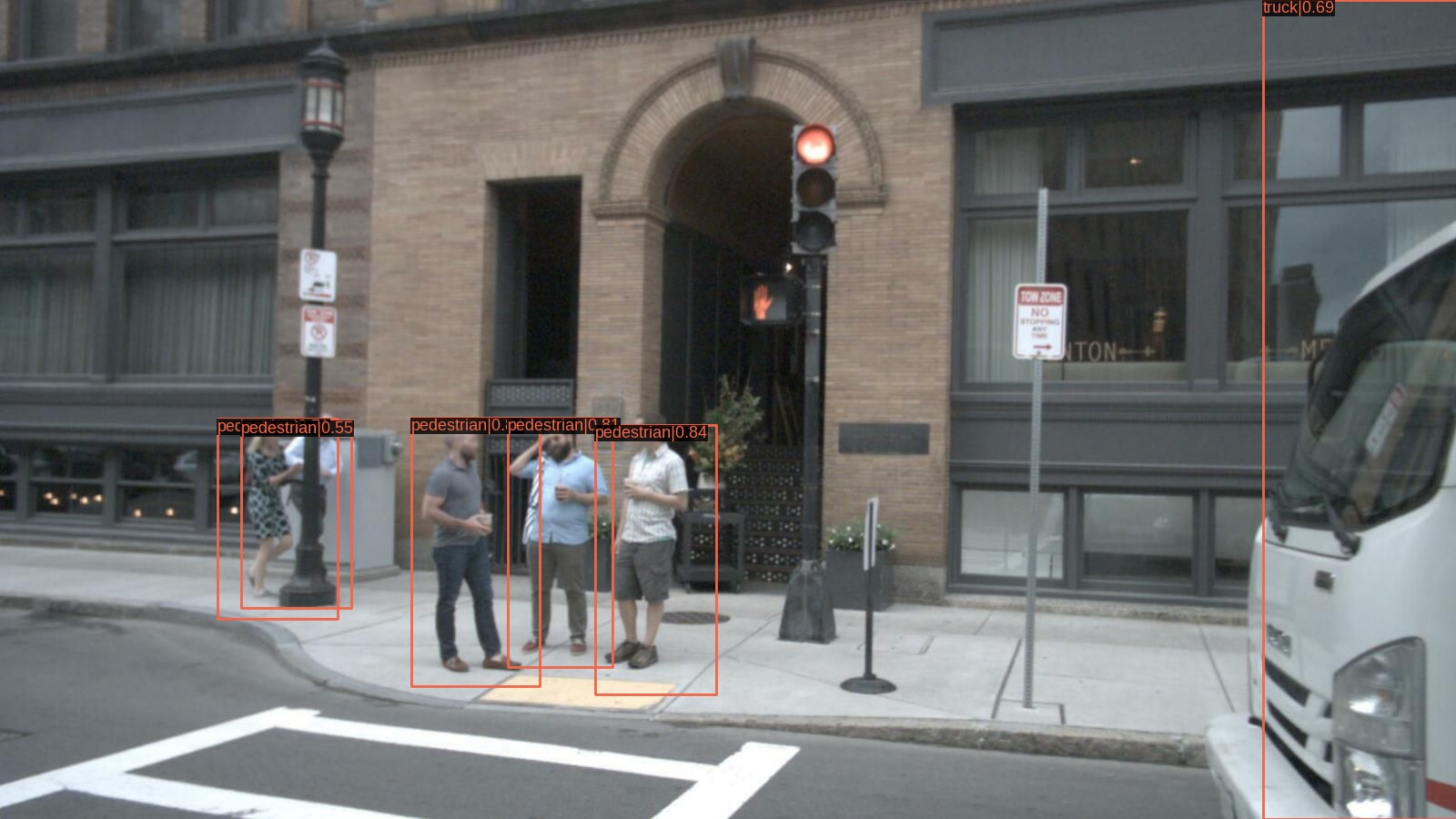} &
\includegraphics[width=0.245\textwidth]{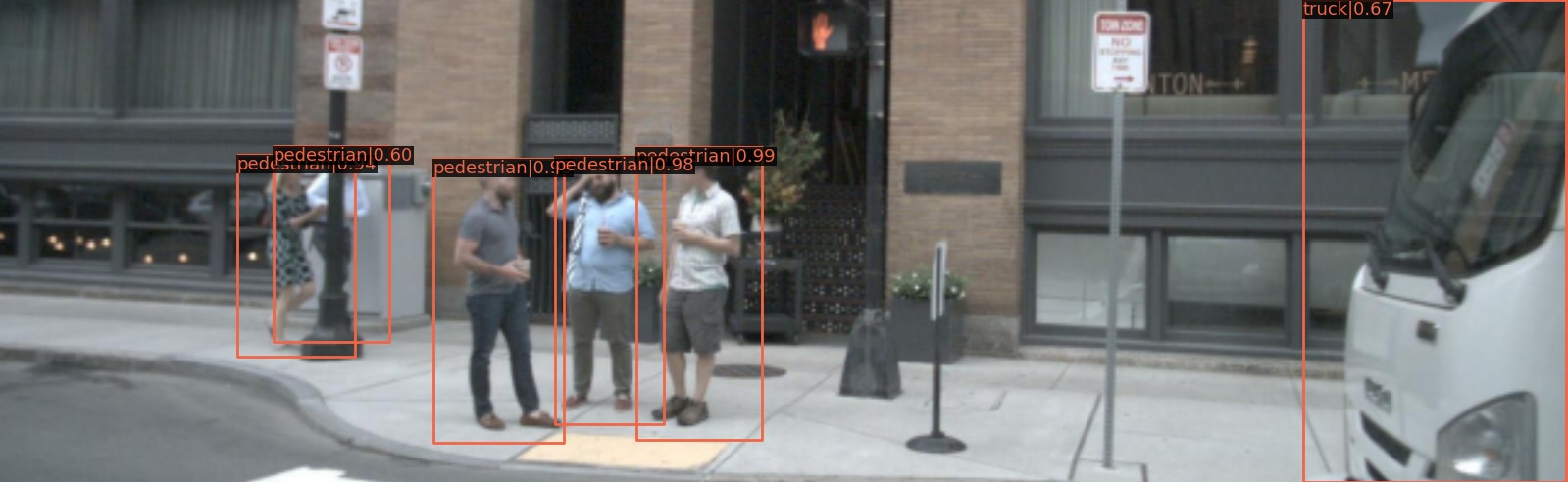} \\

\includegraphics[width=0.245\textwidth]{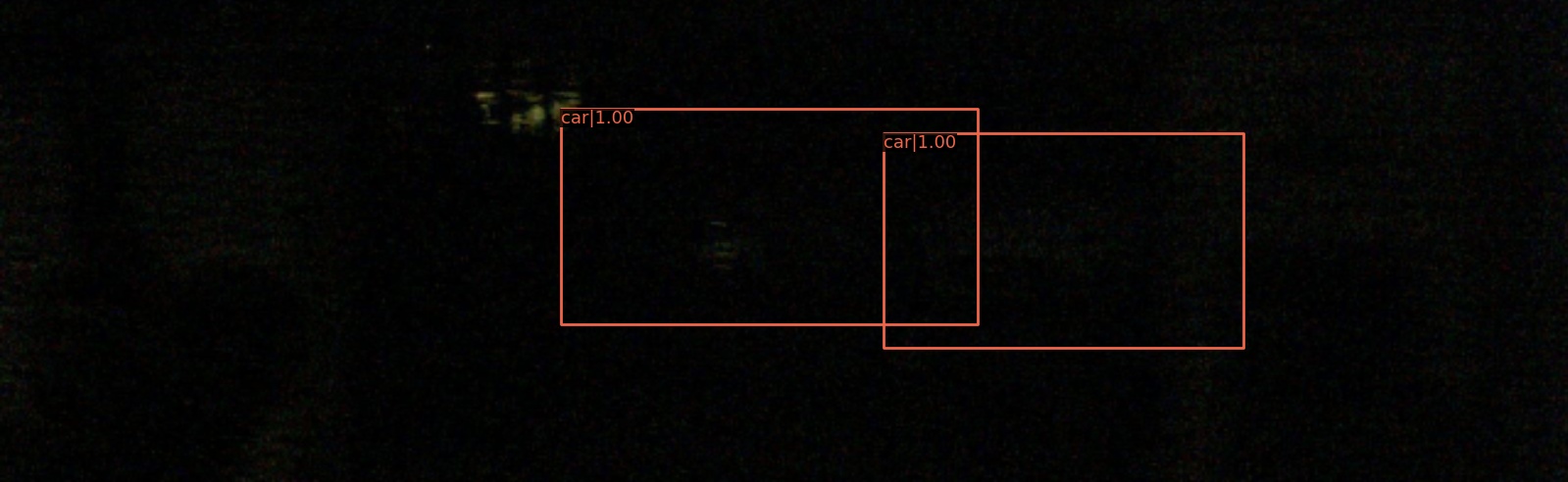} &
\includegraphics[width=0.245\textwidth]{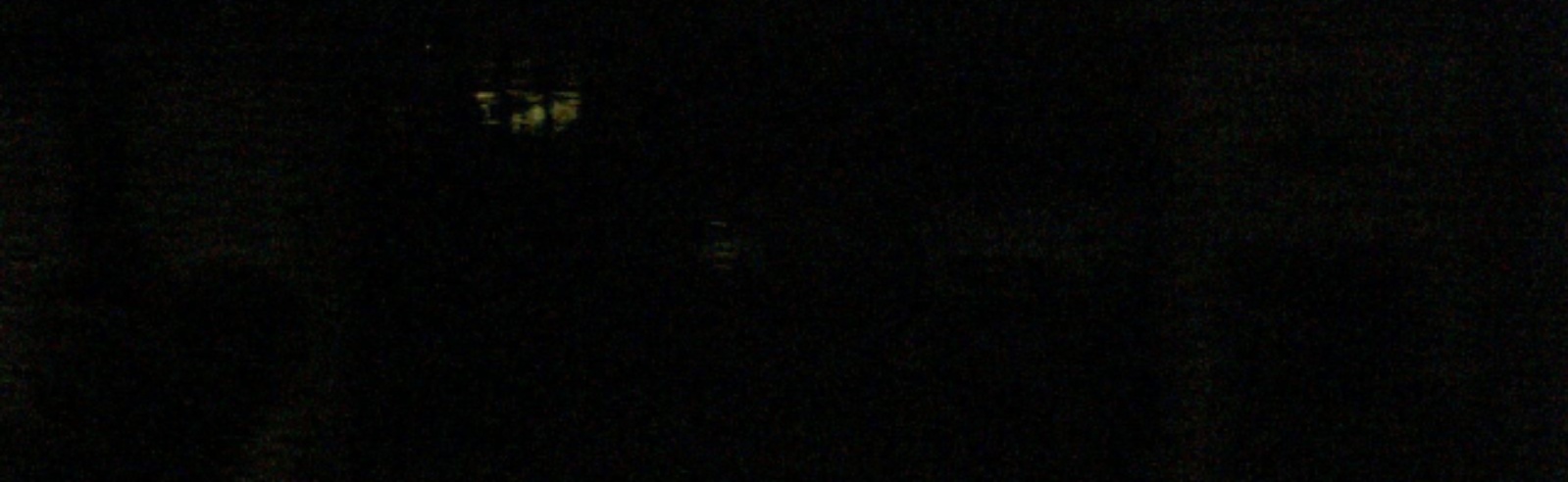} &
\includegraphics[clip,width=0.245\textwidth,trim=1 108 0 300]{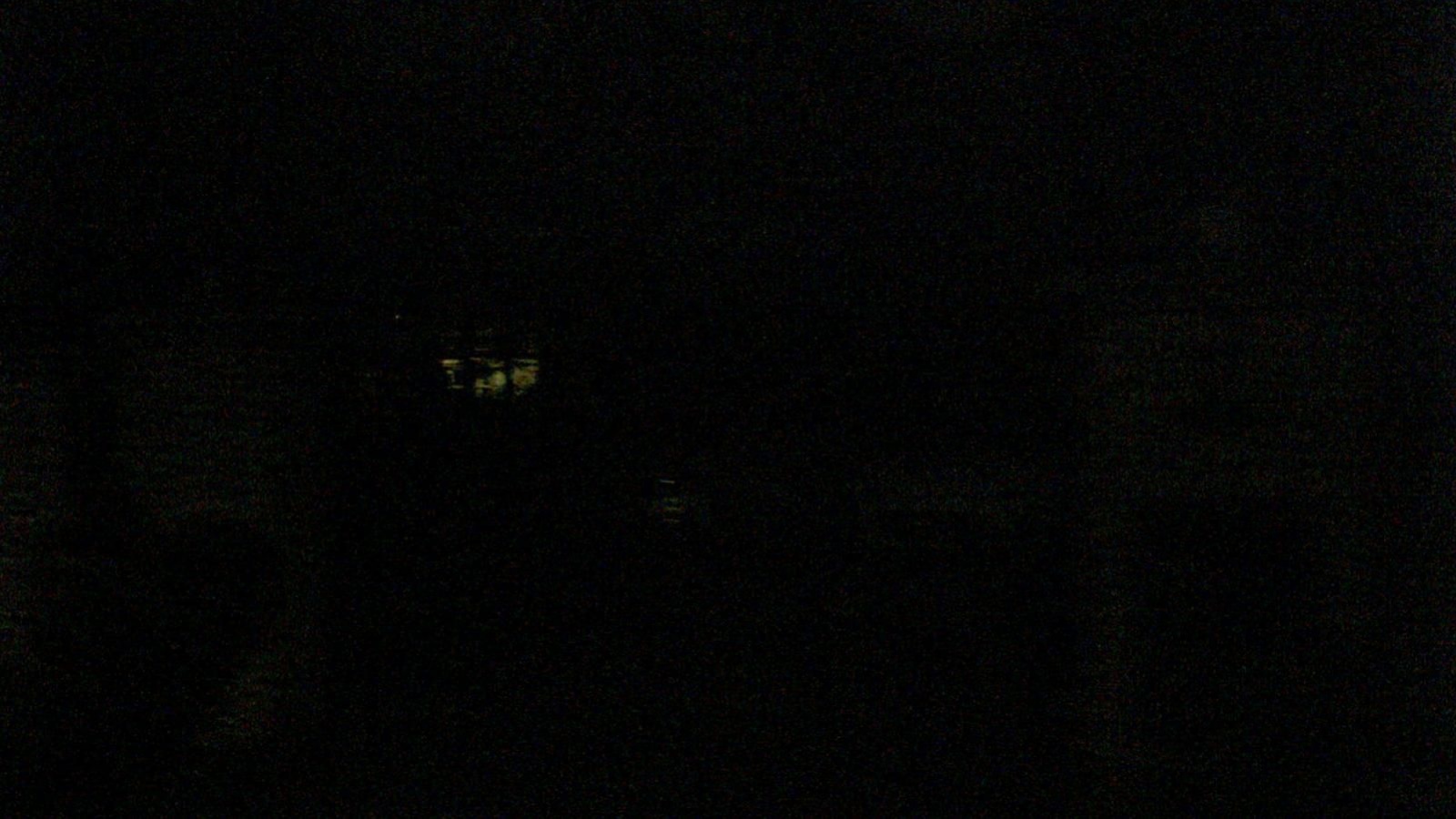} &
\includegraphics[width=0.245\textwidth]{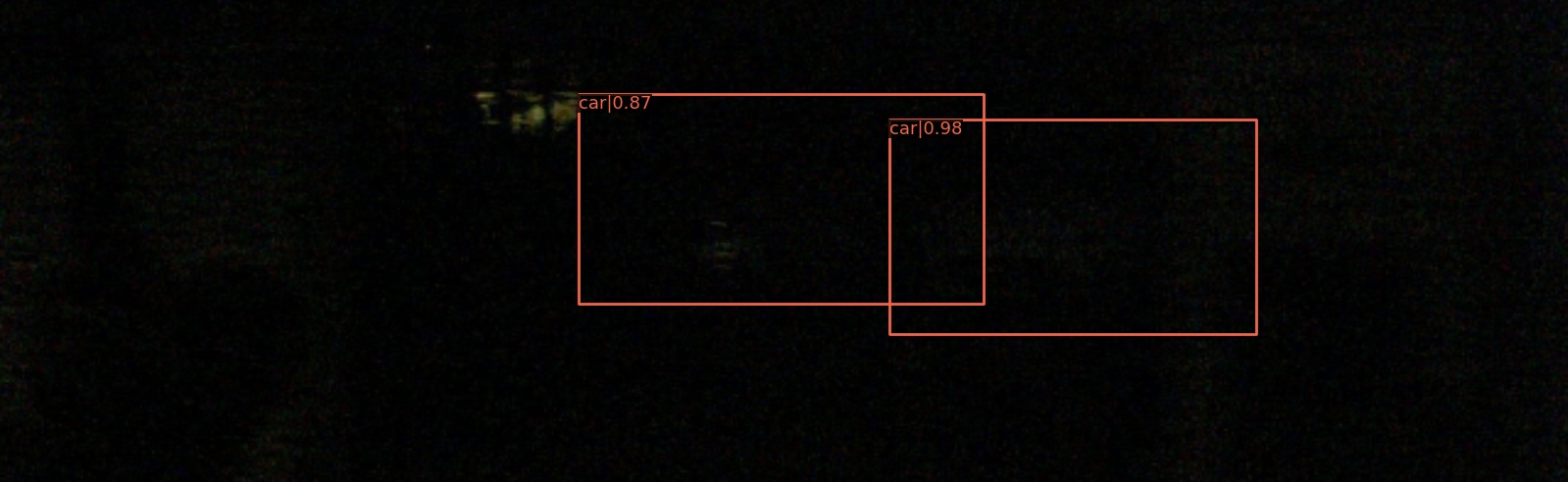} \\
\end{tabular}
\caption{Qualitative detection results on nuScenes. Best viewed on a screen at full zoom.}
\label{fig:results:nuscenes}
\vspace{-1.2em}
\end{figure*}

\PAR{Efficiency.}
We further investigate the number of parameters/flops and the inference speed on an Nvidia Quadro RTX 6000 GPU in \cref{table:stf:modality:ablation}. Our fusion method adds only a minor computational overhead. Even when using \emph{all three} additional modalities besides the camera, the flops increase by only 27.3\% and the parameter count by a marginal 4.2\%. The inference time increases by 27.4\% for one added modality and the full multi-modal network predicts in much less than double the time of the camera-only network.

\subsection{Qualitative Results}

The qualitative results on DENSE in \cref{fig:results:stf} demonstrate that our proposed method is significantly more resilient to adverse conditions
than a strong camera-only model. 
Note e.g.\ the second example, where HRFuser correctly detects obscured cars in the fog, while HRFormer misses them. Even though HRFuser misses a few distant objects in the other example, it still performs significantly better than HRFormer.
HRFormer struggles particularly in detecting objects at a large distance. This can be attributed to the cumulative effect of atmospheric phenomena such as fog and snow on the appearance of objects as their distance from the camera increases. The good performance of HRFuser demonstrates its greater generalization capability thanks to learning robust features from multiple modalities.

\cref{fig:results:nuscenes} presents detection results on nuScenes of HRFormer, BEVFusion and HRFuser. HRFuser detects the partially occluded pedestrian in the first example, which is missed by HRFormer. The last example includes minimal queues from the camera. However, in contrast to the camera-only HRFormer and the lidar- and camera-based BEVFusion, HRFuser correctly detects both cars of the scene, showcasing its ability to effectively leverage complementary sensors for object detection.

\section{Conclusion}
\label{sec:conclusion}

We have proposed HRFuser, a multi-modal, multi-resolution and multi-level fusion architecture. In particular, we have extended the high-resolution paradigm for dense semantic prediction to multiple modalities by introducing additional high-resolution branches for the extra modalities besides the camera. HRFuser repeatedly fuses the extra modalities into the multi-resolution camera branch with a novel transformer block that applies cross attention in local windows and enables efficient learning of robust multi-modal features.
Our proposed MWCA fusion module attends to discriminative information from additional sensors while ignoring their noisy parts.
We have evaluated HRFuser on DENSE and nuScenes and demonstrated its state-of-the-art performance in 2D object detection across a wide range of scenes and conditions. Our architecture is generic and scales straightforwardly to an arbitrary number of sensors, thus being of particular relevance for practical multi-modal settings in autonomous cars and robots, which usually involve a diverse set of sensors.

\section*{ACKNOWLEDGMENT}

This work was supported by the ETH Future Computing Laboratory (EFCL), financed by a donation from Huawei Technologies.

{\small
\bibliographystyle{IEEEtran}
\bibliography{IEEEabrv,main}
}

\appendix

\subsection{Additional Architectural Details}
\label{app:archit:details}

All branches of HRFuser start with a CNN reducing the resolution by a factor of four, followed by four stages consisting of multiple identical blocks. For all branches, we use basic bottleneck blocks to build the first stage~\cite{hrnet} 
and transformer blocks to build all subsequent stages and streams~\cite{hrformer}. 
A transformer block consists of a local-window self-attention on $7\times7$ windows followed by an feed-forward network with $3\times3$ depth-wise convolution and an expansion ratio of 4. The additional parameters of the transformer blocks for different versions of HRFuser are displayed in \cref{table:parameters}, where $D_s$ and $H_s$ apply to all blocks and MWCA modules within a given stream.

\subsection{Overview of Characteristics of Different Sensors}

We provide a more detailed overview of the characteristics of each sensor we use in our experiments:

\begin{enumerate}
    \item \textbf{Camera:} Very high resolution and rich texture, but poor readings in low illumination and fog and no direct geometric information.
    \item \textbf{Lidar:} Fair resolution, explicit range information, independent from external illumination, degradation when the optical medium is not clear (fog, rain, snow).
    \item \textbf{Radar:} Robust to adverse weather and illumination, velocity information, low resolution, noisy.
    \item \textbf{Gated camera:} High resolution, robust to adverse weather and illumination, still not widely adopted.
\end{enumerate}

\begin{table}[ht]
  \caption{Parameters of the tiny (T), small (S), and base (B) versions of HRFuser. $D_s$ denotes the number of channels and $H_s$ the number of heads, with $s\in\{1,\,\dots,\,4\}$ denoting the corresponding stream. For the camera branch $\alpha$, the values are displayed as: $\left(D_1, D_2, D_3, D_4\right)$ and $\left(H_1, H_2, H_3, H_4\right)$. The secondary branches $\beta$ only have one stream: $D_1$ and $H_1$.
  }
  \label{table:parameters}
  \centering
  \setlength\tabcolsep{2pt}
  \small
  \begin{tabular*}{\linewidth}{l @{\extracolsep{\fill}} cccc}
  \toprule
  Model & Branch & \#channels ($D_s$) & \#heads ($H_s$)\\
  \midrule
  HRFuser-T & $\alpha$ & $\left(18,36,72,144\right)$   & $\left(1,2,4,8\right)$\\
   & $\beta$& $18$   & $1$\\
  HRFuser-S & $\alpha$ & $\left(32,64,128,256\right)$  & $\left(1,2,4,8\right)$\\
   & $\beta$& $32$   & $1$\\
  HRFuser-B & $\alpha$ & $\left(78,156,312,624\right)$ & $\left(2,4,8,16\right)$\\
   & $\beta$& $78$   & $2$\\
  \bottomrule
  \end{tabular*}
\end{table}

\subsection{Additional Details on the Experimental Setup}

For both datasets (DENSE and nuScenes), we use AdamW
with a base learning rate of 0.001, 
weight decay of 0.01, and betas of 0.9 and 0.999. We apply a learning rate warm-up for 500 iterations with a ratio of 0.001.

\begin{figure*}
    \centering
    \begin{tabular}{@{}c@{\hspace{0.05cm}}c@{\hspace{0.05cm}}c@{\hspace{0.05cm}}c@{}}    
        \subfloat{\footnotesize RGB Camera} &
        \subfloat{\footnotesize Gated Camera} &
        \subfloat{\footnotesize Lidar} &
        \subfloat{\footnotesize Radar} \\
        \vspace{-0.07cm}
    \includegraphics[clip,trim=250 115 250 77, width=0.248\textwidth]{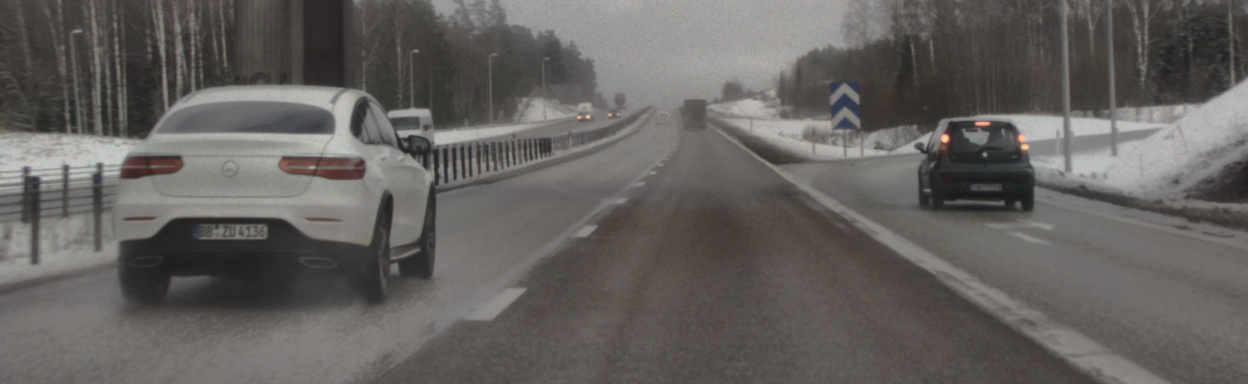} &
    \includegraphics[clip,trim=250 115 250 77, width=0.248\textwidth]{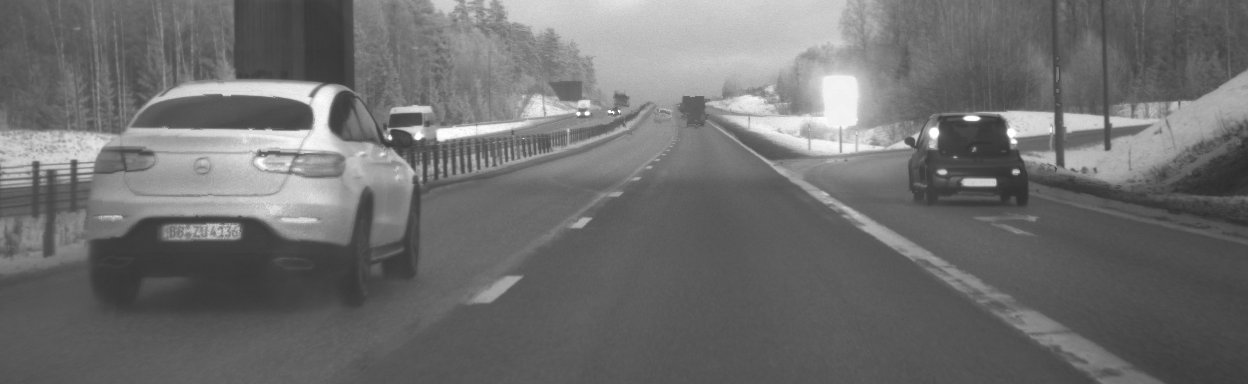} &
    \includegraphics[clip,trim=250 115 250 77, width=0.248\textwidth]{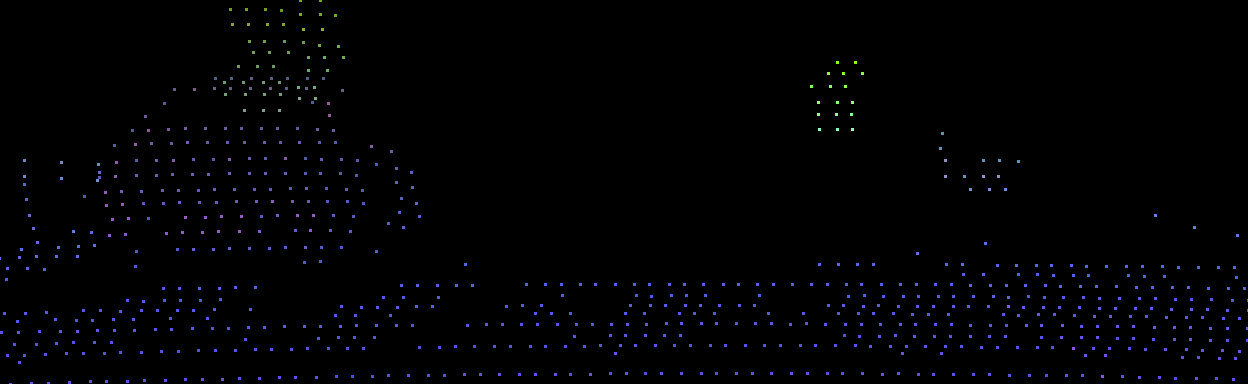} &
    \includegraphics[clip,trim=250 115 250 77, width=0.248\textwidth]{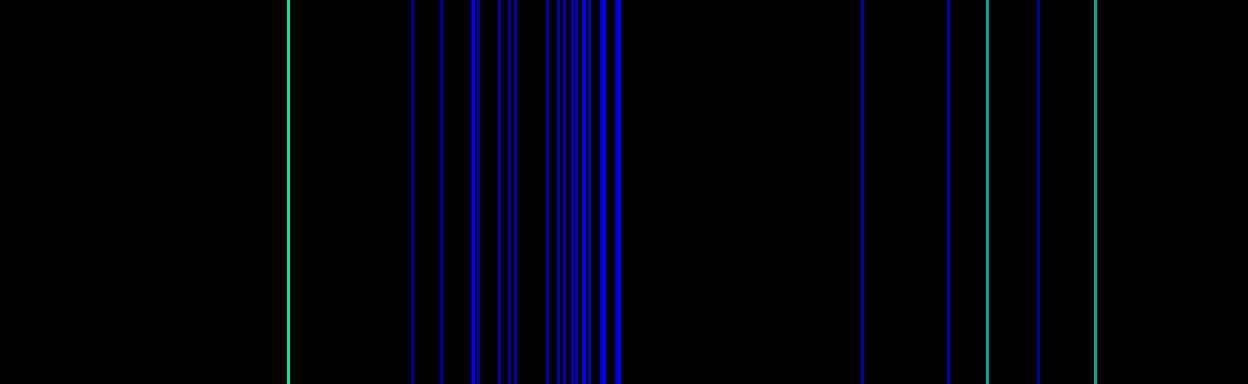} \\
    \vspace{-0.07cm}    
    \includegraphics[clip,trim=250 115 250 77, width=0.248\textwidth]{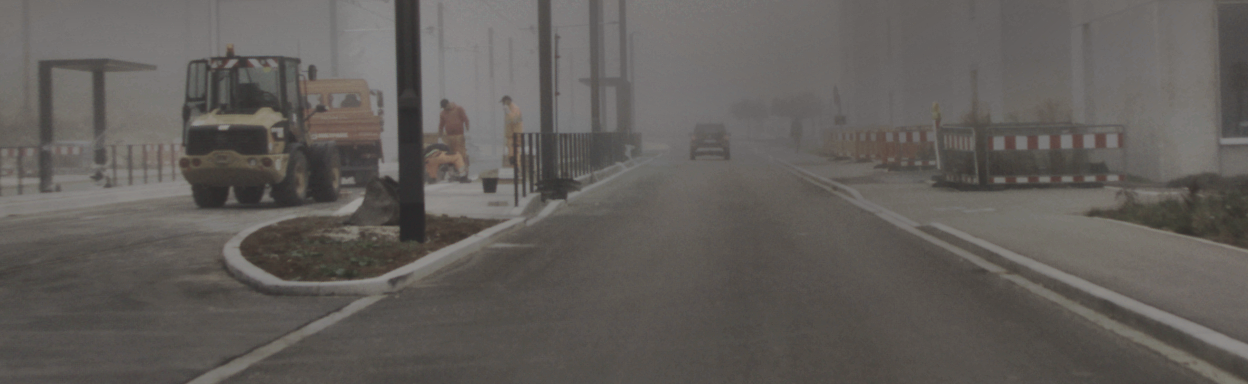} &
    \includegraphics[clip,trim=250 115 250 77, width=0.248\textwidth]{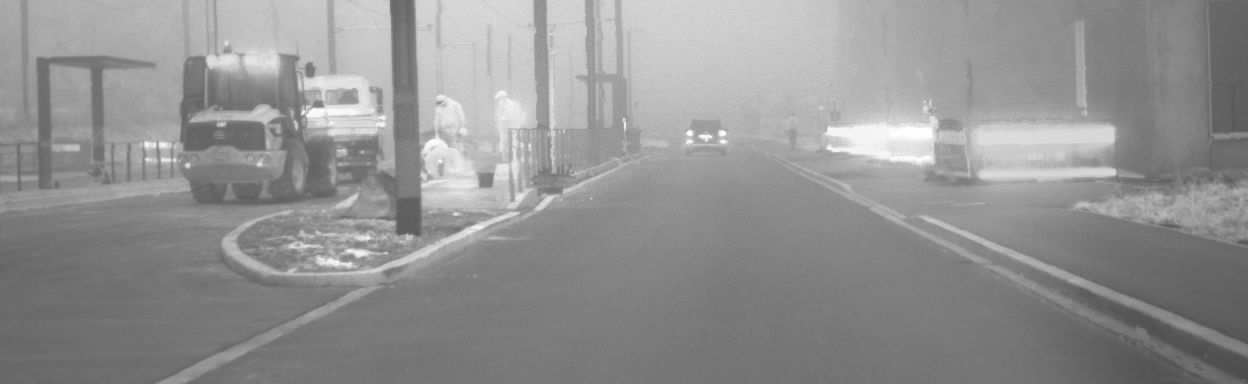} &
    \includegraphics[clip,trim=250 115 250 77, width=0.248\textwidth]{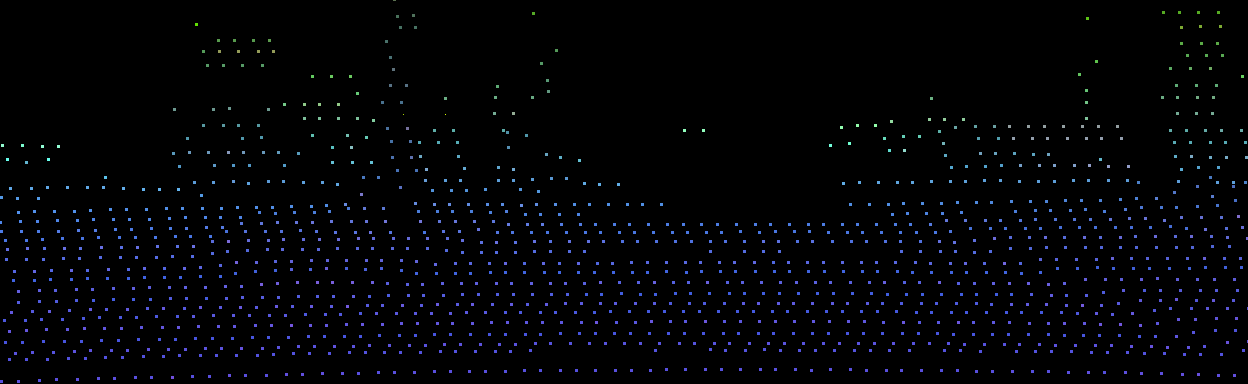} &
    \includegraphics[clip,trim=250 115 250 77, width=0.248\textwidth]{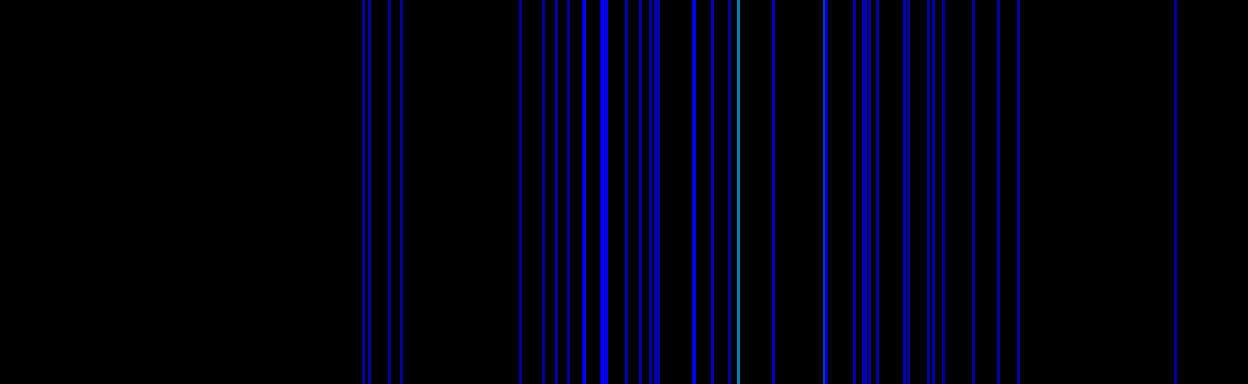} \\
    
    \includegraphics[clip,trim=250 115 250 77, width=0.248\textwidth]{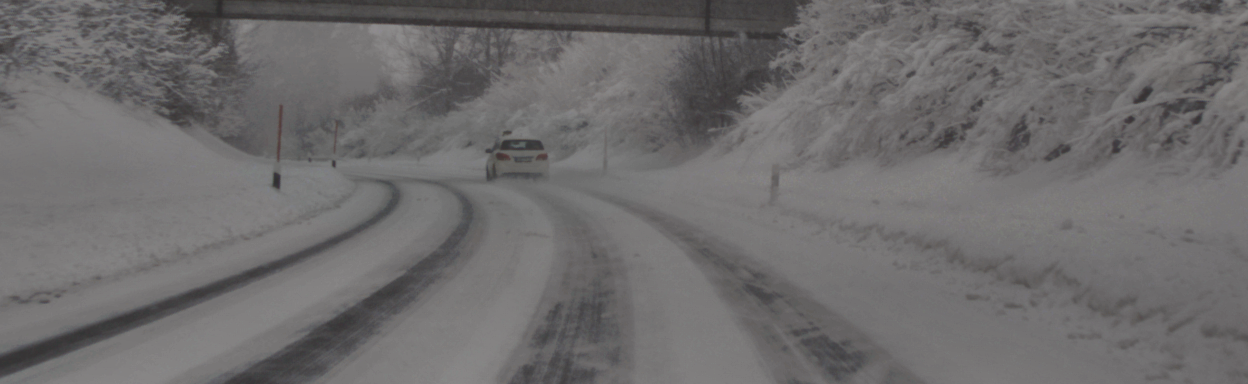} &
    \includegraphics[clip,trim=250 115 250 77, width=0.248\textwidth]{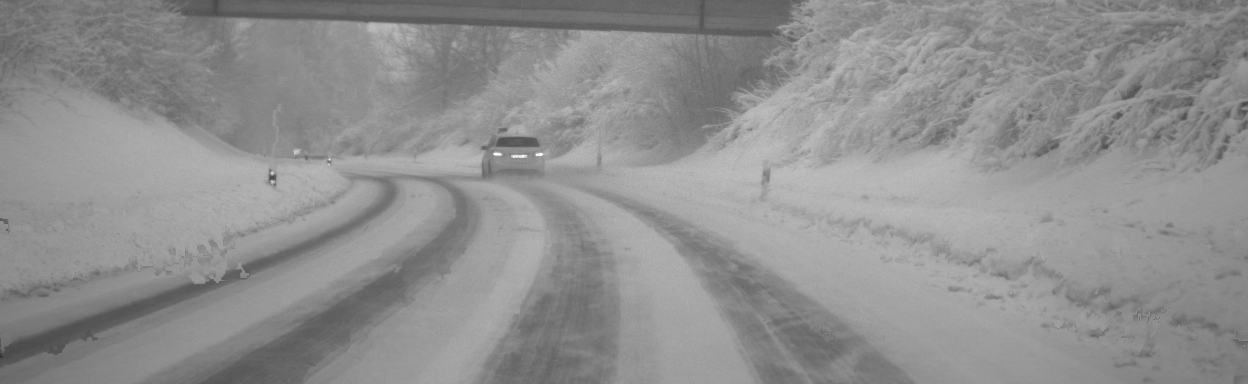} &
    \includegraphics[clip,trim=250 115 250 77, width=0.248\textwidth]{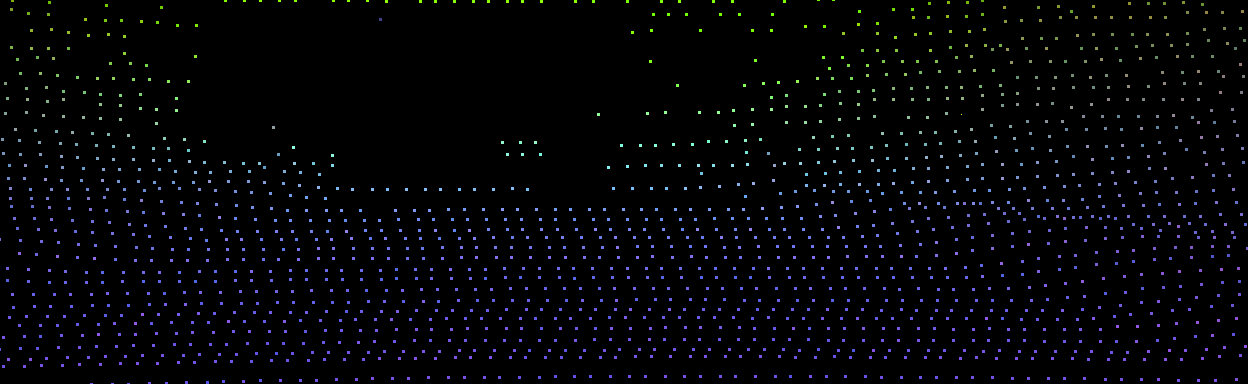} &
    \includegraphics[clip,trim=250 115 250 77, width=0.248\textwidth]{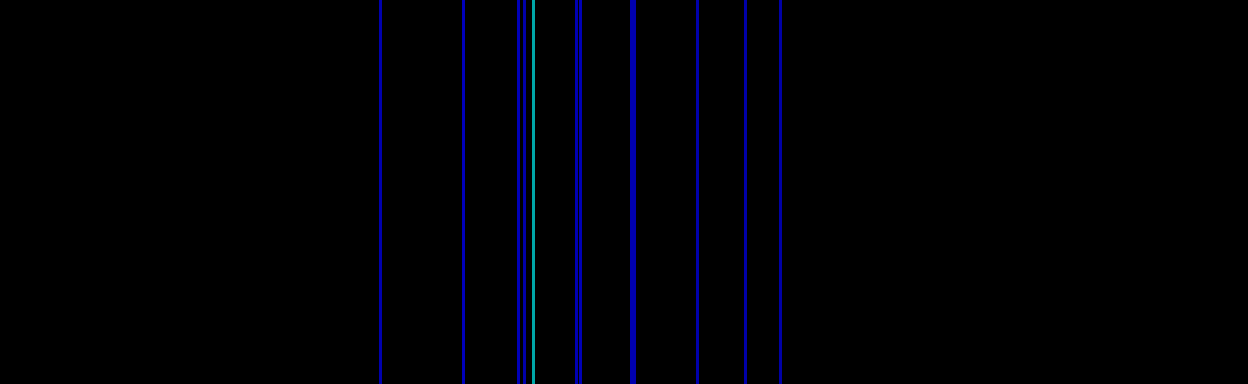} \\
    \end{tabular}        
  \caption{Example inputs to HRFuser from DENSE. From left to right: RGB image, warped gated camera image, projected lidar points, projected radar points. The radar and lidar projections are highlighted and enlarged for better visualization. Best viewed on a screen at full zoom.}
  \label{fig:supl:results:stf}
\end{figure*}

\PAR{DENSE~\cite{seeing:through:fog}.} The dataset provides 1920$\times$1024 camera images, lidar and radar points, and 1280$\times$720 gated camera images, captured under a variety of normal and adverse weather conditions. We process the inputs in the same way as~\cite{seeing:through:fog}. 
The camera is cropped to a 1248$\times$360 window around the center of the gated camera. The image from the gated camera is transformed into the image plane of the camera using a homography mapping as in~\cite{seeing:through:fog}. 
We also crop the annotated 2D bounding boxes to the aforementioned 1248$\times$360 window, discarding boxes for which more than 90\% of the original box area lies outside the crop. The gated camera is cropped to the same window. The strongest lidar return and radar are projected onto the image plane. As the radar cross-section (RCS) data are not publicly available, we use only 2 radar channels: depth and velocity over ground. However, RCS data were used for training the method of~\cite{seeing:through:fog}, 
so comparing~\cite{seeing:through:fog} 
to our method is actually unfair to our method. Example inputs are displayed in \cref{fig:supl:results:stf}.
We train on 3 classes defined as shown in \cref{tab:stf:map}, and evaluate only on the car class, using the KITTI evaluation framework~\cite{kitti}. 
We run the training of HRFuser-T on 4 Nvidia Titan RTX GPUs with a batch size of 12.

\begin{figure*}
    \centering
    \begin{tabular}{@{}c@{\hspace{0.05cm}}c@{\hspace{0.05cm}}c@{}}    
        \subfloat{\footnotesize RGB Camera} &
        \subfloat{\footnotesize Lidar} &
        \subfloat{\footnotesize Radar} \\
        \vspace{-0.07cm}
        \includegraphics[width=0.328\textwidth]{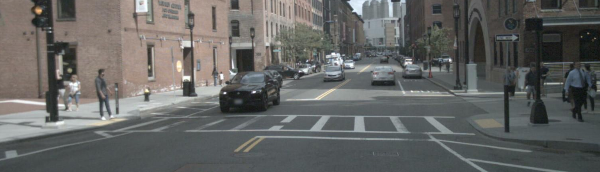} &
        \includegraphics[width=0.328\textwidth]{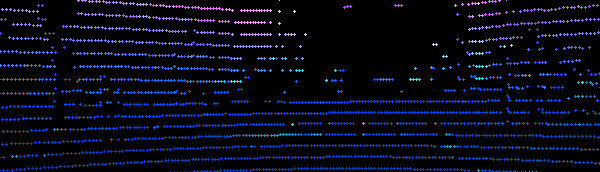} &
        \includegraphics[width=0.328\textwidth]{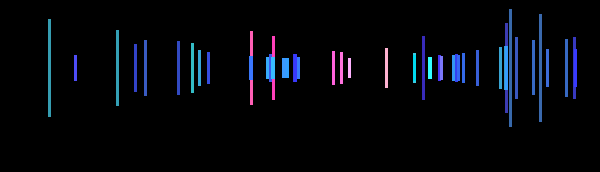} \\
        \vspace{-0.07cm}
        \includegraphics[width=0.328\textwidth]{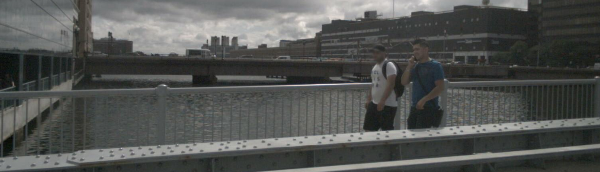} &
        \includegraphics[width=0.328\textwidth]{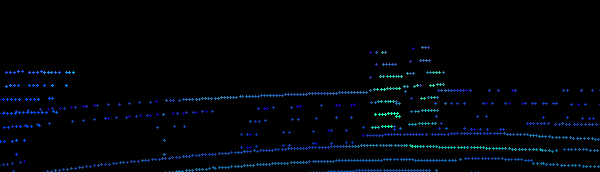} &
        \includegraphics[width=0.328\textwidth]{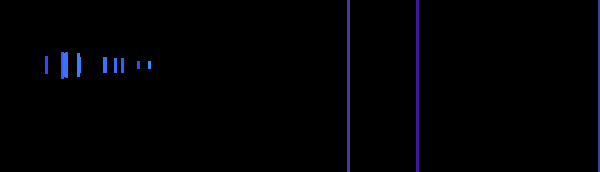} \\
        \includegraphics[width=0.328\textwidth]{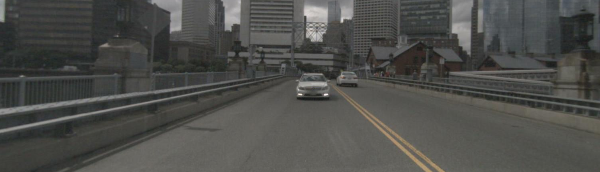} &
        \includegraphics[width=0.328\textwidth]{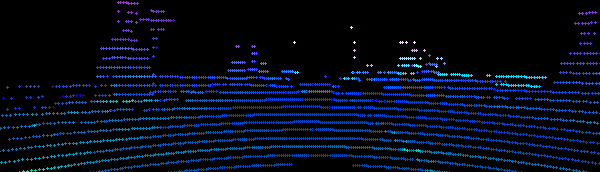} &
        \includegraphics[width=0.328\textwidth]{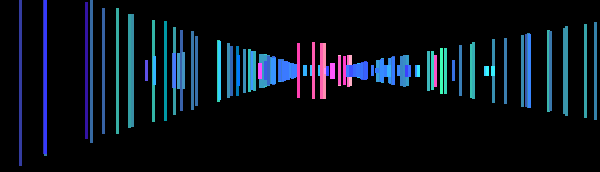} 
    \end{tabular}    
  \caption{Example inputs to HRFuser from nuScenes. From left to right: RGB image, projected lidar points, projected radar points. The radar and lidar projections are highlighted and enlarged for better visualization. Best viewed on a screen at full zoom.}
  \label{fig:supl:results:nuscenes}
\end{figure*}

\PAR{nuScenes~\cite{nuScenes}.} 
Similar to~\cite{deep:learning:based:radar:camera:fusion:architecture}, 
we resize the recorded 1600$\times$900 images to 640$\times$360 and project the radar points as 3m-high pillars onto the image plane. This creates a 640$\times$360 projected radar image with 3 channels: range, RCS and velocity over ground. Compared to~\cite{deep:learning:based:radar:camera:fusion:architecture}, 
we do not accumulate radar data across time or filter them in any way. The lidar points are projected onto the image plane, yielding a 640$\times$360 image with 3 channels: range, intensity and height. Example inputs are displayed in \cref{fig:supl:results:nuscenes}. All input channels are normalized over the entire dataset. We run the training of HRFuser-T on 4 Nvidia RTX 2080 TI GPUs with a batch size of 12.
We follow the mmdet3d~\cite{mmdet3d} 
framework and use a set of 10 classes for training and evaluation, which are defined based on the original nuScenes classes as shown in \cref{tab:nus:map}.

\PAR{Clarification on nuScenes classes used in experiments and class-wise performance.}
As we mention in the main paper, ``unless otherwise stated'' we indeed use the set of 10 nuScenes classes in our experiments. The only case where we use a reduced set of 6 classes is in \cref{table:sota:nus}, 
in which this choice of classes was necessary for comparability to the method in~\cite{radar:camera:fusion:joint:detection:distance:estimation}, which only reports results on these 6 classes. This different choice of classes has been explicitly stated in the caption of \cref{table:sota:nus}.

\begin{table}
    \caption{Mapping from original DENSE classes to the set of classes we use for training.}
    \label{tab:stf:map}
    \vspace{1em}
    \centering
    \begin{tabular}{cc}
  \toprule
        DENSE Class & Mapped Class \\
  \midrule
        PassengerCar & Car\\
        Pedestrian & Pedestrian \\
        RidableVehicle &  Cyclist\\
        LargeVehicle & DontCare\\
        Vehicle &  DontCare\\
        DontCare &  DontCare\\
  \bottomrule
    \end{tabular}
\end{table}

\begin{table}
    \caption{Mapping from original nuScenes classes to our default set of training and evaluation classes~\cite{mmdet3d}.
    }
    \label{tab:nus:map}
    \vspace{1em}
  \resizebox{\columnwidth}{!}{%
    \centering
    \begin{tabular}{cc}
  \toprule
        NuScenes Class & Mapped Class \\
  \midrule
        vehicle.car & car\\
        vehicle.truck & truck\\
        vehicle.trailer  &  trailer\\
        vehicle.bus.bendy  & bus \\
        vehicle.bus.rigid  & bus \\
        vehicle.construction  & construction\_vehicle \\
        vehicle.bicycle &  bicycle\\
        vehicle.motorcycle  &  motorcycle\\
        human.pedestrian.child & pedestrian \\
        human.pedestrian.adult & pedestrian\\
        human.pedestrian.construction\_worker  & pedestrian \\
        human.pedestrian.police\_officer  & pedestrian \\
        movable\_object.trafficcone  &  traffic\_cone\\
        movable\_object.barrier  & barrier \\
  \bottomrule
    \end{tabular}}
\end{table}

\subsection{Additional Ablations}
\label{app:additional:ablations}

\PAR{Choice of primary modality.}
As mentioned in~\cref{subsubsec:exp:modablation}, we investigate the effect of different sensors as a primary modality and provide a comparison on DENSE in \cref{table:supl:abl:mod}. While we have selected the camera as the primary modality for our HRFuser architecture, in this experiment we set each available sensor besides the camera, i.e., lidar, radar, and gated camera, as the primary modality. The RGB camera becomes thereby a secondary modality and is treated as the others. 
We observe that having either the camera or the gated camera as the primary sensor generally attains higher performance than having lidar and radar as the primary sensor, even though the respective difference is slight. Our intuition for this finding is that the high spatial resolution of the camera and the gated camera makes them better choices for serving as the primary modality, as the primary modality is used in our cross-attention block to compute the queries and thus to determine which regions of the other modalities to attend to, a function which can be carried out with higher spatial accuracy in case high-resolution readings are available from the primary modality. However, since fusion across all modalities starts at an early stage in the network, HRFuser can learn meaningful features even when using a sparse modality such as radar as the primary modality, and in any case, HRFuser is fairly robust with respect to which modality serves as the primary one.

\begin{table*}[!tb]
  \caption{Ablation on DENSE on the selection of different primary modalities. Results are in AP.}
  \label{table:supl:abl:mod}
  \centering
  \setlength\tabcolsep{2pt}
  \begin{tabular*}{\linewidth}{l @{\extracolsep{\fill}} cccccccccccc}
  \toprule
  Primary &  \multicolumn{3}{c}{\textbf{clear}} & \multicolumn{3}{c}{\textbf{light fog}} &  \multicolumn{3}{c}{\textbf{dense fog}} & \multicolumn{3}{c}{\textbf{snow/rain}}\\
  Sensor & easy & mod. & hard & easy & mod. & hard & easy & mod. & hard & easy & mod. & hard \\
  \midrule
  RGB camera (ours) &     90.15 & \textbf{87.10} & 79.48 & 90.60 & 89.34 & \textbf{86.50} & 87.93 & 80.27 & 78.21 & \textbf{90.05} & \textbf{85.35} & \textbf{78.09}\\
  Lidar & 90.02 & 86.89 & 79.35 & 90.61 & \textbf{89.45} & 86.28 & 88.64 & 80.60 & 78.59 & 89.86 & 84.98 & 77.44\\
  Radar & 89.99 & 86.96 & 79.47 & \textbf{90.65} & 89.30 & 80.89 & 88.45 & 80.53 & 72.33 & 89.86 & 85.08 & 77.46\\
  Gated camera & \textbf{90.20} & 86.82 & \textbf{79.54} & 90.58 & 89.33 & 80.92 & \textbf{88.73} & \textbf{80.87} & \textbf{79.05} & 90.03 & 85.29 & 77.82\\
  \bottomrule
  \end{tabular*}
\end{table*}

\PAR{PVTv2 adaptations.} We create the alternative attention mechanisms PVTv2-CA and PVTv2-Li-CA, which are presented in \cref{table:stf:modality:ablation}
of the main paper, by adapting the state-of-the-art transformer PVTv2~\cite{PVTv2:vision:transformer}. 
We adapt its spatial reduction attention module for cross-attention by entering the query from our primary branch and the key and value from our secondary branch, and pass this into their proposed convolutional feed-forward module employing depth-wise convolution. In contrast to PVTv2, we do not incorporate the overlapping patch embedding, in order to keep the spacial dimensions of the feature maps unchanged. No pre-training is applied when training PVTv2, which is also the case for all other presented methods. 

\PAR{Parallel cross-attention skip connection.} To examine the benefit of the skip connections for the secondary modalities which are involved in our parallel CA block as shown in \cref{fig:parallel:ca}
of the main paper, we remove these skip connections (blue and orange in 
\cref{fig:parallel:ca})
and observe in \cref{table:details:nus} a drop of 0.6\% in AP relative to our default MWCA. This finding indicates that skip connections from all modalities are beneficial for cross-attention, as they allow the network to attend to details without having to learn the identity function.

\begin{table}[!tb]
  \caption{Additional ablation of MWCA on nuScenes. MWCA: our MWCA fusion block, MWCA$_{\text{w/o skip}}$: our fusion block without skip connections for secondary modalities in the parallel CA block.}
  \label{table:details:nus}
  \centering
  \setlength\tabcolsep{2pt}
  \small
  \begin{tabular*}{\linewidth}{l @{\extracolsep{\fill}} ccccccc}
  \toprule
  Method & AP & AP$_{0.5}$ & AP$_{0.75}$ & AP$_m$ & AP$_l$ & AR\\
  \midrule
  HRFuser-T (MWCA$_{\text{w/o skip}}$)~ & 30.9& 56.6 & 30.0 & 22.1 &41.9 &41.9 \\
  HRFuser-T (MWCA) & \best{31.5} & \best{57.4} & \best{31.1} & \best{22.7} & \best{42.5} &\best{42.3}\\
  \bottomrule
  \end{tabular*}
\end{table}

\PAR{Amount of training data.}
We investigate whether the performance gain of HRFuser is due to the increased number of ``inputs'' or due to general network strength and better learned features. We trained HRFuser-T with two modalities (lidar + camera) on a subset of nuScenes containing half the training data. As seen in the \cref{table:supl:nus:data}, this model substantially outperforms the camera-only model trained on the complete training set, recovering ~90\% of the performance gain of the fusion model that sees the complete training set. Thus, even with the same quantity of ``inputs'' as the camera-only model trained on the full nuScenes training set, HRFuser delivers a significant performance improvement.

\begin{table}[!tb]
  \caption{Ablation study on using a smaller training set. Validation set results of HRFuser-T trained on a subset of nuScenes containing either all or half the training split are reported. C: camera, L: lidar.}
  \label{table:supl:nus:data}
  \centering
  \setlength\tabcolsep{2pt}
  \begin{tabular*}{\linewidth}{l @{\extracolsep{\fill}} ccc}
  \toprule
   & nuScenes (\%) & Modalities & AP \\
  \midrule
  HRFuser-T & 100 & C & 26.5\\
  HRFuser-T & 50 & CL & 30.8 (+4.3)\\
  HRFuser-T & 100 & CL & 31.2 (+4.7)\\
  \bottomrule
  \end{tabular*}
\end{table}

Note that the volume of information is not the same across modalities. In nuScenes, an average of only 0.71\% of the pixels per radar image and 1.61\% per lidar image have a measurement. Thus, adding a second modality does not double the volume of information but only increases it slightly, which means that the comparison of a camera-only model to a fusion model is reasonably fair.

\subsection{Class-wise Performance on nuScenes}

We report in \cref{table:supl:nus:classwise} the average precision of our HRFuser-T versus the baseline camera-only HRFormer-T on each individual class on nuScenes. HRFuser consistently outperforms HRFormer across all classes by large margins. 

\begin{table}[!tb]
  \caption{Class-wise performance on nuScenes. Results are in AP.}
  \label{table:supl:nus:classwise}
  \centering
  \setlength\tabcolsep{2pt}
  \begin{tabular*}{\linewidth}{l @{\extracolsep{\fill}} ccc}
  \toprule
  Class & HRFormer-T & HRFuser-T\\
  \midrule
  car & 50.2 & \textbf{53.1}\\
  truck & 29.2 & \textbf{36.8}\\
  trailer & 14.9 & \textbf{20.6}\\
  bus & 39.4 & \textbf{48.1}\\
  construction vehicle & 7.0 & \textbf{9.6}\\
  bicycle & 20.8 & \textbf{26.6}\\
  motorcycle & 22.3 & \textbf{28.8}\\
  pedestrian & 25.4 & \textbf{30.7}\\
  traffic cone & 26.8 & \textbf{28.6}\\
  barrier & 28.8 & \textbf{32.0}\\
  \bottomrule
  \end{tabular*}
\end{table}

\subsection{HRFuser with an HRNet-based Backbone}

\cref{table:suppl:nus:hrnet}
 includes \textit{HRFuser-w18 (HRNet)}, a variant of HRFuser built upon HRNetV2-w18~\cite{hrnet}. 
In this variant, we keep the same transformer-based MWCA fusion mechanism with the same parameters as for the default, HRFormer-based HRFuser. However, the camera branch of this variant, in which our MWCA fusion blocks are inserted, resembles HRNetV2p-w18 and follows the HRNet architecture using ``Basic'' blocks introduced in~\cite{hrnet}. 
The secondary modality branches we introduce follow analogously the design of the highest-resolution branch of HRNetV2p-w18. \cref{table:suppl:nus:hrnet}
shows a 4.3\% improvement in AP of our HRFuser-w18 over the camera-only HRNetV2-w18, which demonstrates the generality of the components introduced in HRFuser, as they benefit various dense prediction networks such as HRNet and HRFormer.

\begin{table}
  \caption{Comparison of 2D detection methods on nuScenes evaluated on 6 classes: car, truck, bus, bicycle, motorcycle and pedestrian. 
  C: camera, R: radar, L: lidar.}
  \label{table:suppl:nus:hrnet}
  \smallskip
  \resizebox{\columnwidth}{!}{%
  \centering
  \setlength\tabcolsep{3pt}
  \small
  \begin{tabular}{lccccccc}
  \toprule
  Method & Modalities & AP & AP$_{0.5}$ & AP$_{0.75}$ & AP$_m$ & AP$_l$ & AR\\
  \midrule
  HRNetV2p-w18~\cite{hrnet} & C & 32.4 & 56.6 & 33.5  & 21.0 & 43.7 & 43.4\\ 
  HRFuser-w18 (HRNet) & CRL & \best{36.7} & \best{63.1} & \best{38.1}  & \best{24.9} & \best{48.6} & \best{47.0} \\
  \bottomrule
  \end{tabular}}
\end{table}

\subsection{Comparison to CRF-Net Using Only Radar}

In \cref{tab:crf:radar}, we compare on nuScenes the CRF-Net~\cite{deep:learning:based:radar:camera:fusion:architecture} 
to a version of HRFuser which only uses radar besides the camera, i.e., omitting lidar. This comparison serves in investigating whether HRFuser can leverage information from the radar better than the competing CRF-Net, which focuses explicitly on the radar modality. Indeed, HRFuser-T$_{\text{radar}}$ yields a 4.9\% improvement in AP over CRF-Net, even when the radar is the only secondary modality.

\begin{table*}[!tb]
  \caption{Additional comparison on nuScenes of CRF-Net~\cite{deep:learning:based:radar:camera:fusion:architecture} 
against a radar-only HRFuser, both evaluated on 6 classes: car, truck, bus, bicycle, motorcycle and pedestrian, using the split from~\cite{deep:learning:based:radar:camera:fusion:architecture} 
for training and evaluation. C: camera, R: radar.}
  \label{tab:crf:radar}
  \centering
  \setlength\tabcolsep{2pt}
  \small
  \begin{tabular*}{\linewidth}{l @{\extracolsep{\fill}} ccccccc}
  \toprule
  Method & Modalities & AP & AP$_{0.5}$ & AP$_{0.75}$ & AP$_m$ & AP$_l$ & AR\\
  \midrule
  CRF-Net~\cite{deep:learning:based:radar:camera:fusion:architecture} 
& CR & 27.0 &  42.7 & 29.0 & 22.7 & 35.6 & 31.3\\ 
  HRFuser-T$_{\text{radar}}$ (HRFormer) & CR & \best{31.9} & \best{58.2} & \best{31.6} & \best{23.9} & \best{45.2} & \best{42.6}\\
  \bottomrule
  \end{tabular*}
\end{table*}

\subsection{Multiple Modalities and Overfitting}

Fusing multiple modalities not only allows to build more robust features but also helps against overfitting. This is demonstrated in \cref{table:sota:nus} of the main paper, where HRFormer-T outperforms the much
larger HRFormer-B by 0.5\% in AP, but HRFuser-B outperforms the smaller HRFuser-T by 0.5\% in  AP.

\subsection{MWCA Attention Map Analysis}

\begin{figure*}
    \centering
    \begin{tabular}{@{}c@{\hspace{0.05cm}}c@{}}    
        \subfloat{\footnotesize RGB Image} &
        \subfloat{\footnotesize Lidar Attention Map} \\
        \vspace{-0.07cm}
        \includegraphics[width=0.49\textwidth]{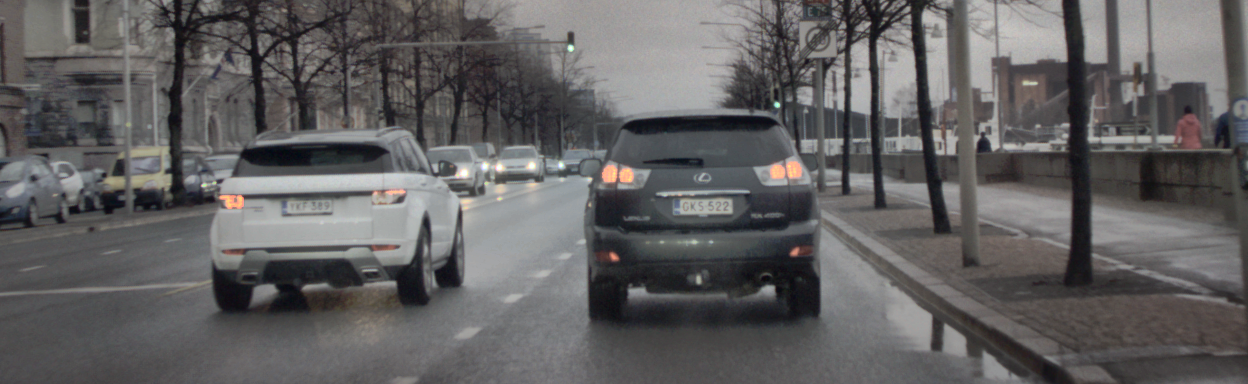} &
        \includegraphics[width=0.49\textwidth]{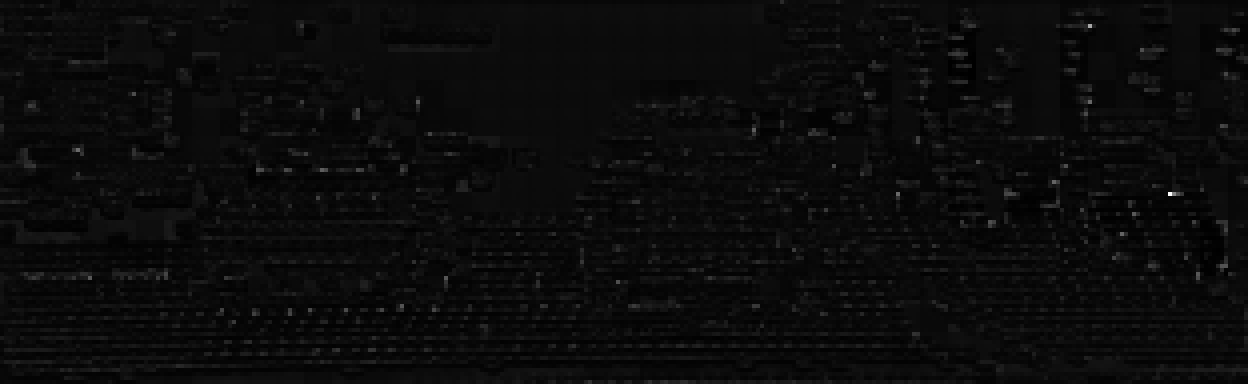} \\
        \subfloat{\footnotesize Radar Attention Map} &
        \subfloat{\footnotesize Gated Camera attention Map} \\
        \vspace{-0.07cm}
        \includegraphics[width=0.49\textwidth]{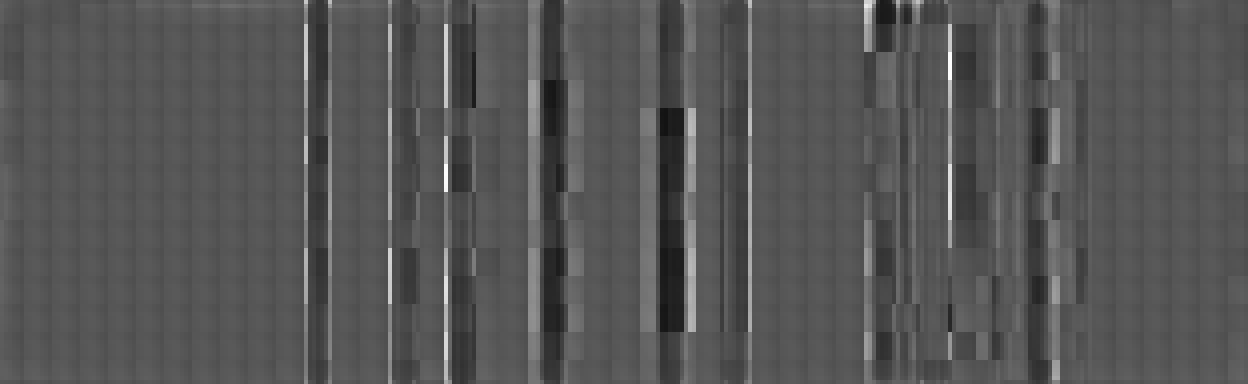} &
        \includegraphics[width=0.49\textwidth]{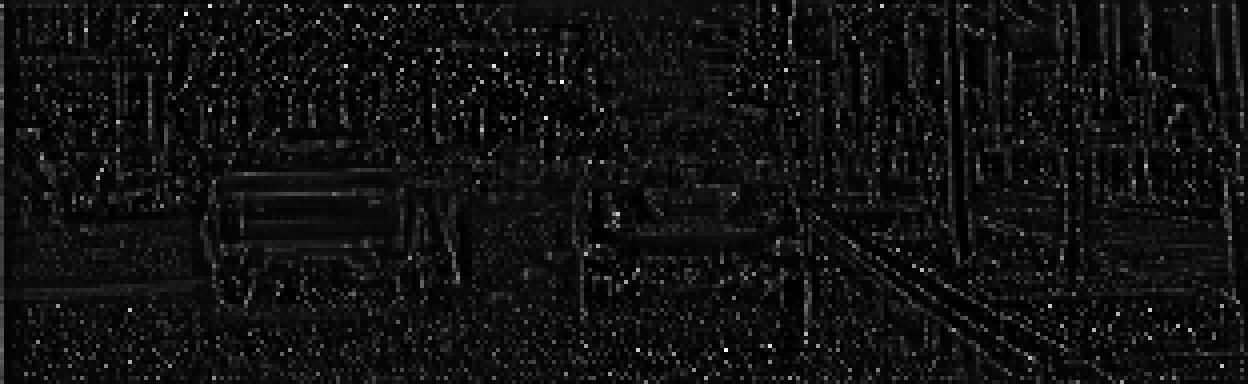}
    \end{tabular}    
  \caption{Visualization of attention maps from MWCA at the first fusion stage into the highest-resolution stream. Top left: RGB image, top right: lidar attention map, bottom left: radar attention map, bottom right: gated camera attention map. Best viewed on a screen at full zoom.}
  \label{fig:attention:map}
\end{figure*}

To investigate what our proposed MWCA learns, we visualize some exemplary attention maps of the cross-attention fusion. As seen in \cref{fig:attention:map}, MWCA attends to each modality individually and in a sensor-specific fashion. The sampling pattern thereby reflects the characteristics of the sensors:

\begin{itemize}
    \item The lidar attention map is sparse and highlights only individual points. Areas without lidar returns--such as the sky--are ignored. 
    \item The radar attention map follows the vertical columns of the radar input, but puts additional highlight on horizontal areas corresponding to cars.
    \item The gated camera attention map highlights the dominant edges and structures in the image. The high resolution and density of the gated camera allows to attend to fine details.
\end{itemize}

MWCA learns continuous reasoning across windows, as demonstrated by the continuous nature of the attention maps. This is especially noticeable in the gated camera attention map, where the highlighted edges are continuous and uninterrupted by the boundaries of each local window.

\subsection{Additional Qualitative Results}

\begin{figure*}
    \centering
    \begin{tabular}{@{}c@{\hspace{0.05cm}}c@{\hspace{0.05cm}}c@{}}
        \multicolumn{1}{c}{\footnotesize Ground Truth} &
        \multicolumn{1}{c}{\footnotesize HRFormer~\cite{hrformer}} &
        \multicolumn{1}{c}{\footnotesize HRFuser} \\
        \vspace{-0.07cm}
        \includegraphics[width=0.328\textwidth]{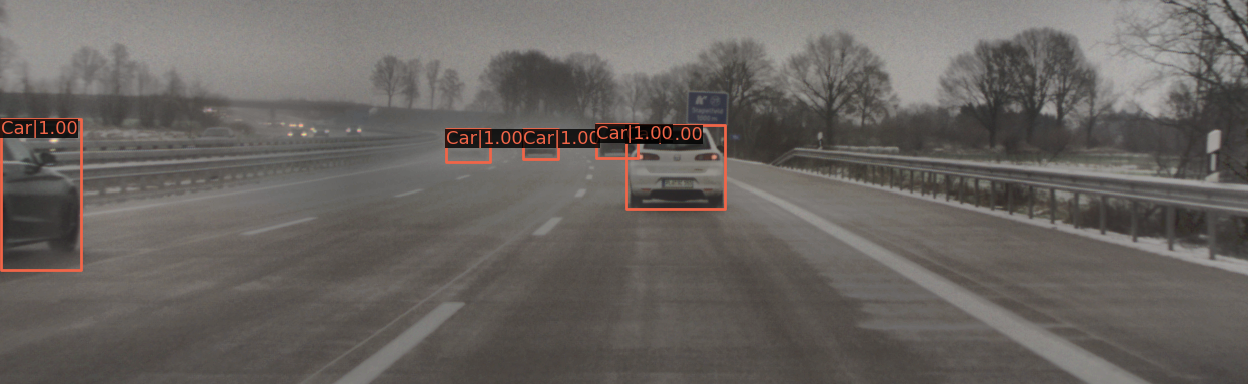} &
        \includegraphics[width=0.328\textwidth]{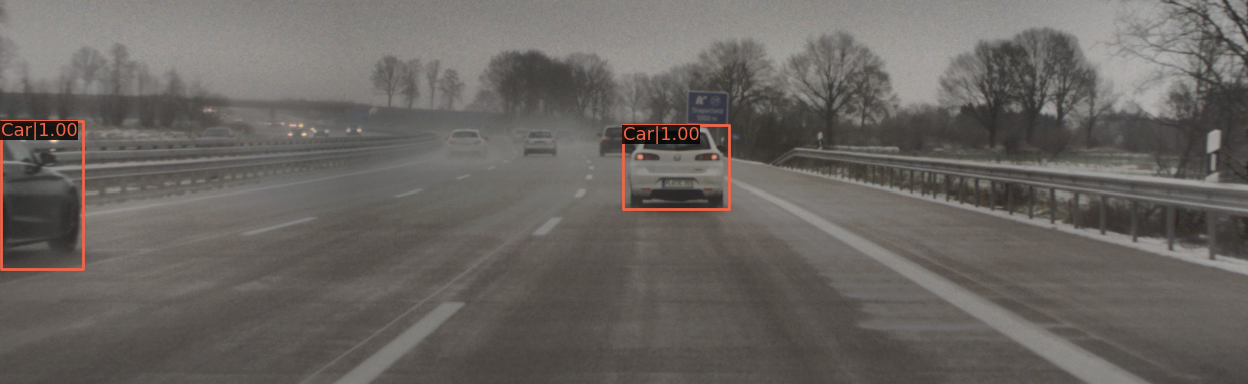} &
        \includegraphics[width=0.328\textwidth]{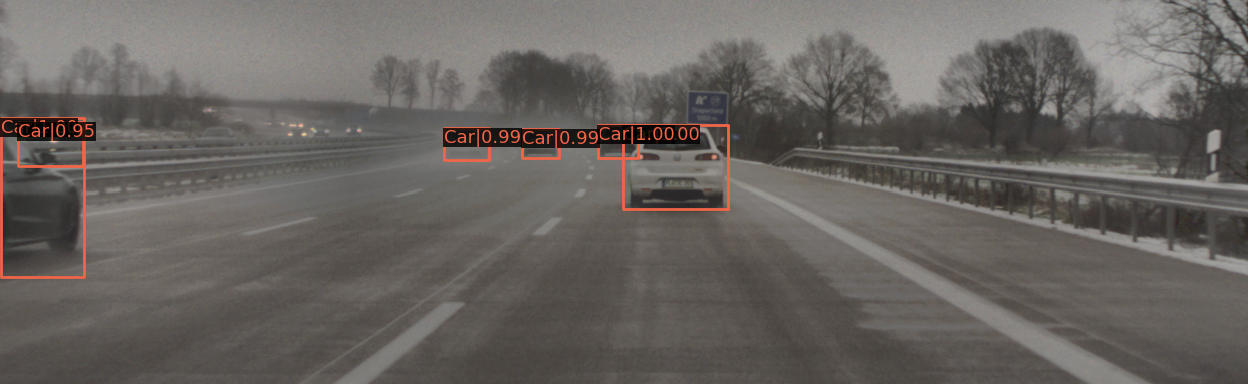} \\
        \vspace{-0.07cm}
        \includegraphics[width=0.328\textwidth]{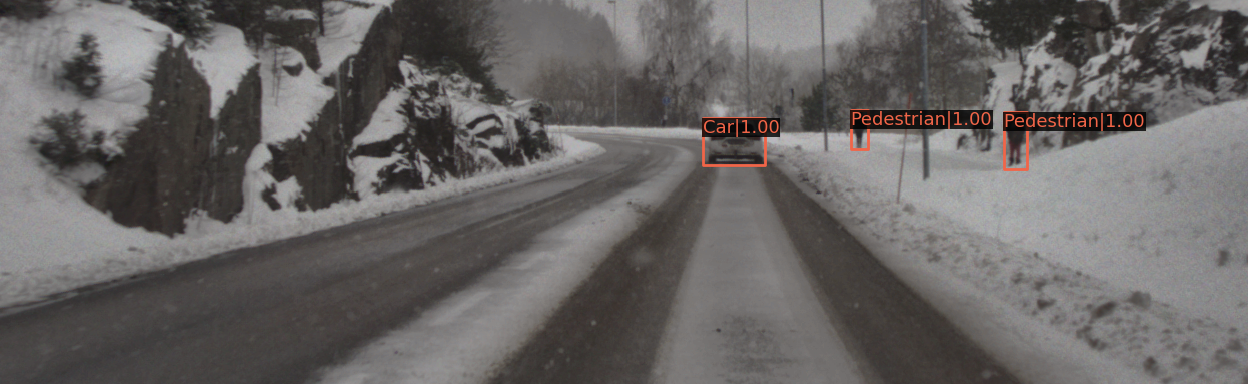} &
        \includegraphics[width=0.328\textwidth]{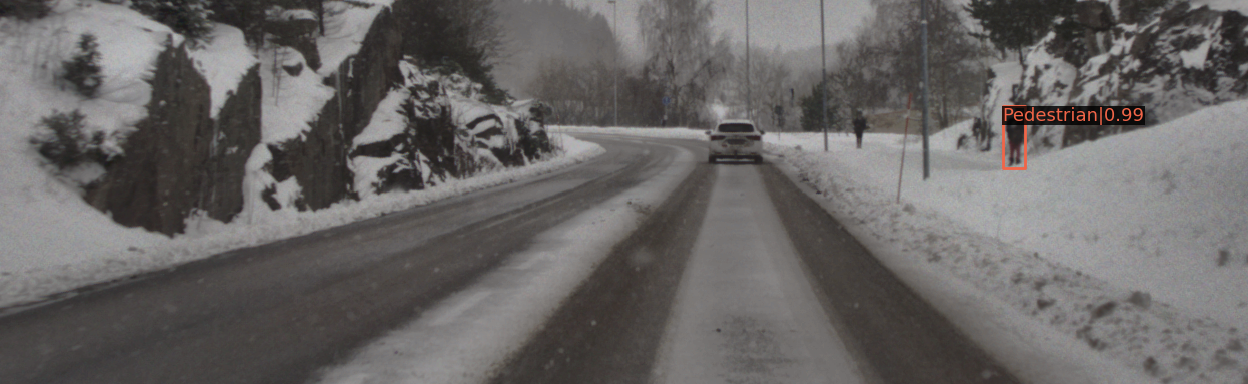} &
        \includegraphics[width=0.328\textwidth]{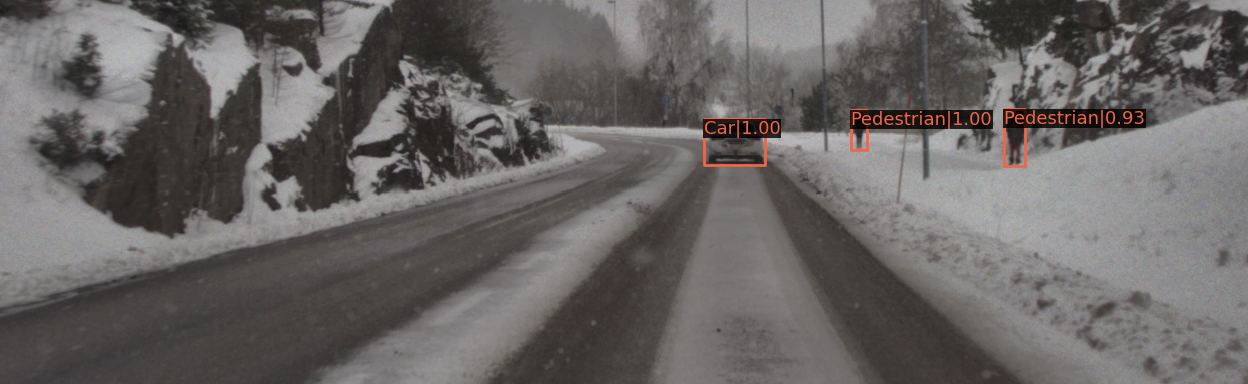} \\
        \vspace{-0.07cm}
        \includegraphics[width=0.328\textwidth]{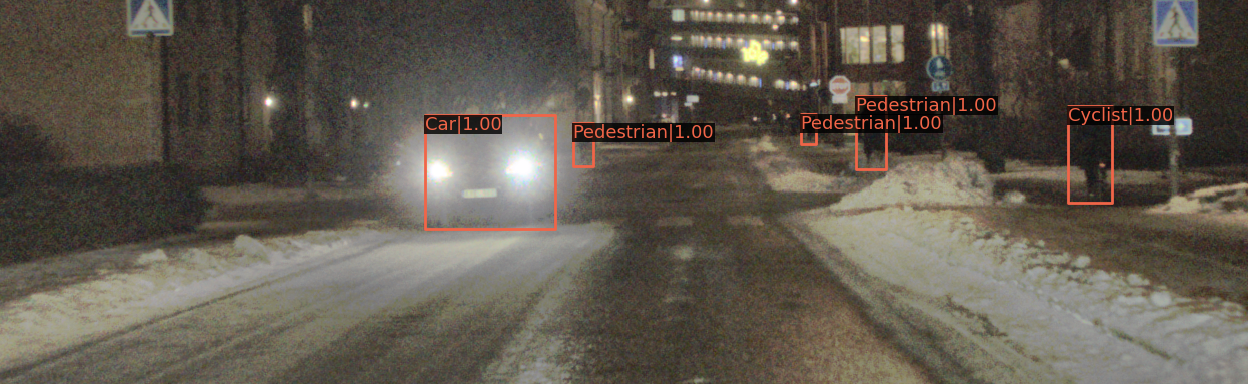} &
        \includegraphics[width=0.328\textwidth]{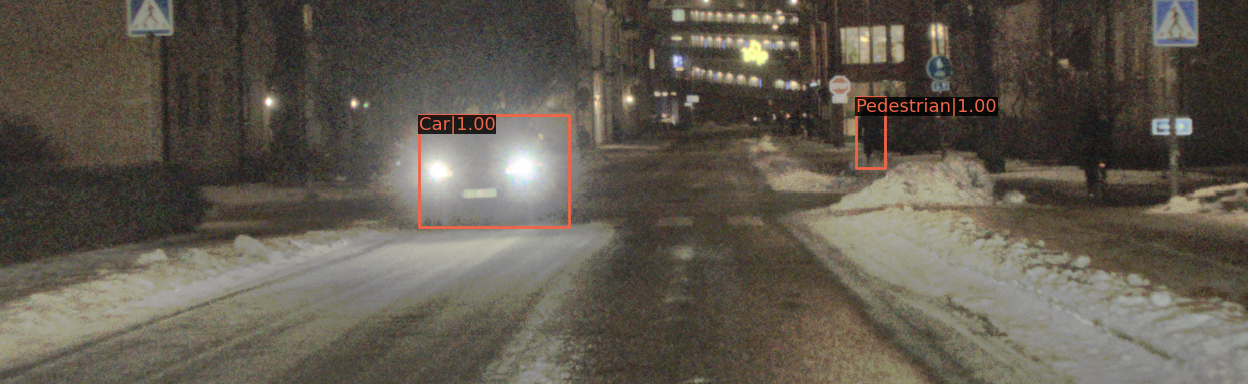} &
        \includegraphics[width=0.328\textwidth]{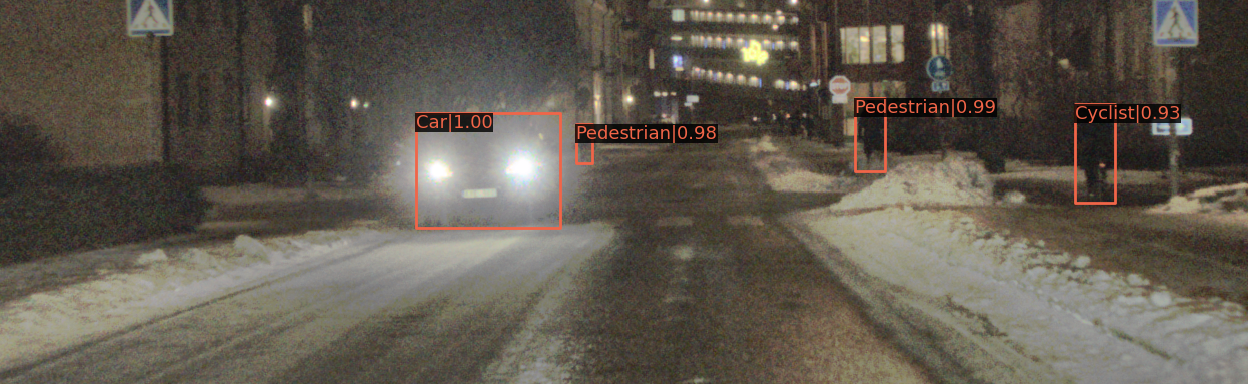} \\
        \vspace{-0.07cm}
        \includegraphics[width=0.328\textwidth]{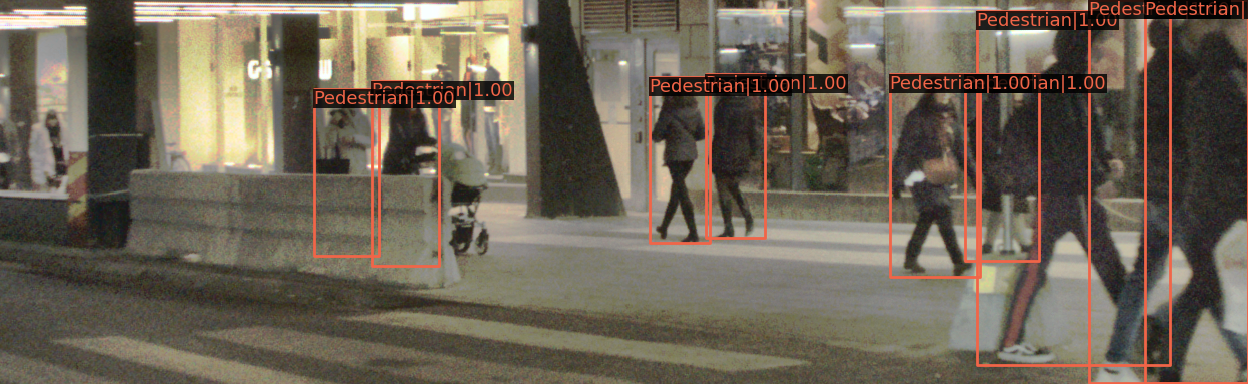} &
        \includegraphics[width=0.328\textwidth]{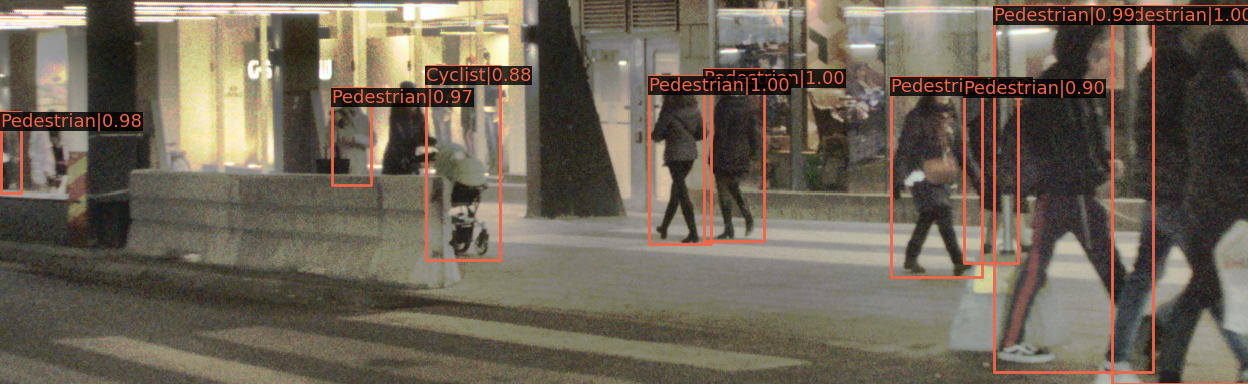} &
        \includegraphics[width=0.328\textwidth]{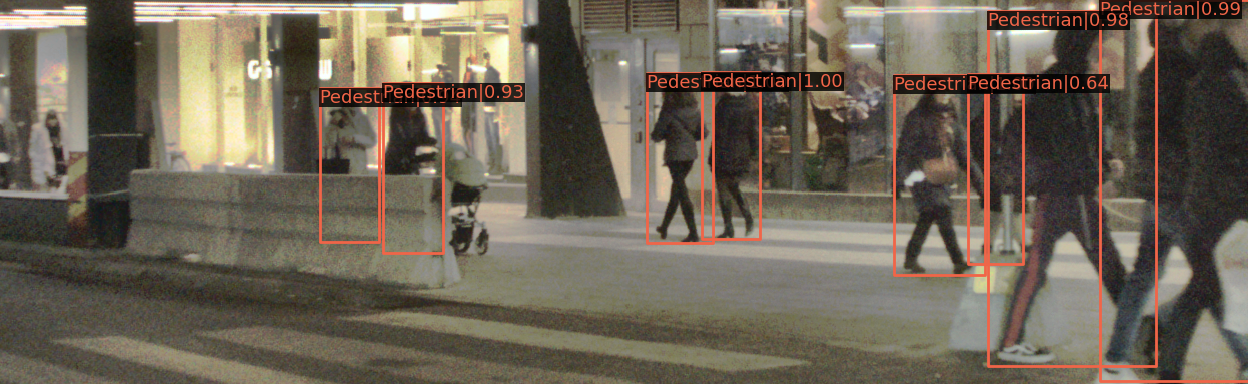} \\
        \includegraphics[width=0.328\textwidth]{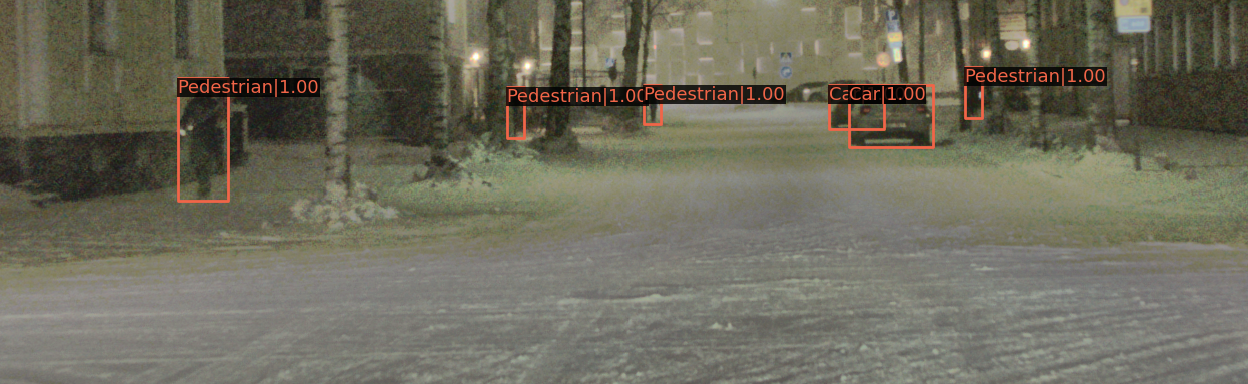} &
        \includegraphics[width=0.328\textwidth]{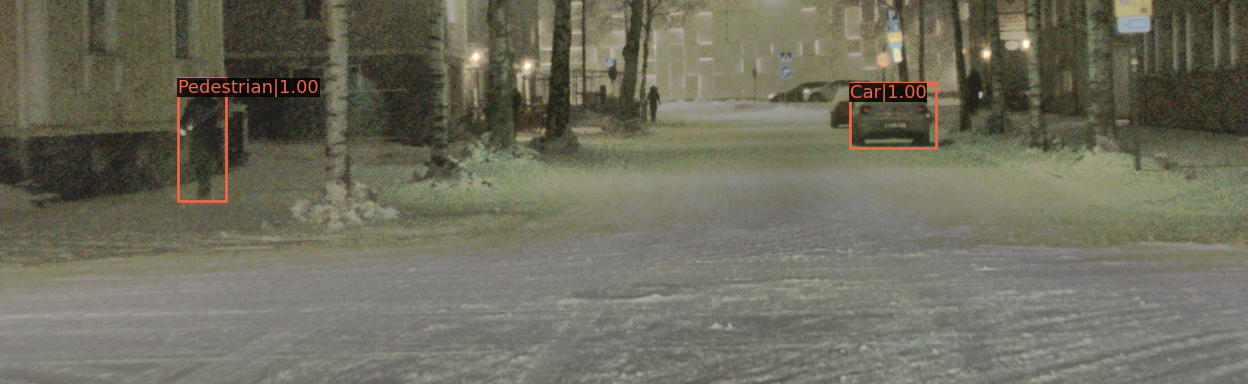} &
        \includegraphics[width=0.328\textwidth]{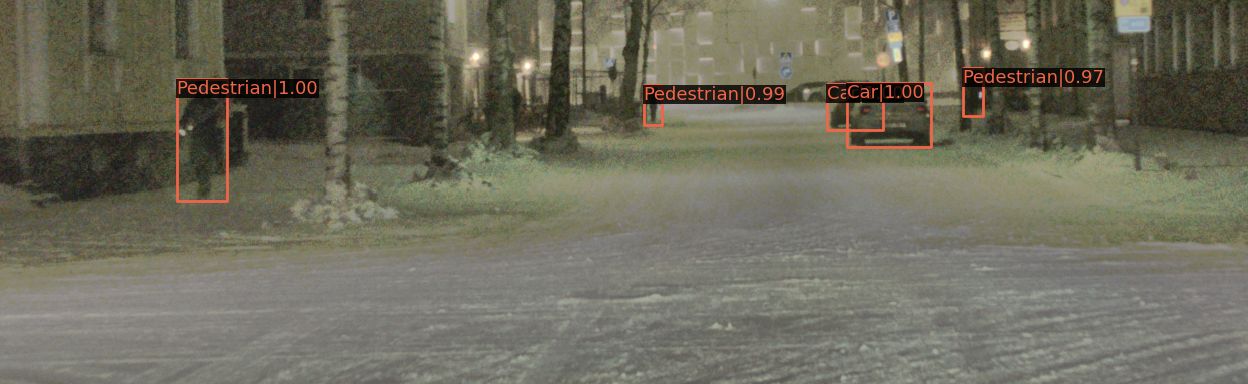}
    \end{tabular}
    \caption{Further qualitative detection results on DENSE. From left to right: image with ground-truth annotation, prediction of HRFormer, prediction of HRFuser. Best viewed on a screen at full zoom.}
    \label{fig:further:results:stf}
\end{figure*}

\begin{figure*}
    \centering
    \begin{tabular}{@{}c@{\hspace{0.05cm}}c@{\hspace{0.05cm}}c@{}}
        \multicolumn{1}{c}{\footnotesize Ground Truth} &
        \multicolumn{1}{c}{\footnotesize HRFormer~\cite{hrformer}} &
        \multicolumn{1}{c}{\footnotesize HRFuser} \\
        \vspace{-0.07cm}
        \includegraphics[width=0.328\textwidth]{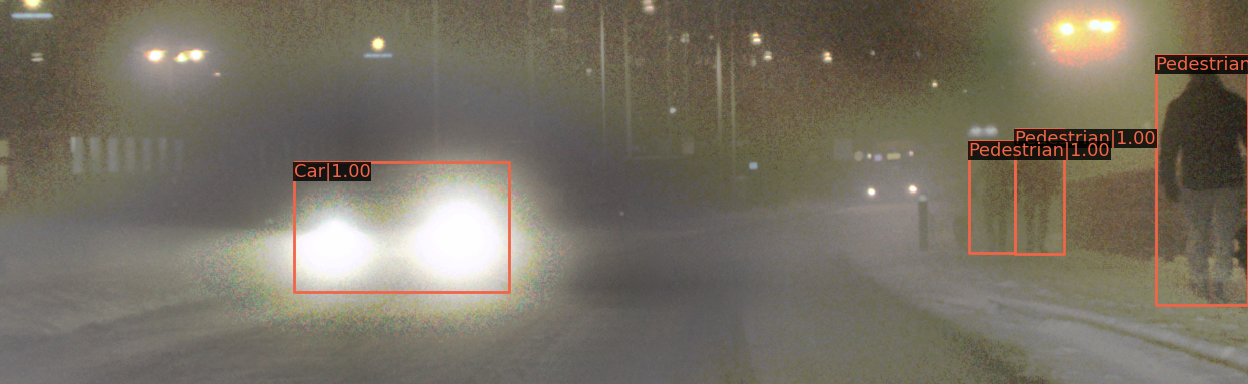} &
        \includegraphics[width=0.328\textwidth]{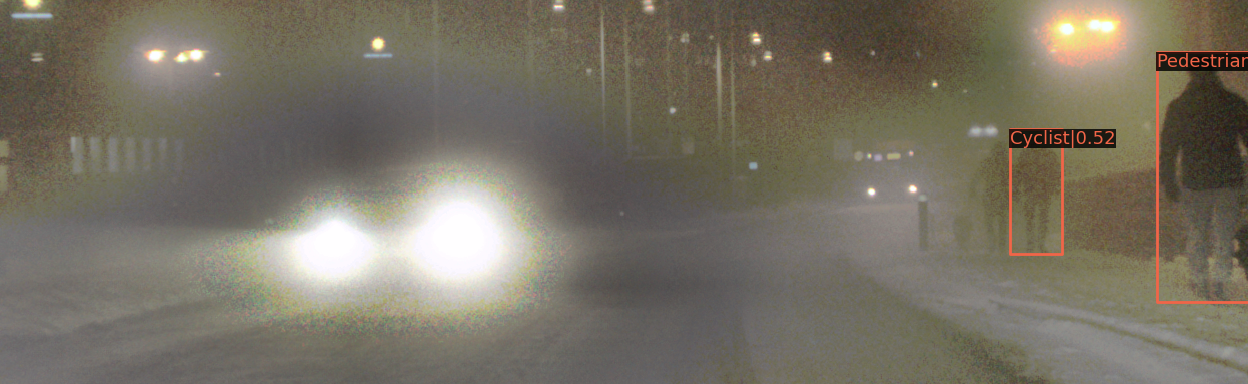} &
        \includegraphics[width=0.328\textwidth]{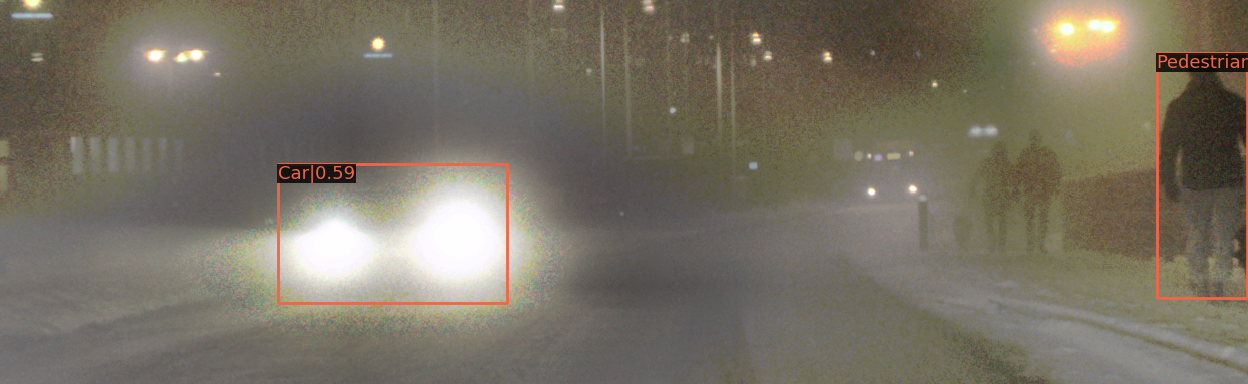} \\
        \includegraphics[width=0.328\textwidth]{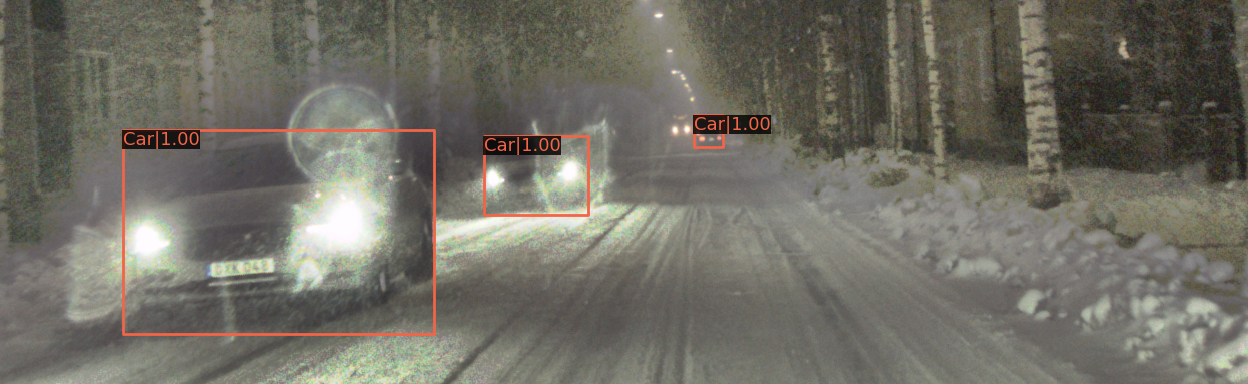} &
        \includegraphics[width=0.328\textwidth]{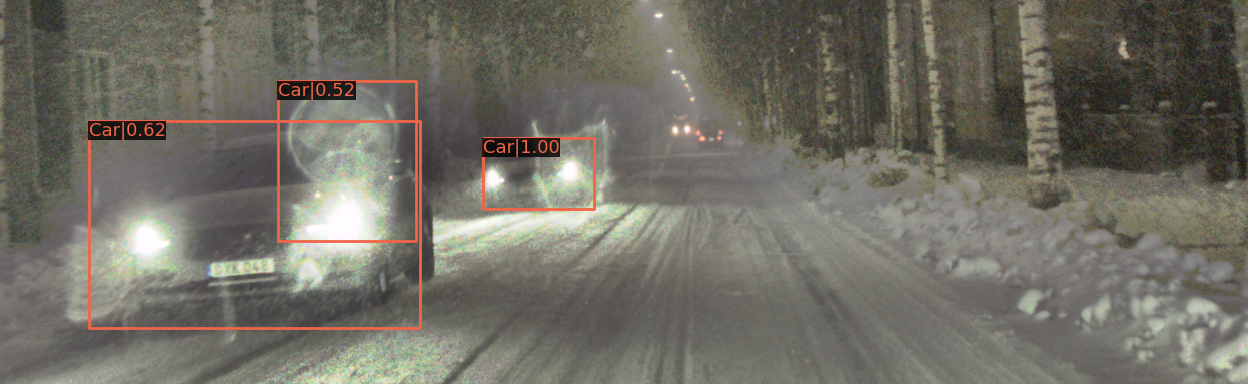} &
        \includegraphics[width=0.328\textwidth]{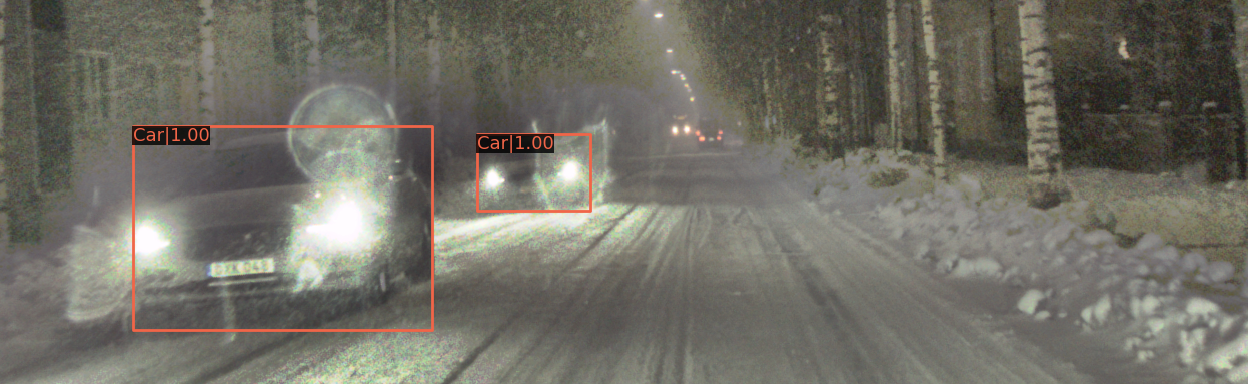}
    \end{tabular}
    \caption{Further qualitative detection results on DENSE showing failure cases of HRFuser. From left to right: image with ground-truth annotation, prediction of HRFormer, prediction of HRFuser. Best viewed on a screen at full zoom.}
    \label{fig:further:results:stf:failure}
\end{figure*}

\begin{figure*}
    \centering
    \begin{tabular}{@{}c@{\hspace{0.05cm}}c@{\hspace{0.05cm}}c@{\hspace{0.05cm}}c@{}}
        \subfloat{\footnotesize Ground Truth} &
        \subfloat{\footnotesize HRFormer~\cite{hrformer}} &
        \subfloat{\footnotesize BEVFusion~\cite{liu2022bevfusion}} &
        \subfloat{\footnotesize HRFuser} \\
        \vspace{-0.07cm}
        \includegraphics[width=0.245\textwidth]{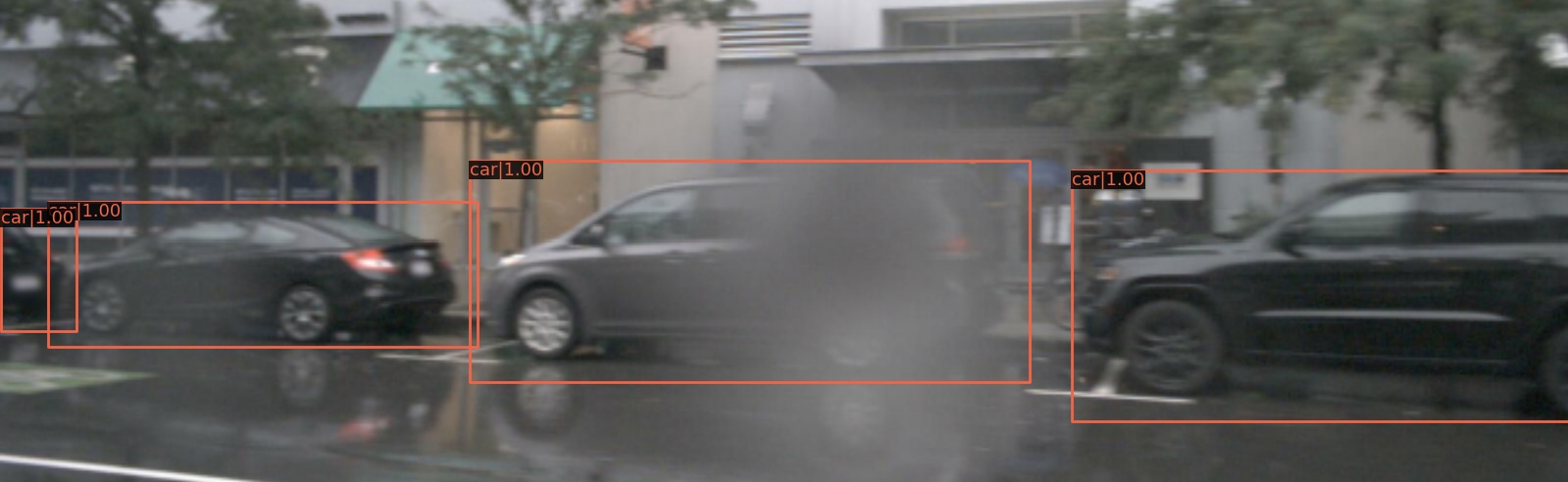} &
        \includegraphics[width=0.245\textwidth]{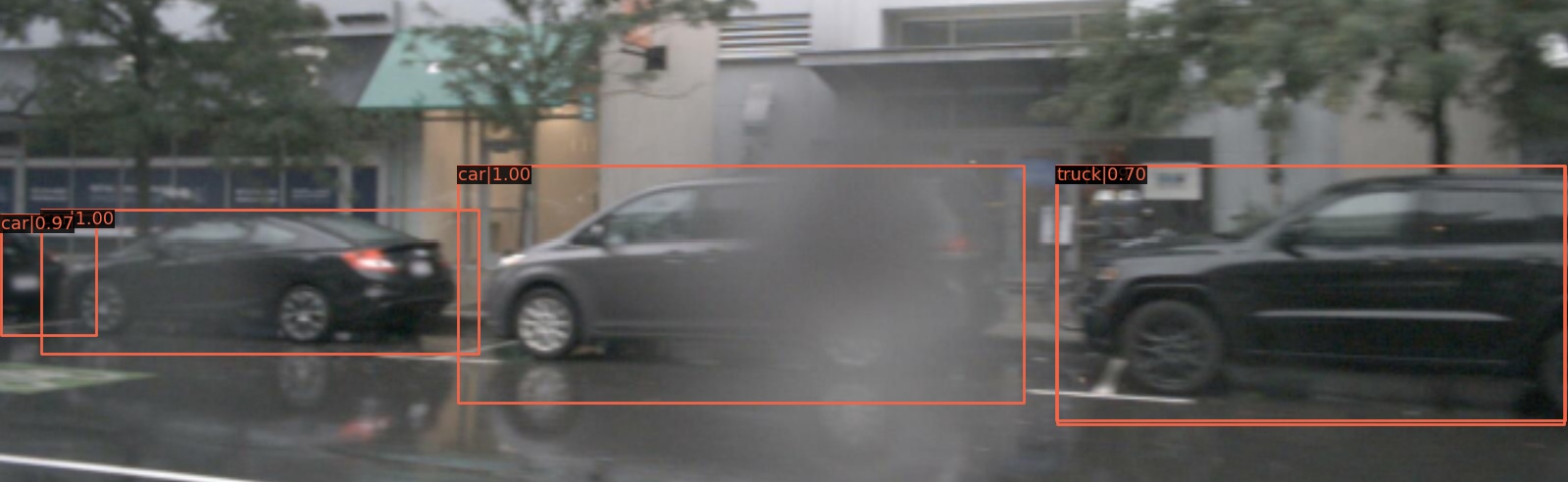} &
        \includegraphics[clip,width=0.245\textwidth,trim=1 108 0 300]{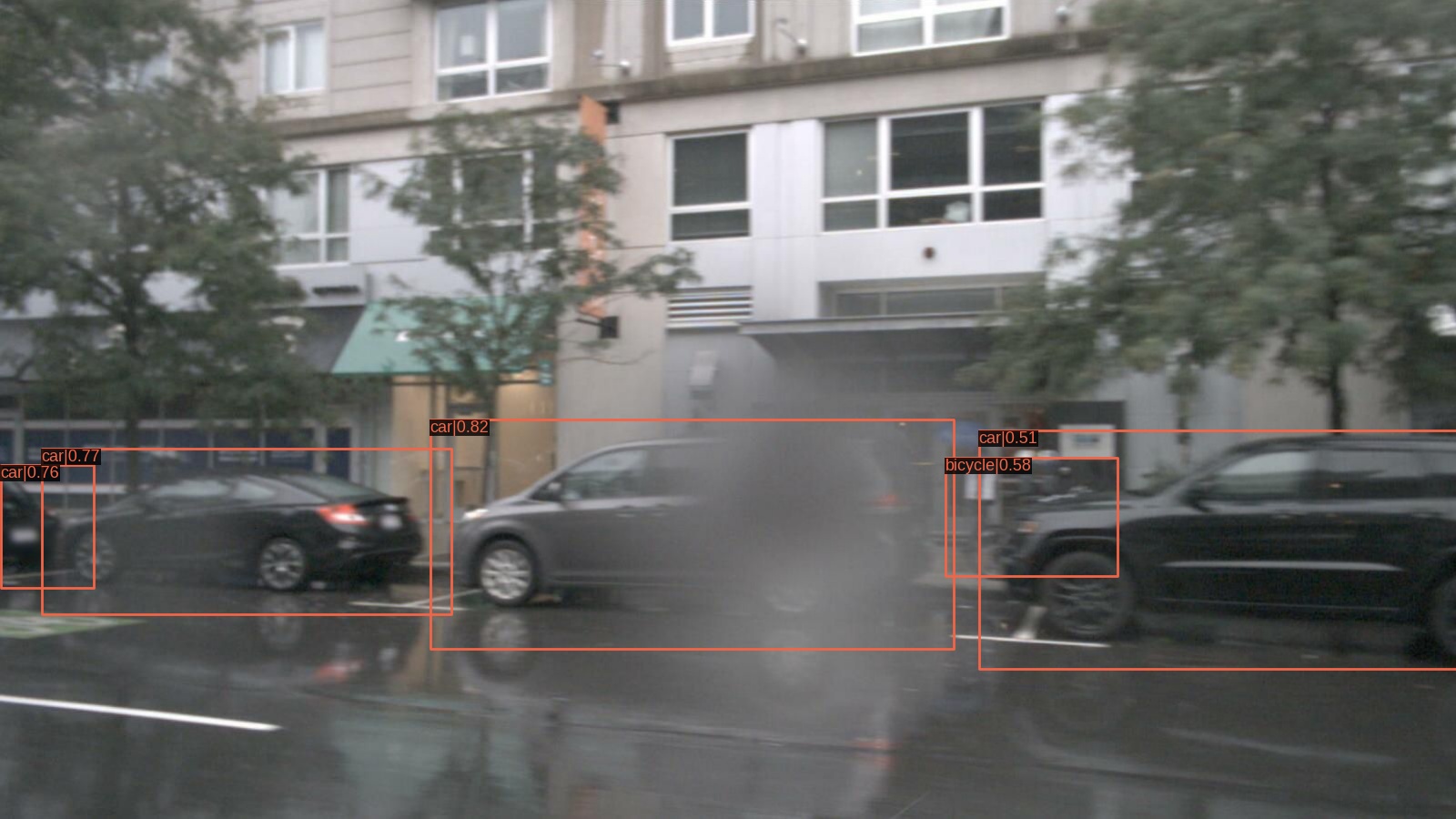} &
        \includegraphics[width=0.245\textwidth]{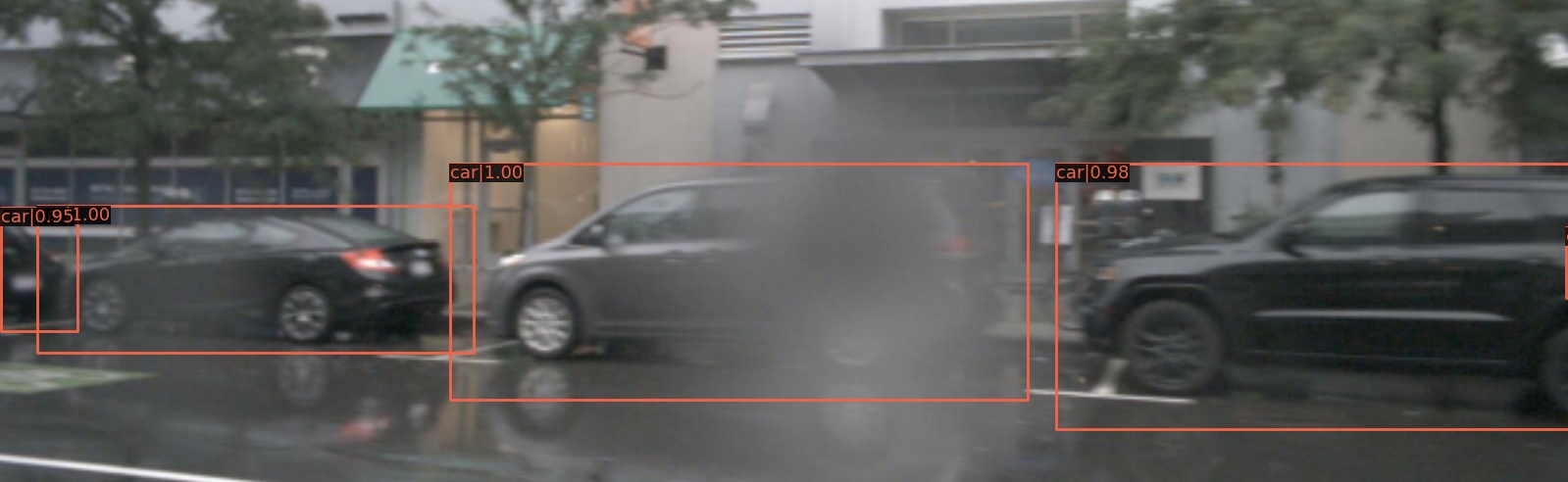} \\
        \vspace{-0.07cm}
        \includegraphics[width=0.245\textwidth]{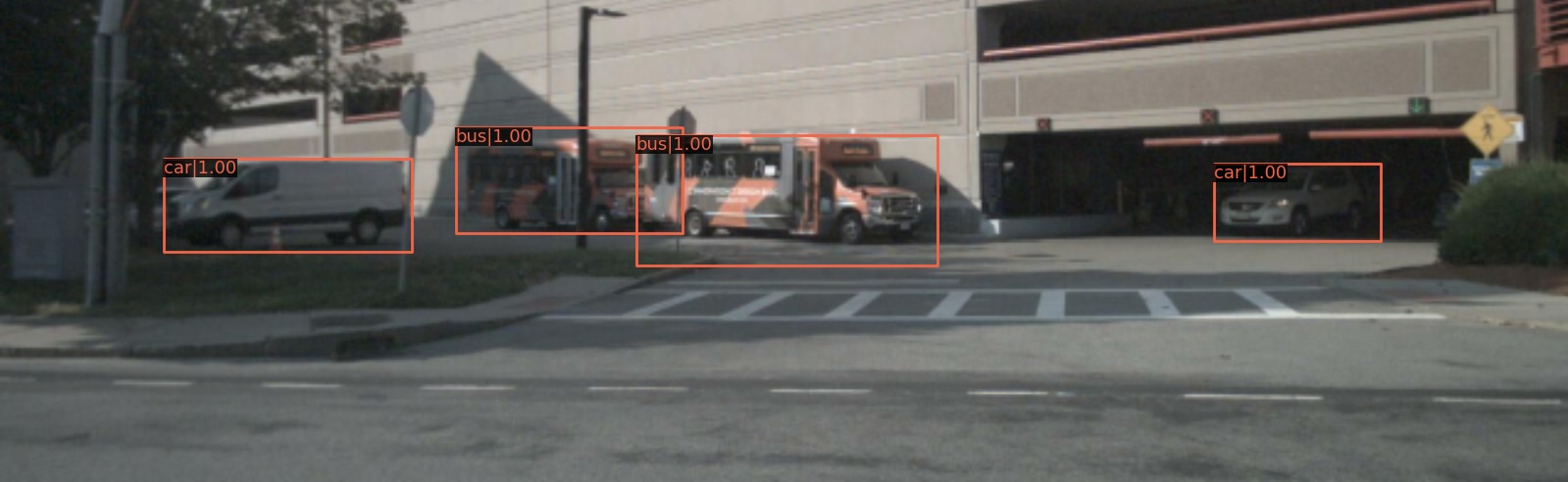} &
        \includegraphics[width=0.245\textwidth]{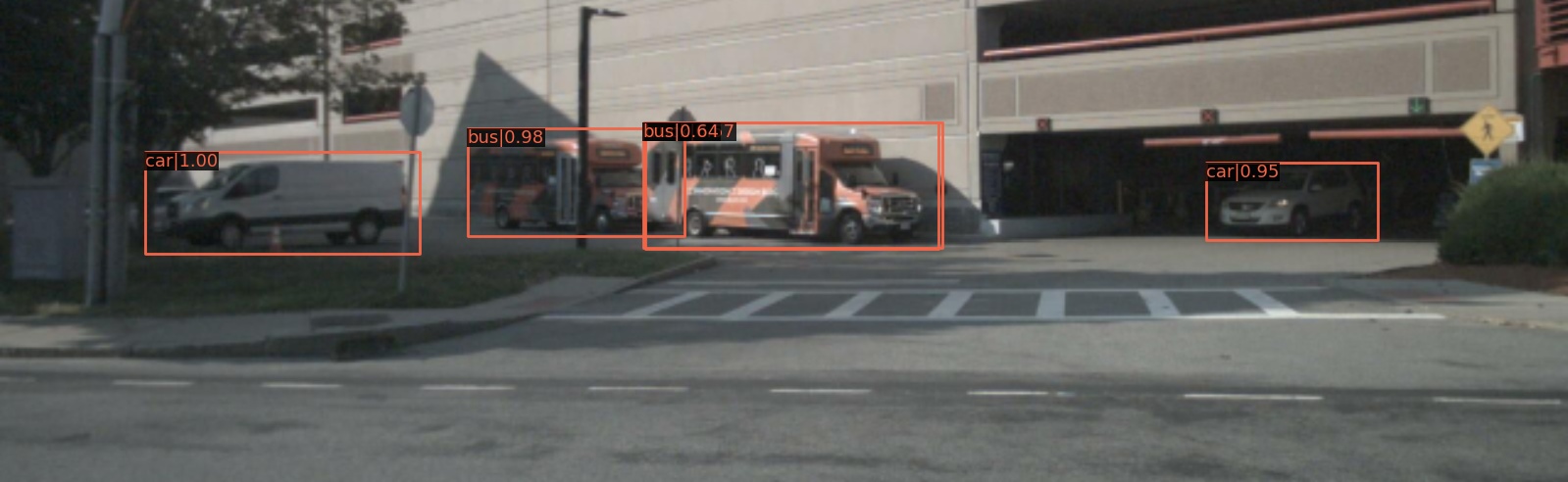} &
        \includegraphics[clip,width=0.245\textwidth,trim=1 108 0 300]{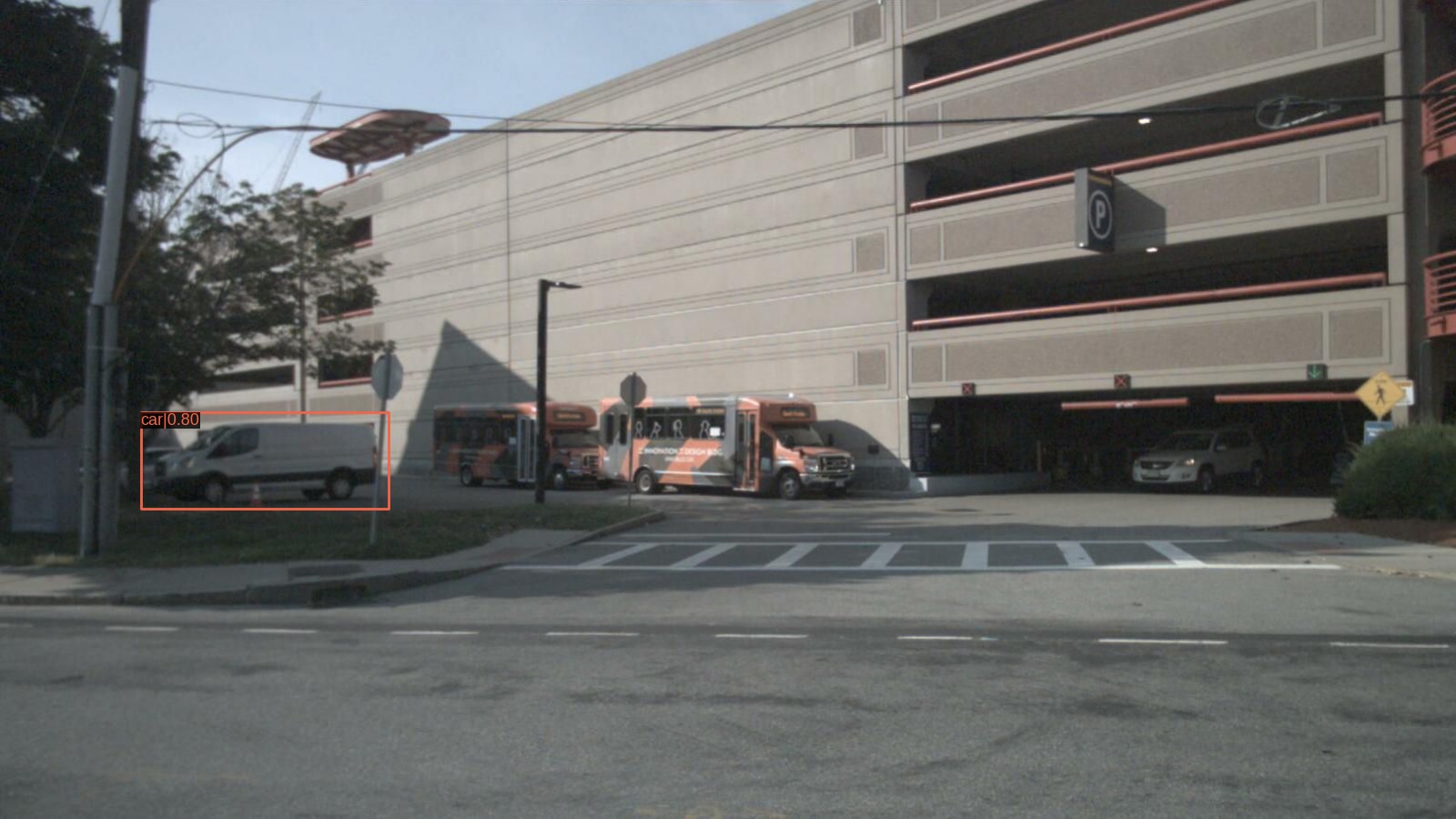} &
        \includegraphics[width=0.245\textwidth]{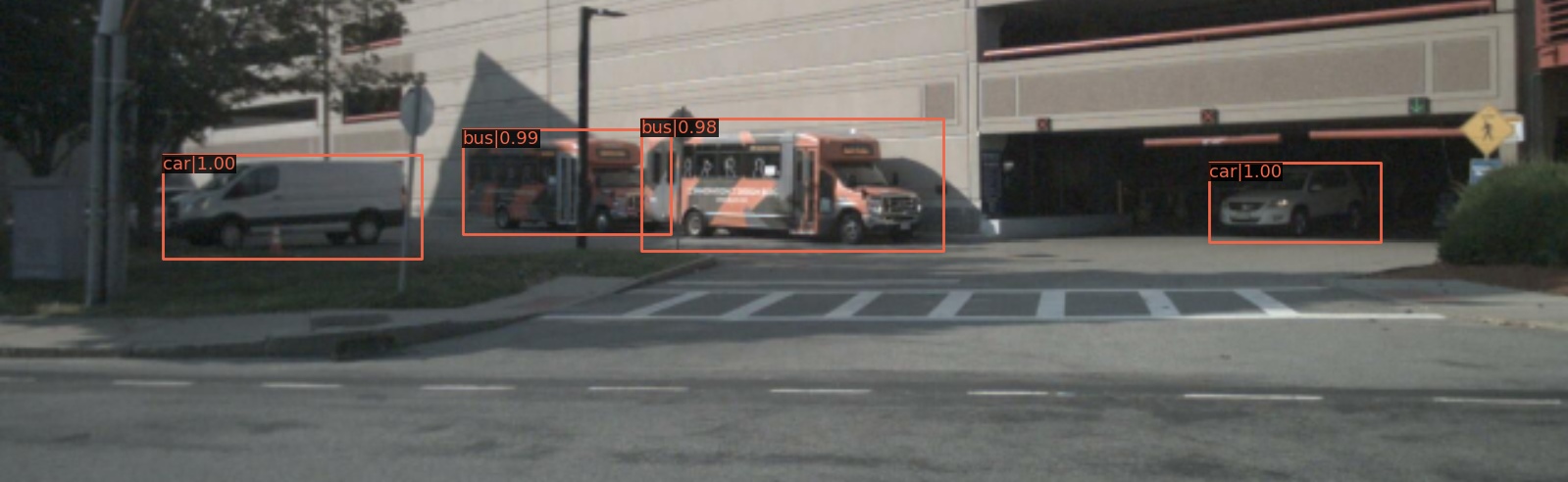} \\
        \vspace{-0.07cm}
        \includegraphics[width=0.245\textwidth]{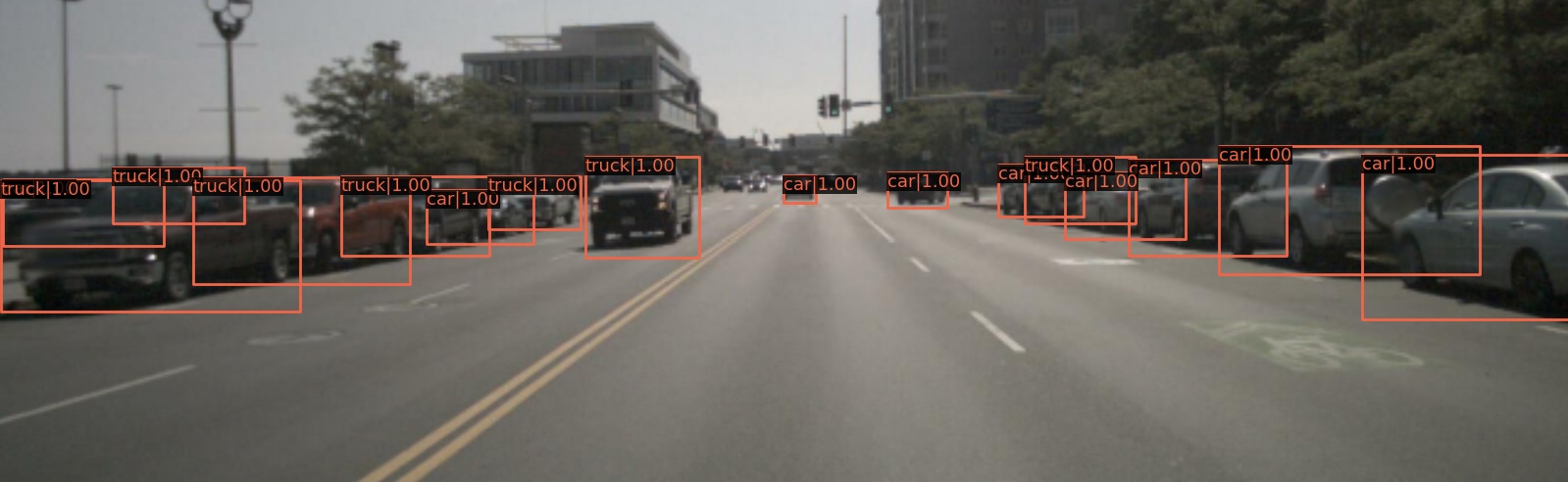} &
        \includegraphics[width=0.245\textwidth]{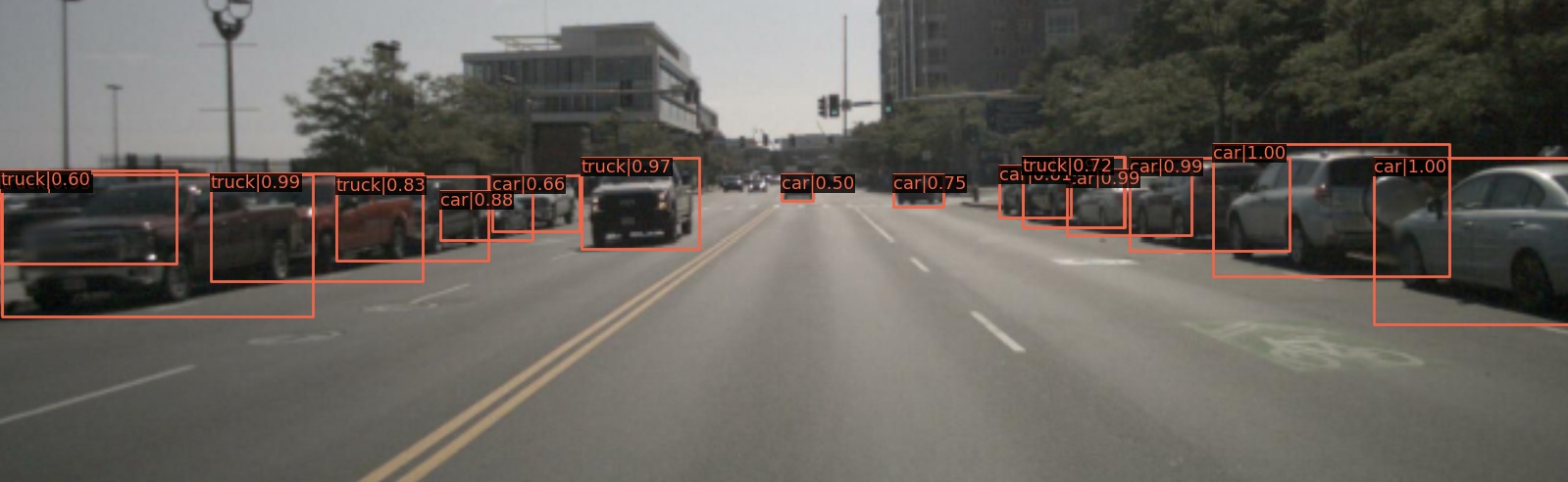} &
        \includegraphics[clip,width=0.245\textwidth,trim=1 108 0 300]{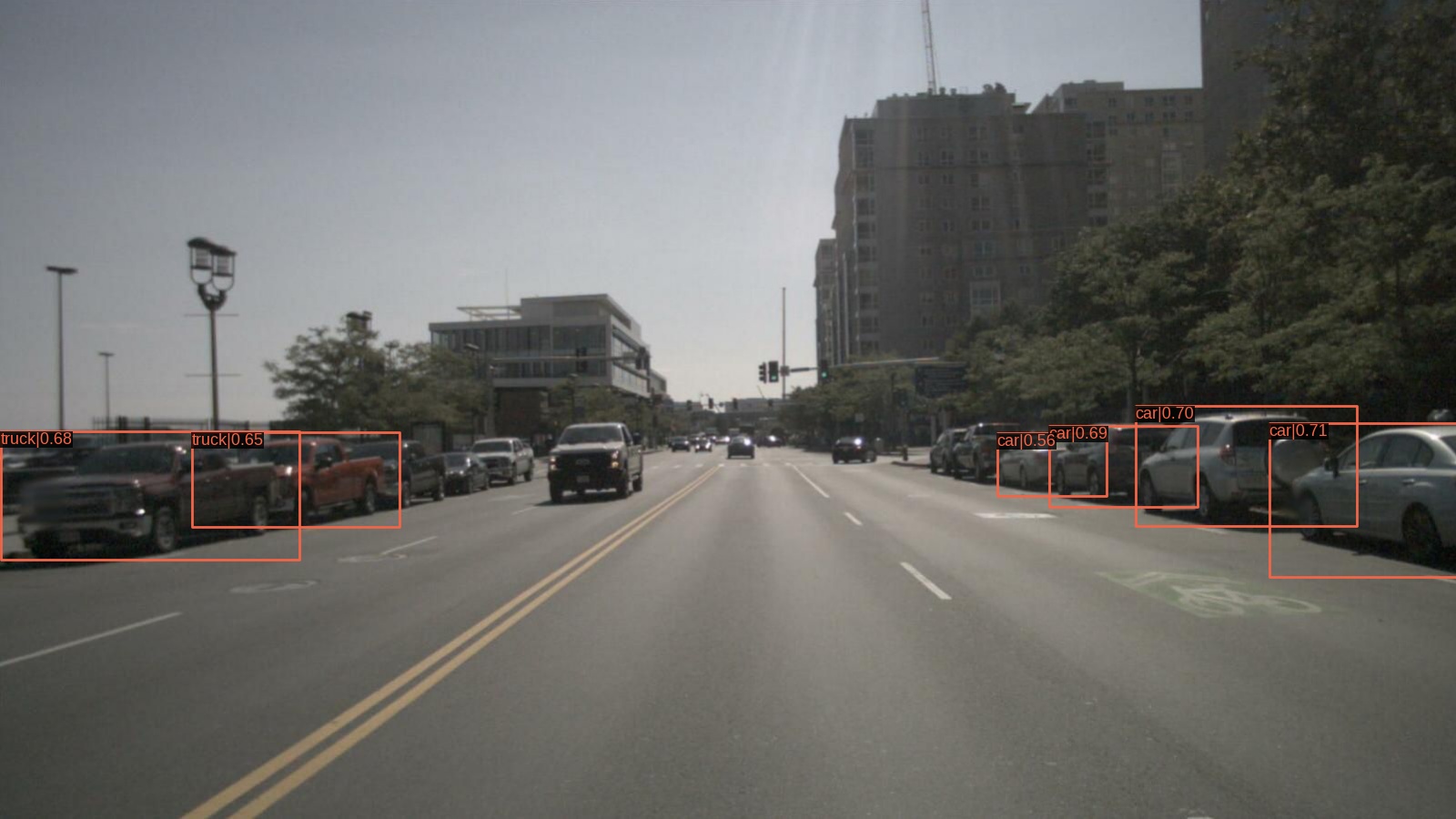} &
        \includegraphics[width=0.245\textwidth]{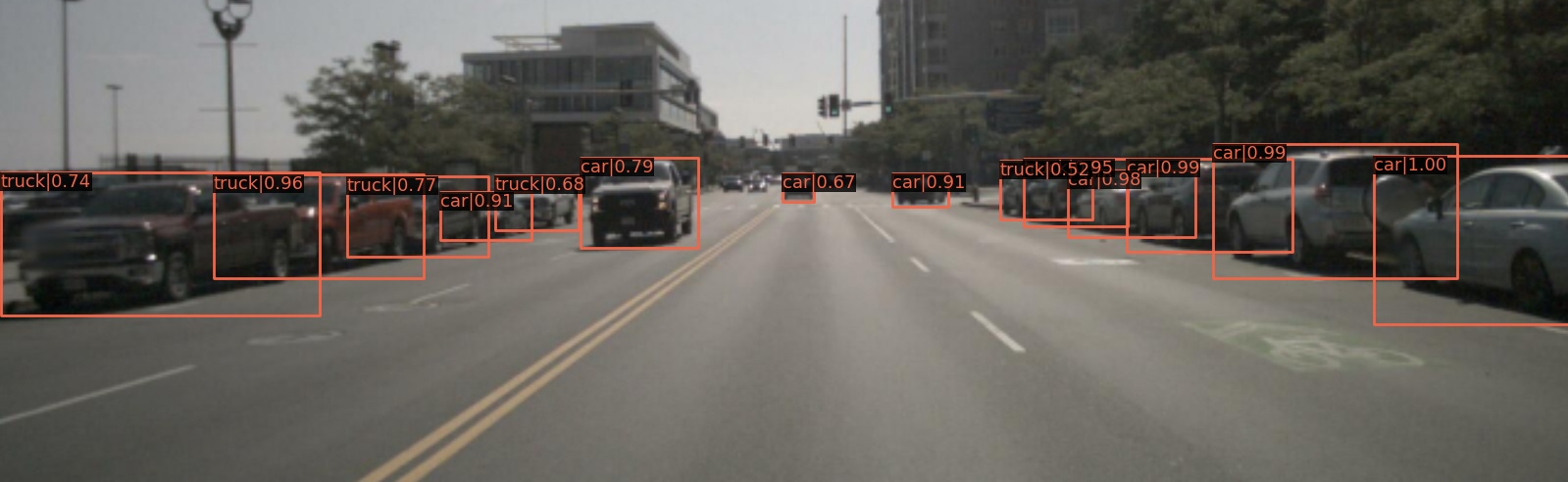} \\
        \vspace{-0.07cm}
        \includegraphics[width=0.245\textwidth]{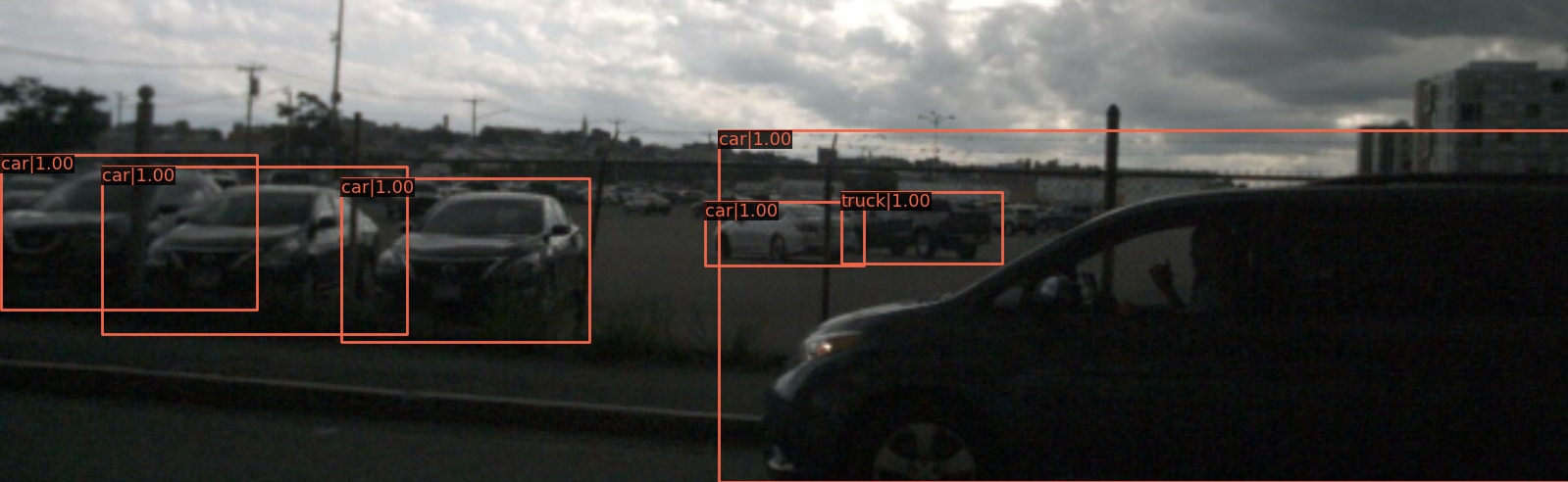}&
        \includegraphics[width=0.245\textwidth]{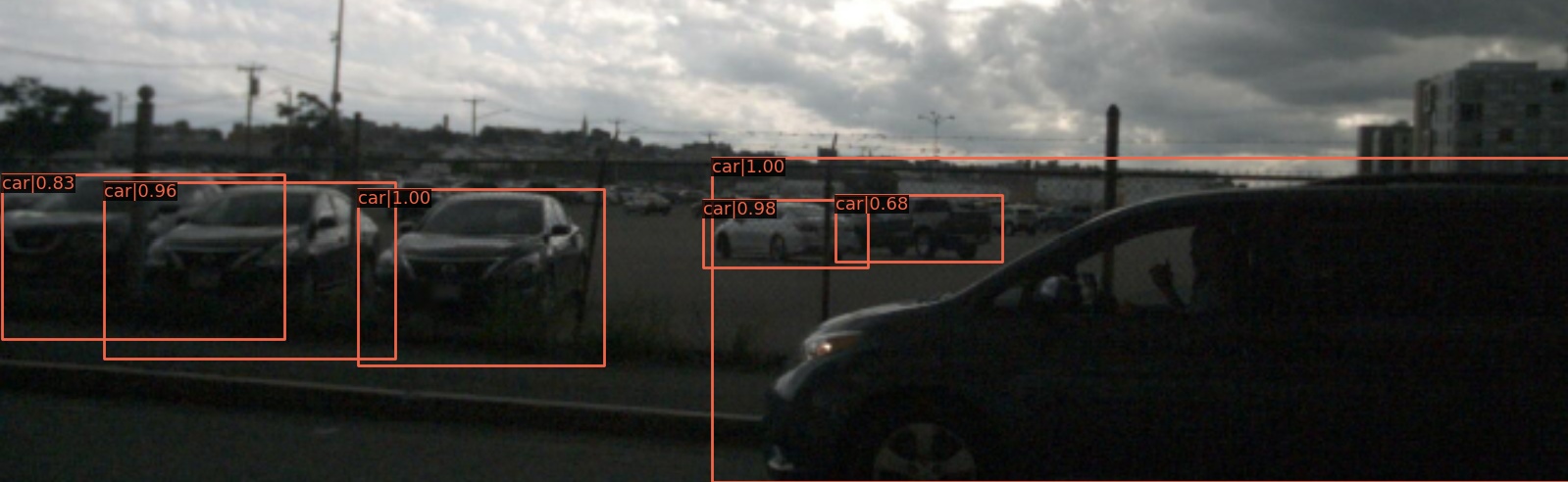}&
        \includegraphics[clip,width=0.245\textwidth,trim=1 108 0 300]{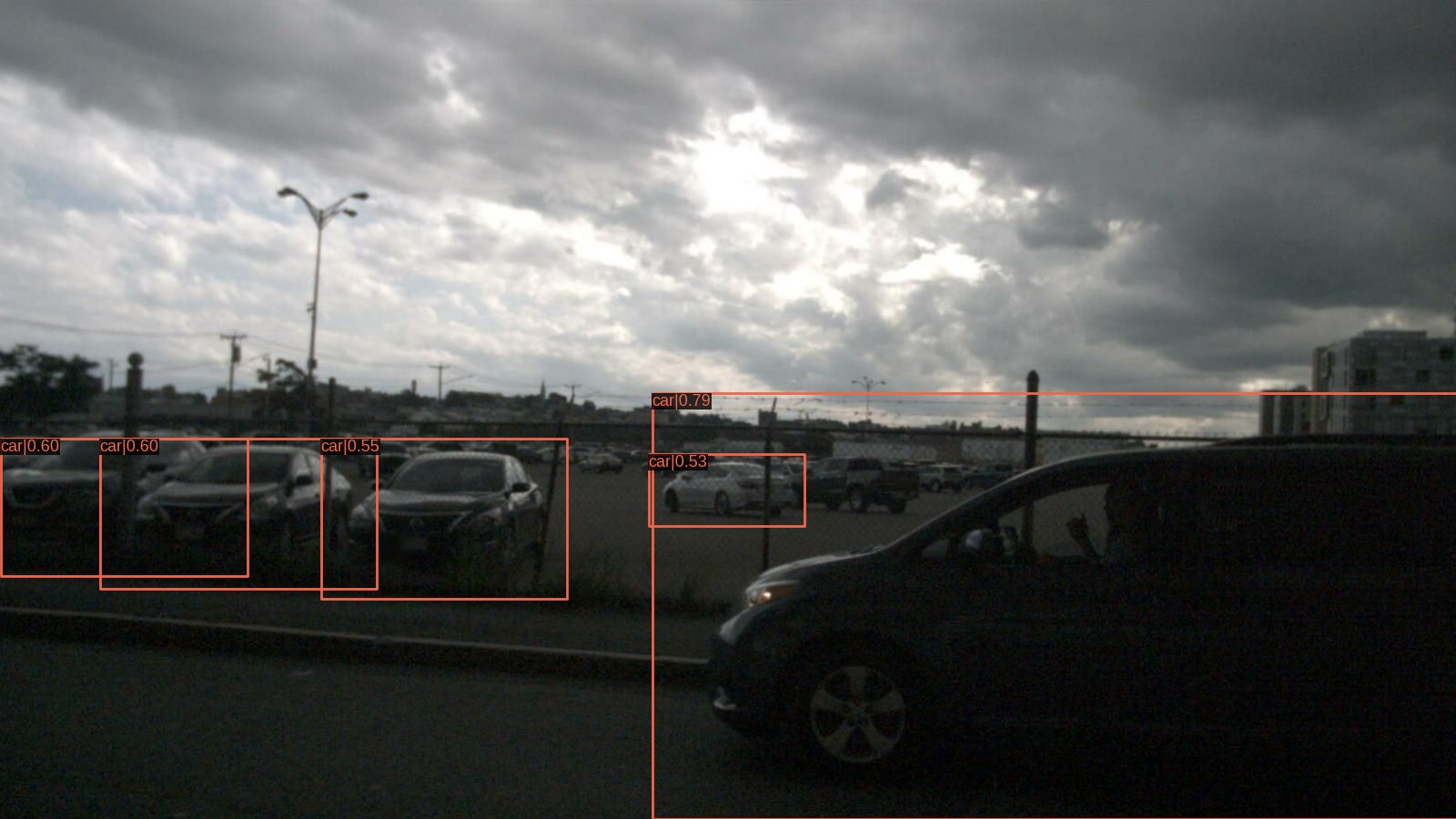}&
        \includegraphics[width=0.245\textwidth]{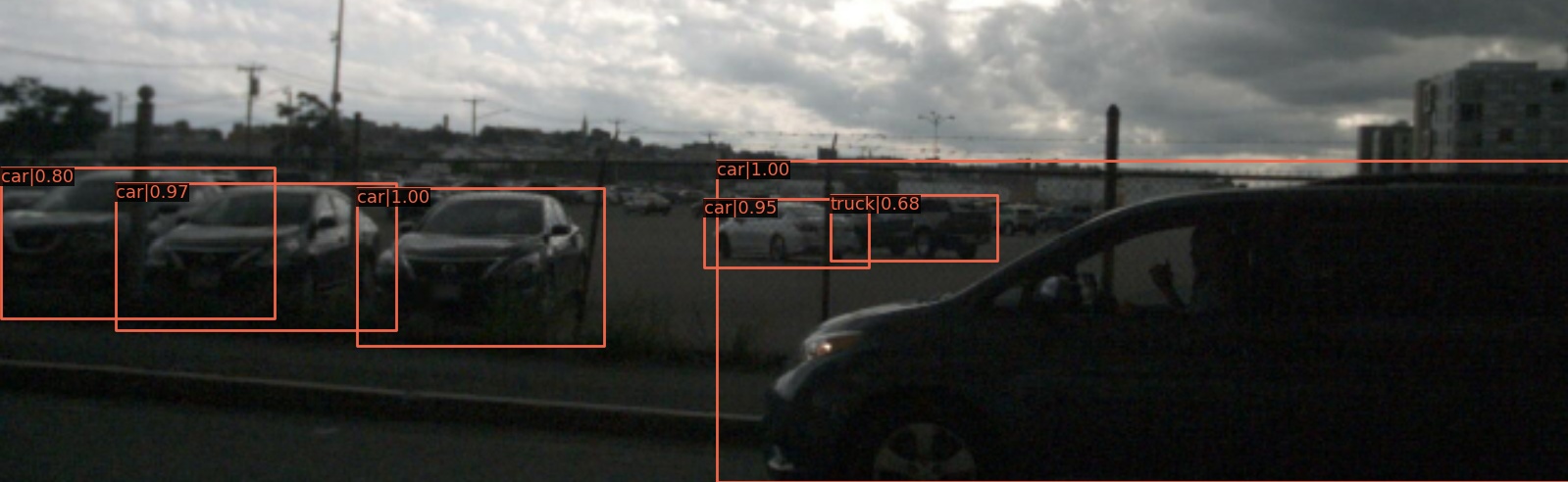} \\
        \includegraphics[width=0.245\textwidth]{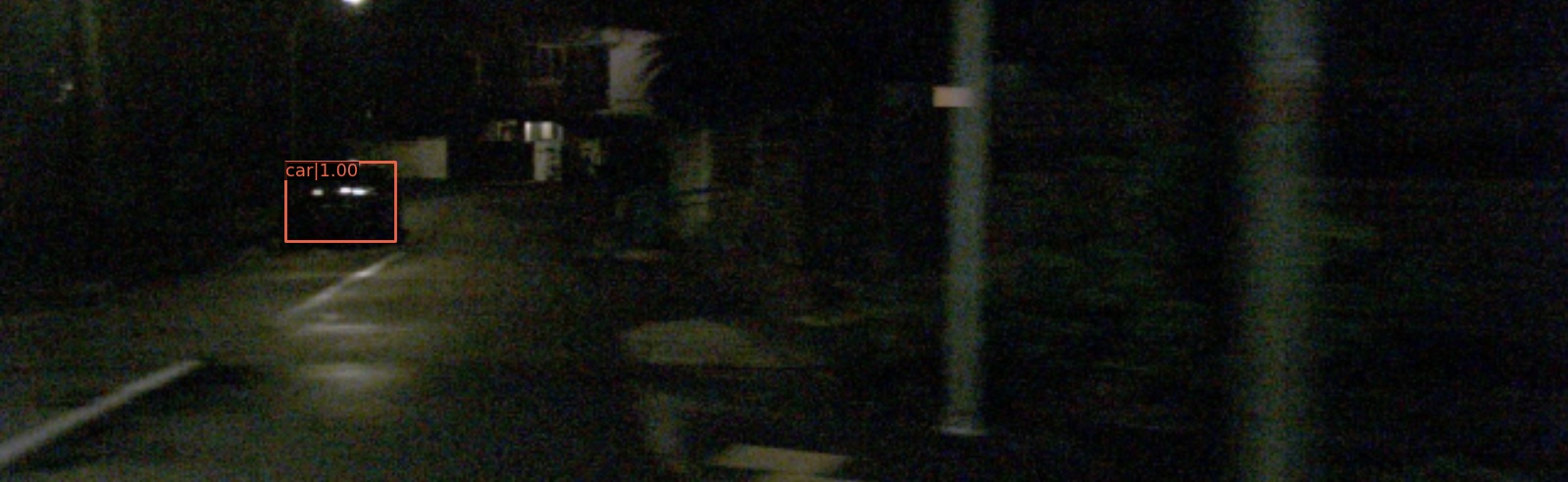}&
        \includegraphics[width=0.245\textwidth]{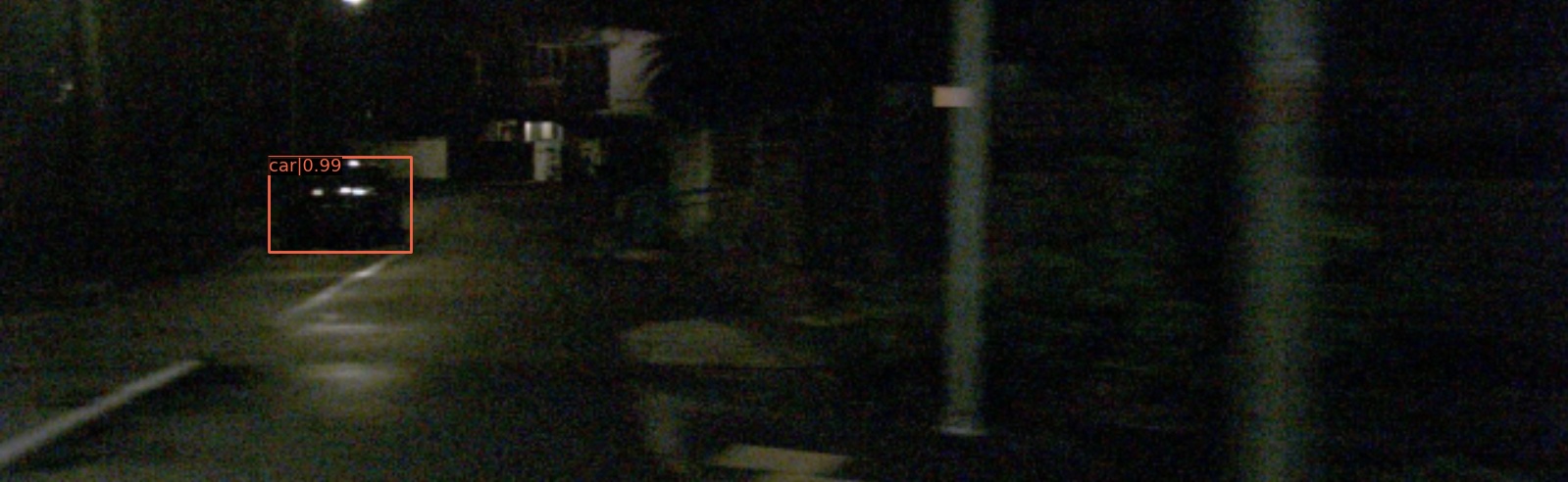}&
        \includegraphics[clip,width=0.245\textwidth,trim=1 108 0 300]{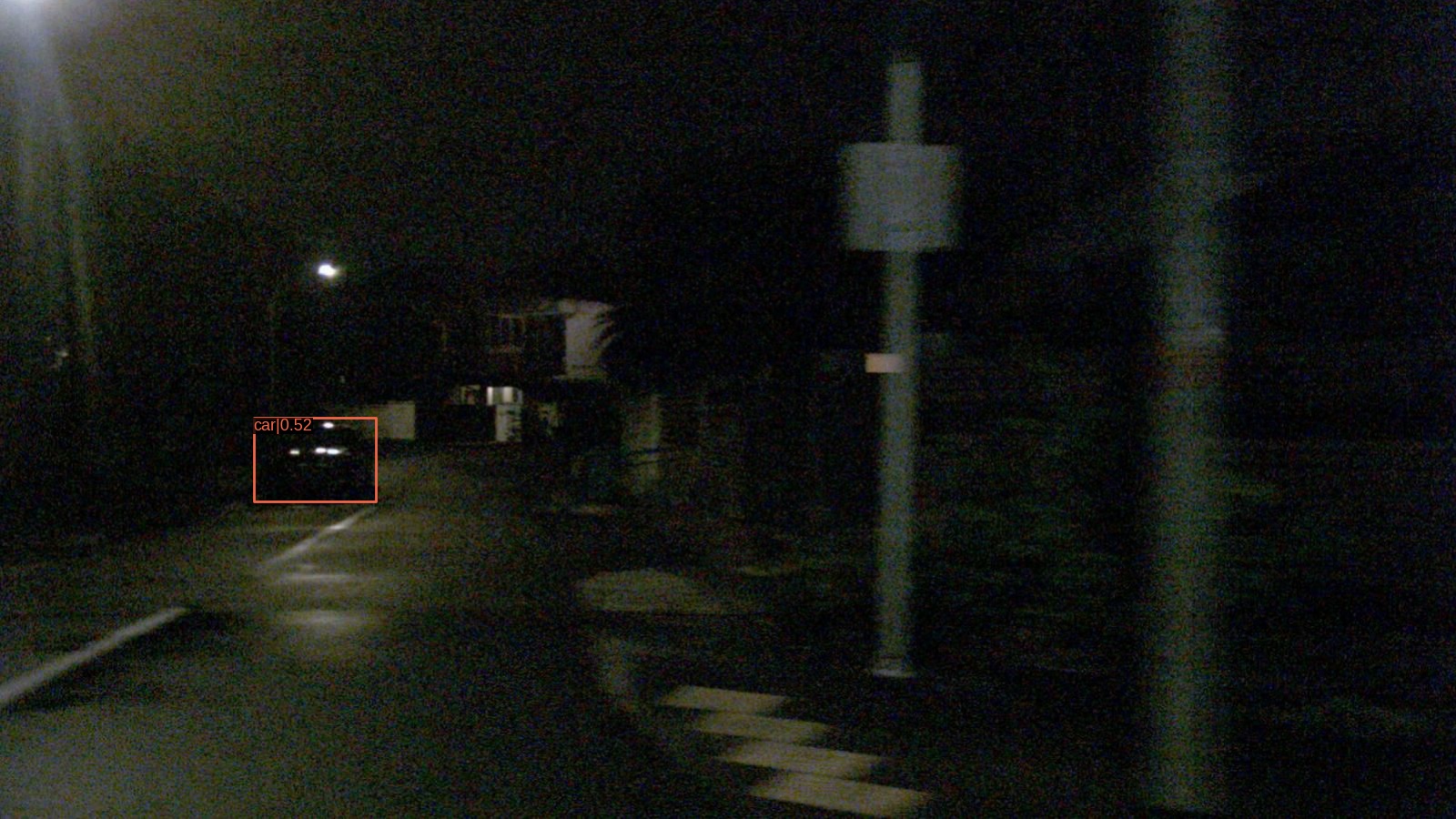}&
        \includegraphics[width=0.245\textwidth]{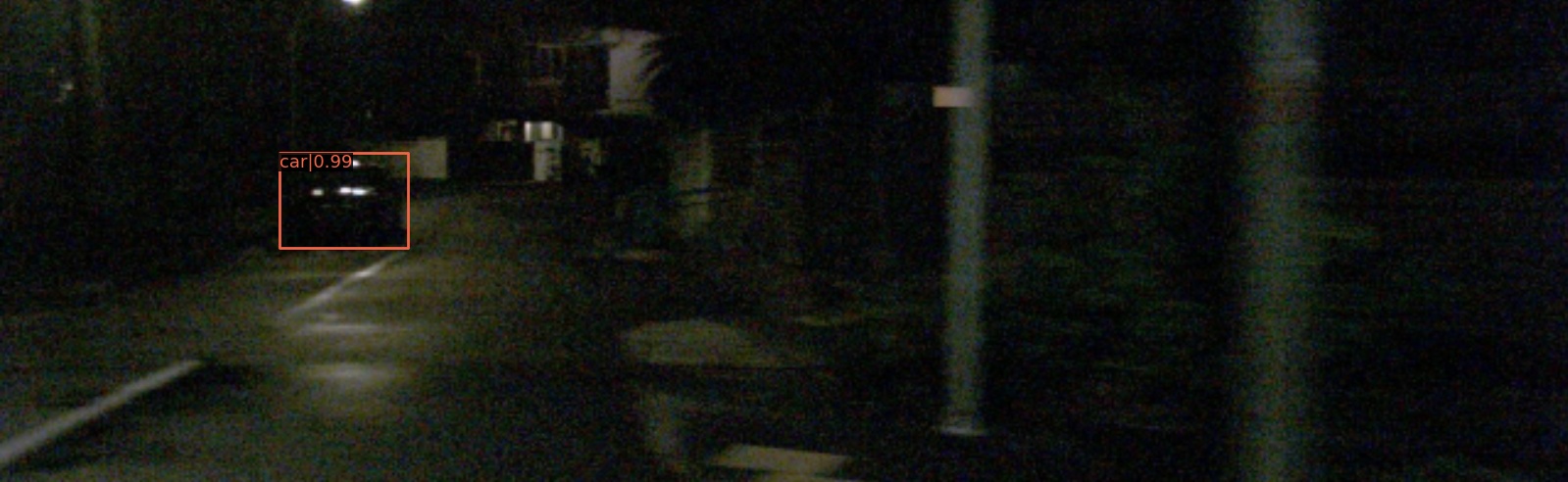}
    \end{tabular}
    \caption{Further qualitative detection results on nuScenes with 6 classes visualized: car, truck, bus, bicycle, motorcycle and pedestrian. From left to right: image with ground-truth annotation, prediction of HRFormer, 3D$\to$2D projected predictions of BEVFusion, prediction of HRFuser. Best viewed on a screen at full zoom.}
    \label{fig:further:results:nuscenes}
\end{figure*}

\begin{figure*}
    \centering
    \begin{tabular}{@{}c@{\hspace{0.05cm}}c@{\hspace{0.05cm}}c@{\hspace{0.05cm}}c@{}}
        \subfloat{\footnotesize Ground Truth} &
        \subfloat{\footnotesize HRFormer~\cite{hrformer}} &
        \subfloat{\footnotesize BEVFusion~\cite{liu2022bevfusion}} &
        \subfloat{\footnotesize HRFuser} \\
        \vspace{-0.07cm}
        \includegraphics[width=0.245\textwidth]{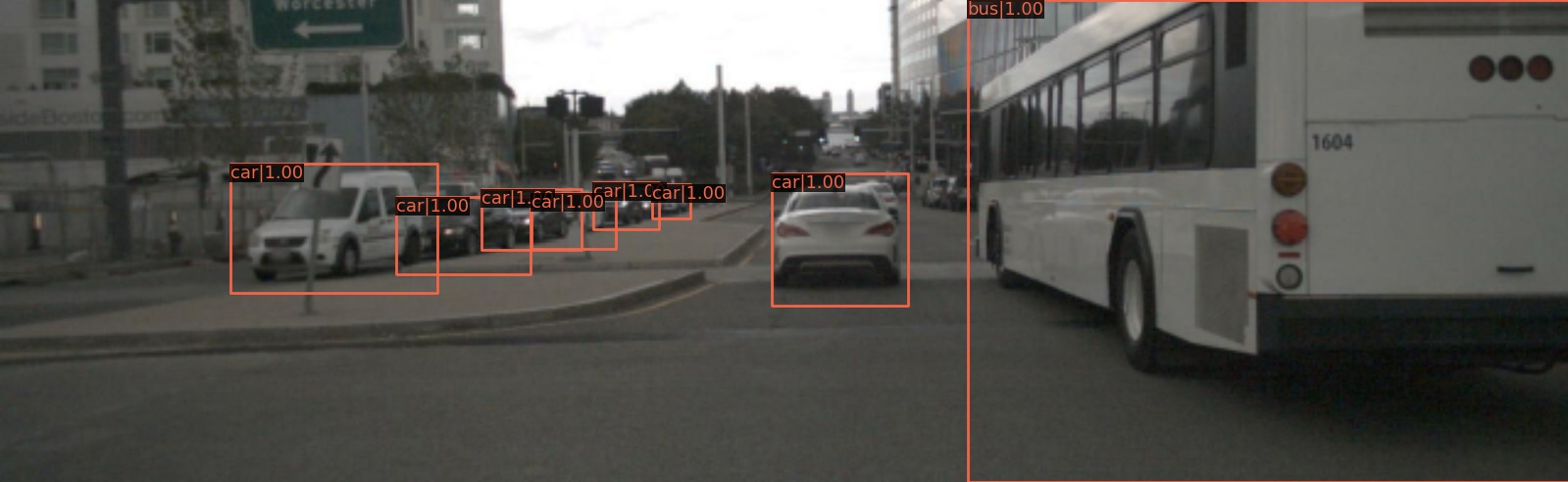}&
        \includegraphics[width=0.245\textwidth]{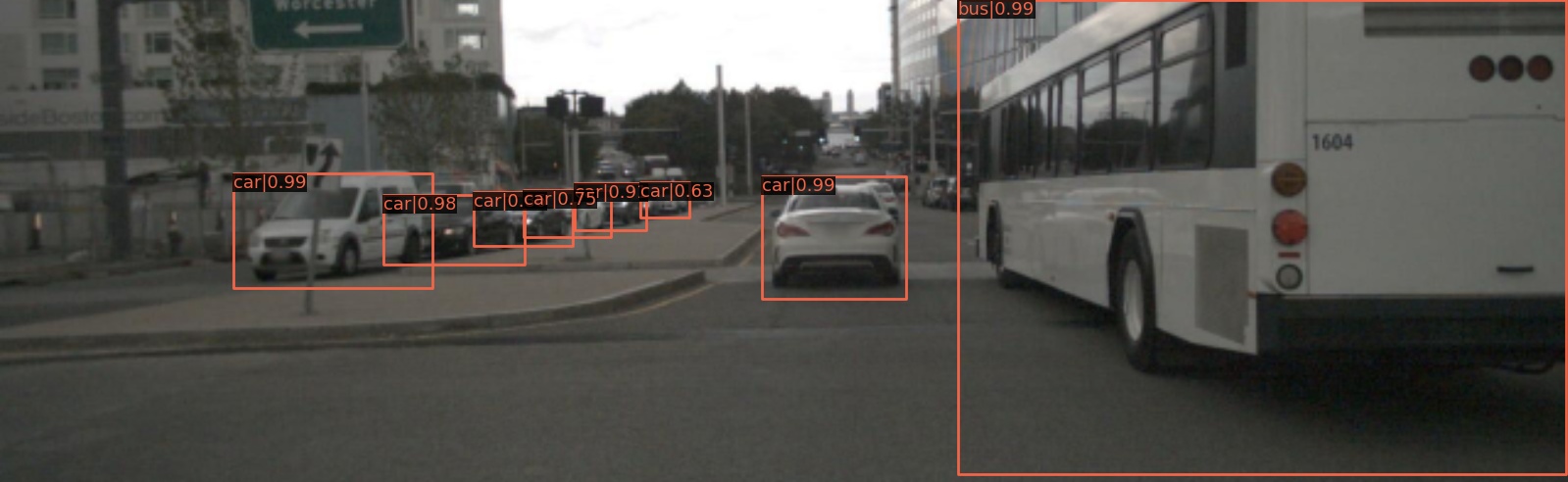}&
        \includegraphics[clip,width=0.245\textwidth,trim=1 108 0 300]{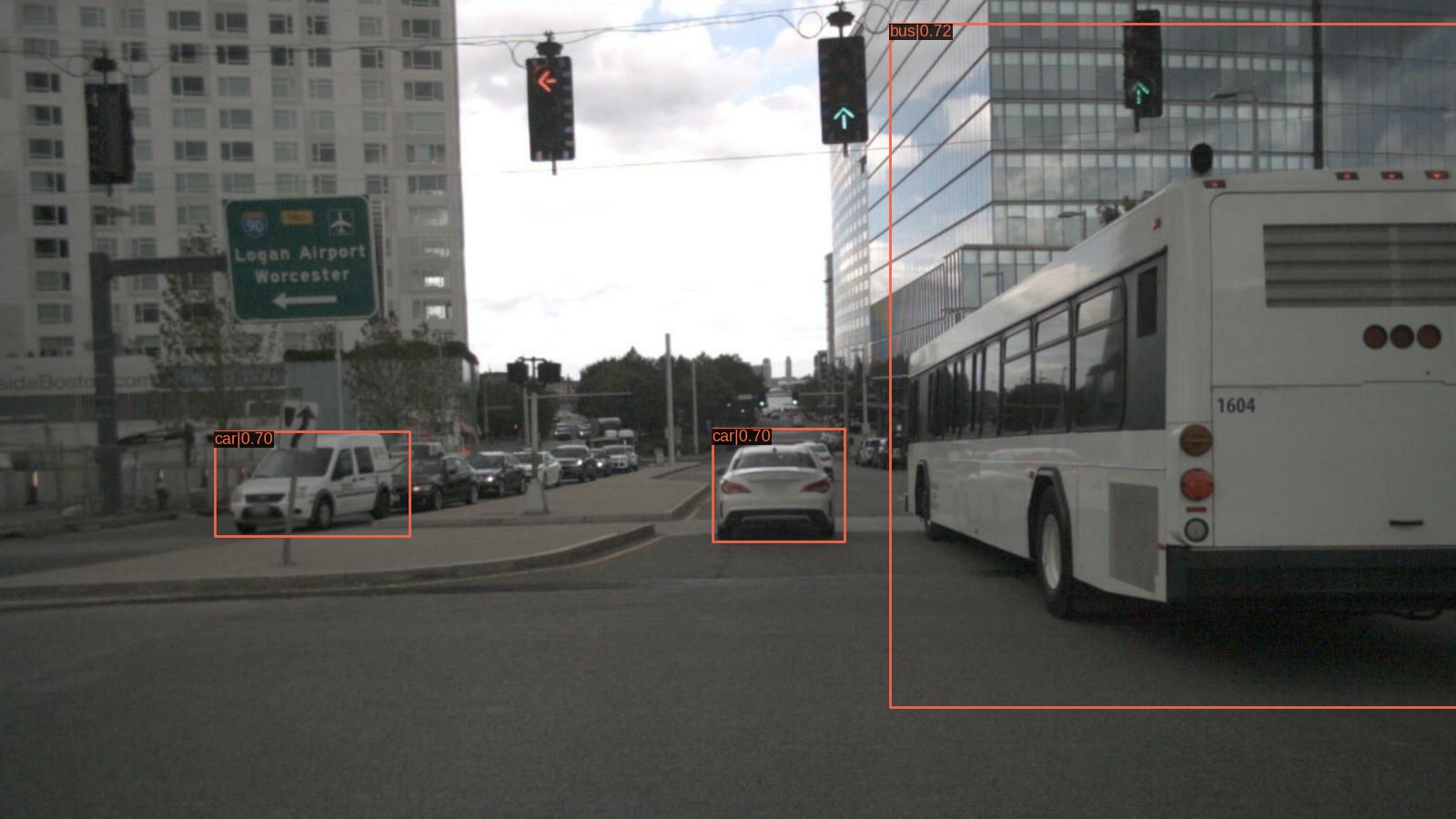}&
        \includegraphics[width=0.245\textwidth]{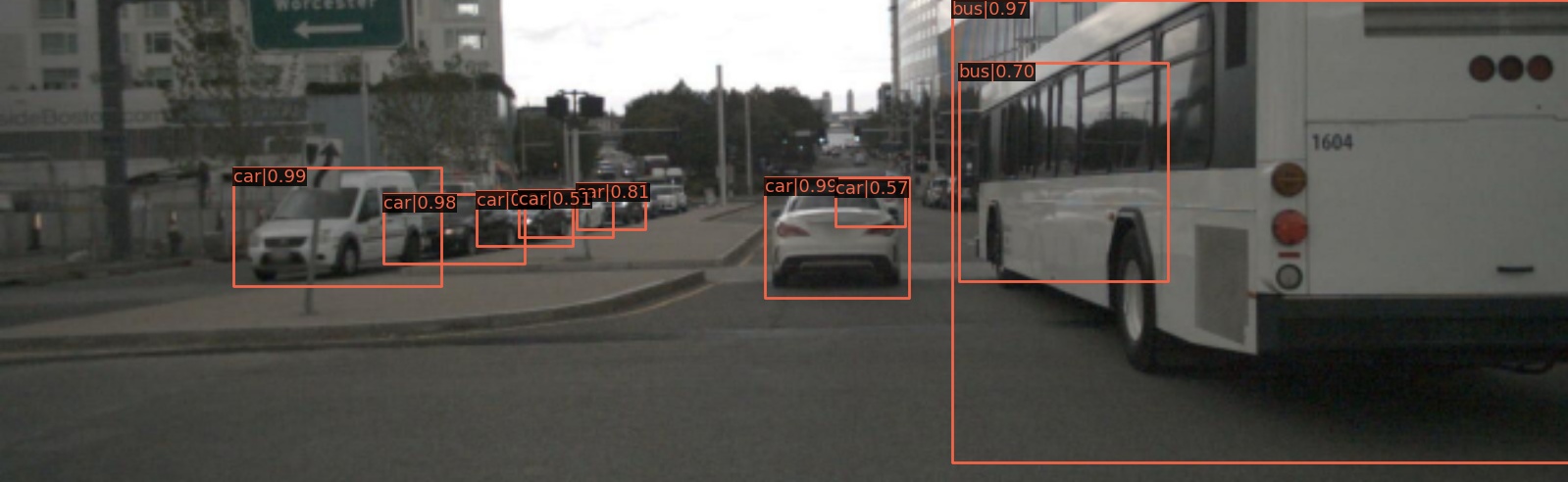}
        \\
        \includegraphics[width=0.245\textwidth]{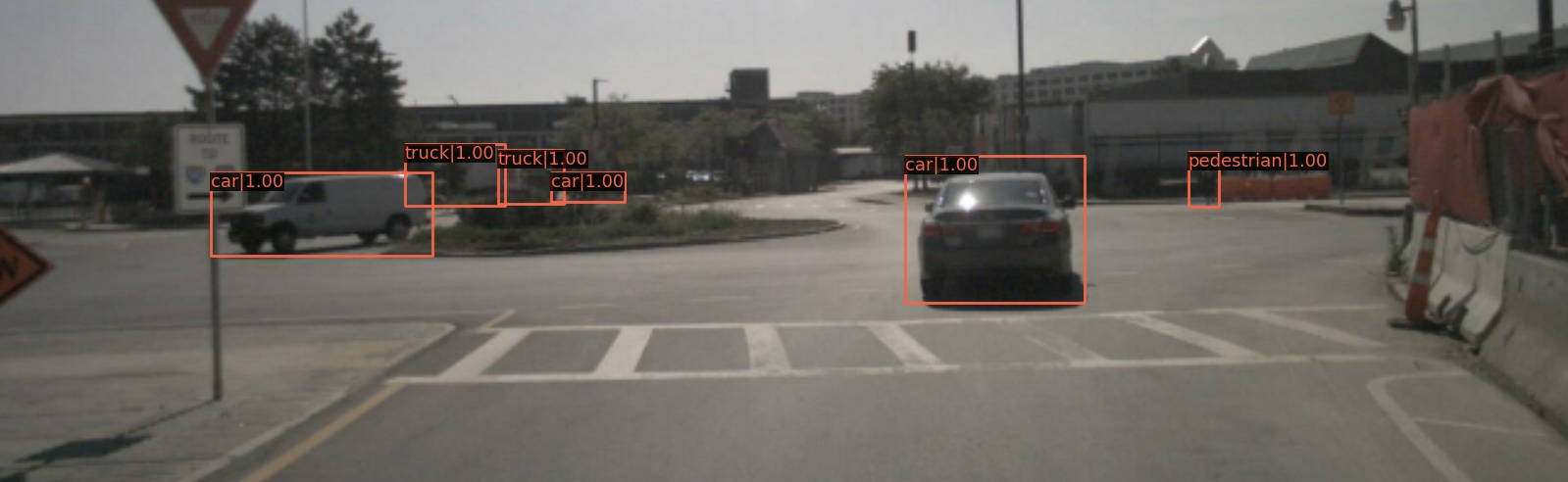}&
        \includegraphics[width=0.245\textwidth]{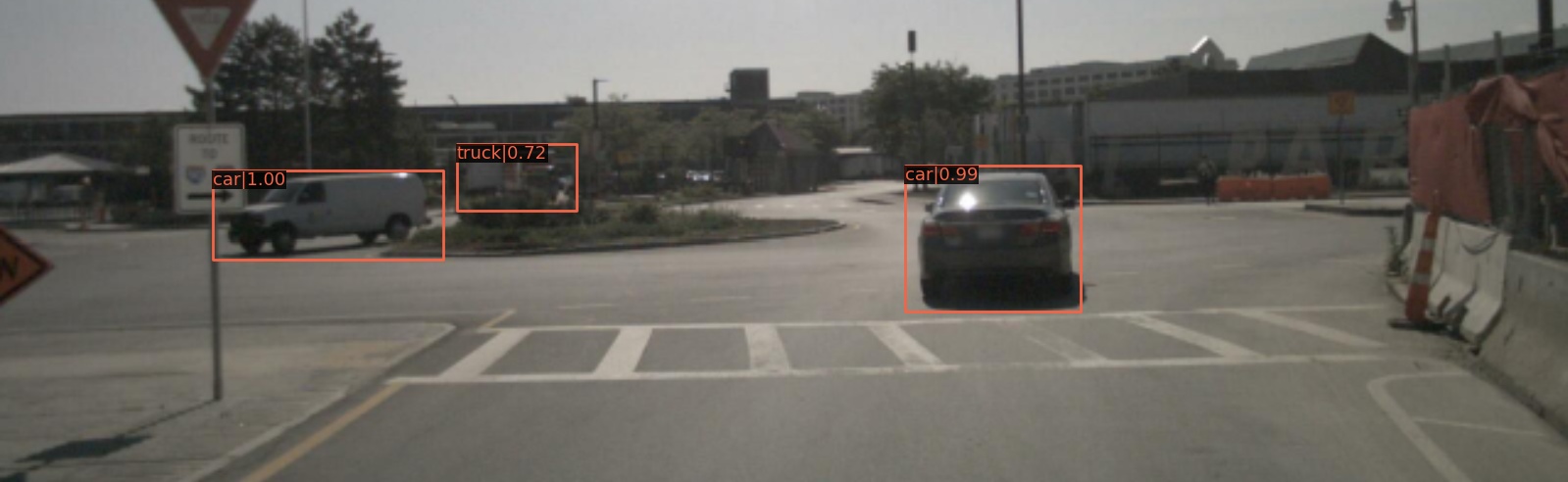}&
        \includegraphics[clip,width=0.245\textwidth,trim=1 108 0 300]{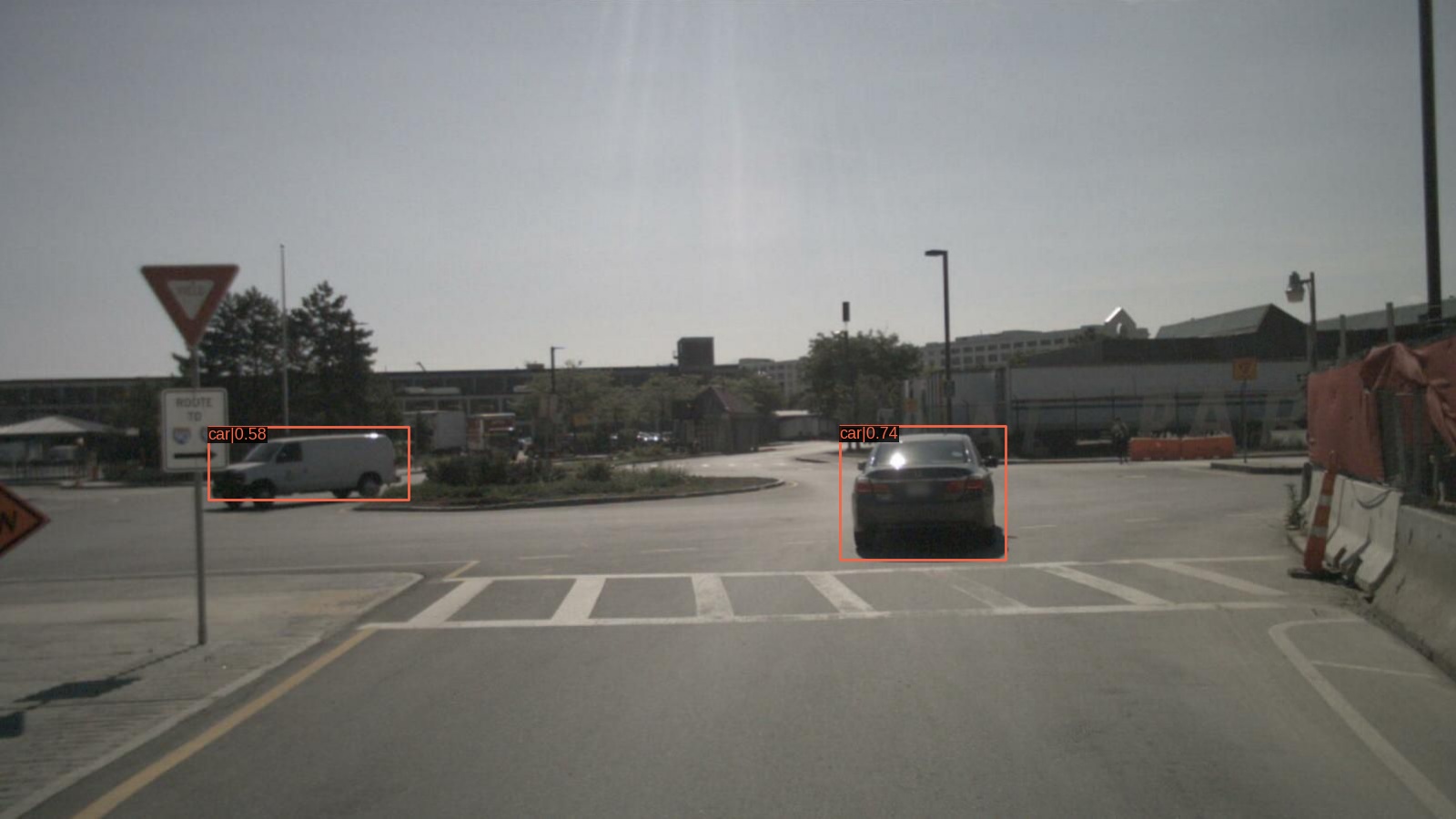}&
        \includegraphics[width=0.245\textwidth]{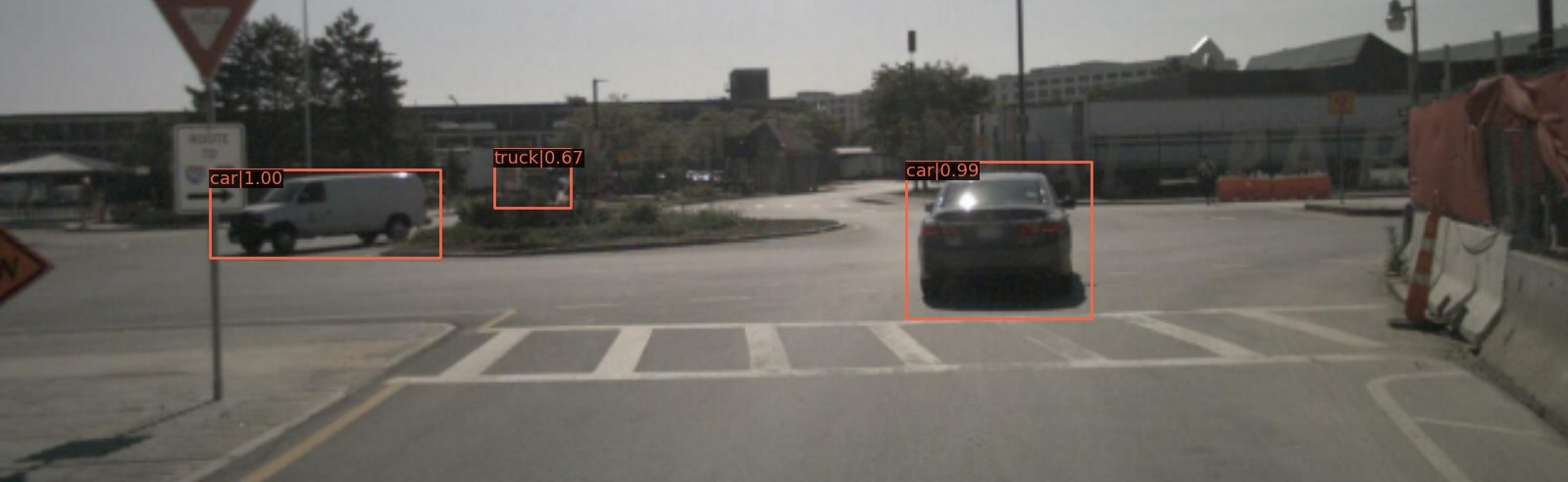}
    \end{tabular}        
    \caption{Further qualitative detection results on nuScenes showing failure cases of HRFuser with 6 classes visualized: car, truck, bus, bicycle, motorcycle and pedestrian. From left to right: image with ground-truth annotation, prediction of HRFormer, 3D$\to$2D projected predictions of BEVFusion, prediction of HRFuser. Best viewed on a screen at full zoom.}
    \label{fig:further:results:nuscenes:failure}
\end{figure*}

We show further qualitative results on DENSE in \cref{fig:further:results:stf} with example failure cases in \cref{fig:further:results:stf:failure} and on nuScenes in \cref{fig:further:results:nuscenes} with example failure cases in \cref{fig:further:results:nuscenes:failure}.

\end{document}